\documentclass[10pt,a4paper]{article}
\usepackage[
    top=4cm,
    bottom=4cm,
    left=5cm,
    right=8cm,
    marginparwidth=2.5cm,
    marginparsep=2cm
]{geometry}
\usepackage{marginnote}
\usepackage{dsfont}
\usepackage{background}
\usepackage{xcolor} 
\usepackage{subcaption}
\usepackage[numbers]{natbib}
\definecolor{lightline}{gray}{0.8}
\definecolor{notegray}{gray}{0.5}

\newlength{\customlineheight}
\backgroundsetup{
    scale=1,
    color=lightline,
    opacity=0.75,
    angle=0,
    position={current page.east},
    hshift=-4cm,
    vshift=0cm,
    contents={\rule{0.2pt}{\customlineheight}}
}

\usepackage{ragged2e}

\newcommand{\greyline}{%
    \par\noindent
    \begin{tikzpicture}
        \draw[gray!40, line width=0.3pt] (0,0) -- (0.3\textwidth,0);
    \end{tikzpicture}
    \par\vspace{0em}
}

\newcommand{\greylinelong}{%
    \par\noindent
    \begin{tikzpicture}
        \draw[gray!40, line width=0.5pt] (0,0) -- (0.8\textwidth,0);
    \end{tikzpicture}
    \par\vspace{-1em}
}

\usepackage{times}
\usepackage{epsfig}
\usepackage{graphicx}
\usepackage{amsmath}
\usepackage{amssymb}
\usepackage{overpic}
\usepackage{appendix}
\usepackage{tcolorbox}
\usepackage{tikz}
\usetikzlibrary{patterns}
\usetikzlibrary{arrows.meta}
\usetikzlibrary{arrows.meta,shapes.geometric}

\usepackage{cancel}

\usepackage{amsthm}
\usepackage{setspace}
\usepackage[boxed]{algorithm2e}
\usepackage{bm}
\usepackage{fancyhdr}

\newtheorem{theorem}{Theorem}

\newcommand{\itab}[1]{\hspace{0em}\rlap{#1}}

\renewcommand\det{\mathrm{det}\,}

\begin{document}

\title{Mathematical Foundations \\ of Geometric Deep Learning}
\date{}
\author{Haitz S\'{a}ez de Oc\'{a}riz Borde and Michael Bronstein}

\maketitle
\vspace{-1cm}
\begin{center}
    University of Oxford
\end{center}

\vfill

\textit{We review\marginnote{These notes were originally developed by Haitz S\'{a}ez de Oc\'{a}riz Borde for the ANAIS 2024 Geometric Deep Learning course in Kathmandu, Nepal. They are based on Michael Bronstein's notes for the 2019 Computer Vision and Pattern Recognition course~\cite{bronstein2019cvpr} at USI Lugano, Switzerland, as well as lecture slides from the 2024 Geometric Deep Learning course~\cite{bronstein2024oxford} at the University of Oxford, United Kingdom.} the key mathematical concepts necessary for studying Geometric Deep Learning~\cite{bronstein2021geometricdeeplearninggrids}. For a deeper understanding of specific topics, we encourage supplementing studies with additional resources.}

\greylinelong

\section*{Introduction}

Since the dawn of civilization, humans have tried to understand the nature of intelligence. With the advent of computers, there have been attempts to emulate human intelligence using computer algorithms -- a field that was dubbed `Artificial Intelligence' or `AI' by the computer scientist John McCarthy in 1956 and has recently enjoyed an explosion of popularity. Many efforts in AI research have focused on the study and replication of what is considered the hallmark of human cognition, such as playing intelligent games, the faculty of language, visual perception, and creativity. While at the time of writing we have multiple successful takes at the above -- computers nowadays play chess and Go better than any human, can translate English into Chinese without a dictionary, automatically drive a car in a crowded city, and generate poetry and art that wins artistic competitions -- it is fair to say that we still do not have a full understanding of what human-like or `general' intelligence entails and how to replicate it.

Most of the aforementioned examples of AI\marginnote{
The Perceptron, introduced by Frank Rosenblatt in 1957, is perhaps the simplest form of an artificial neural network, consisting of only a single artificial neuron. Modern neural networks can contain millions of neurons with billions of weights.

\includegraphics[width=1\linewidth]{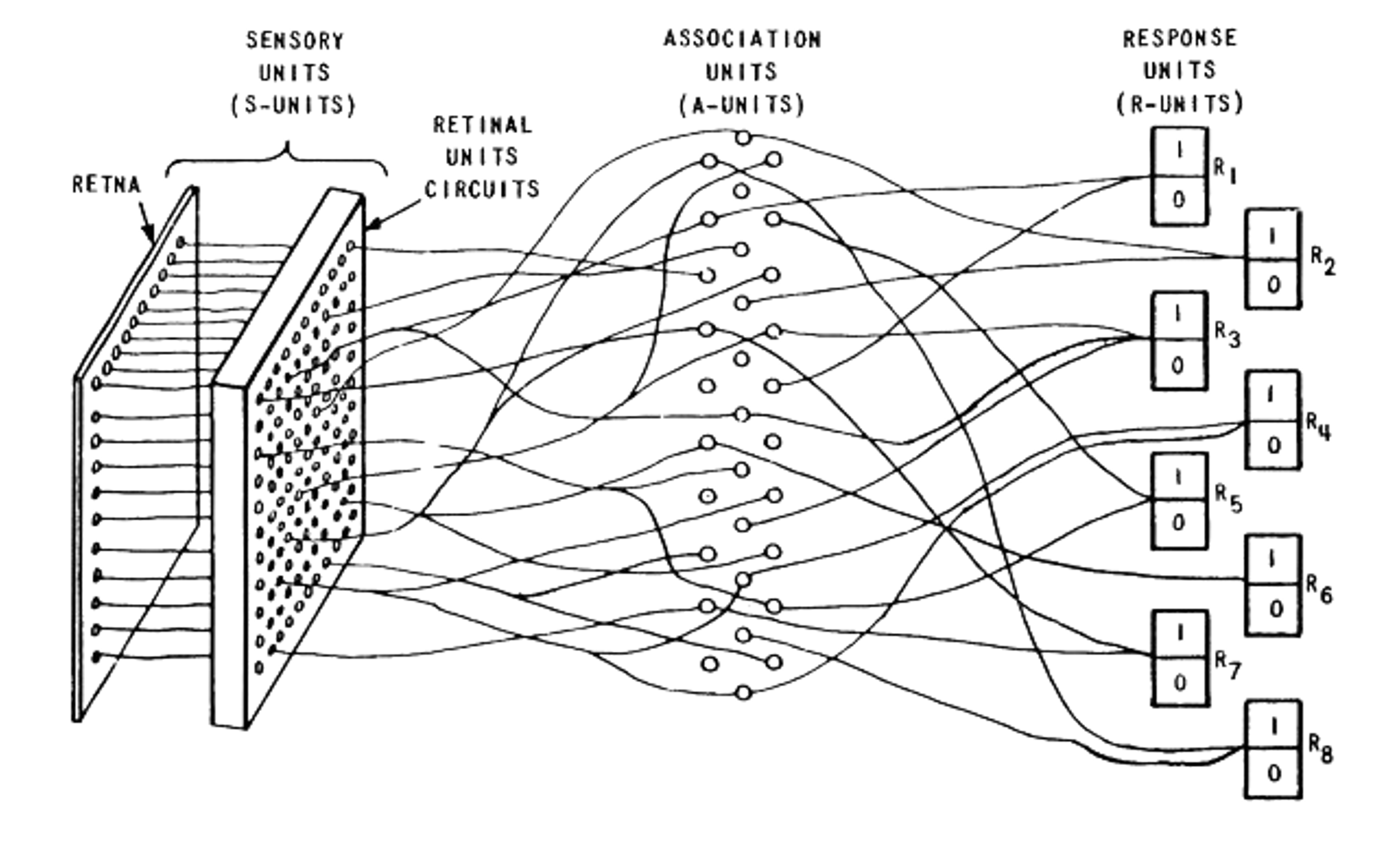}} are powered by Deep Learning, a class of algorithms whose history can be traced back to attempts in the early 20th century to replicate the connectivity and functioning of biological neurons in the brain in computers in a very abstract manner. Such systems are called (artificial) neural networks, by analogy to their biological counterparts, and consist of computational units called neurons, which are typically organized into multiple layers (the term `deep' in Deep Learning refers to neural networks with many such layers). Neurons have parameters that can be tuned for a specific task in an optimization procedure referred to as `learning'. A subfield of AI studying mathematical methods for the design and optimization of such systems is called Machine Learning~(ML).

\begin{tcolorbox}[colback=gray!10, colframe=gray!40]
\textit{Deep Learning} is an umbrella term for Machine Learning algorithms that rely on artificial neural networks typically consisting of a large number of layers. 
\end{tcolorbox}

\clearpage

\subsection*{What is Geometric Deep Learning?}\marginnote{In the early 2020s there has been a clear convergence towards Transformer-based architectures across data modalities~\cite{vaswani2017attention}.} 

In recent years, there has been a rapid proliferation of various artificial neural network architectures, each suggesting different connectivity patterns and internal computations to be performed by the learning systems. 

Geometric Deep Learning is a subfield of Deep Learning~\cite{lecun2015deep,Goodfellow-et-al-2016} that focuses on developing artificial neural networks for data with non-Euclidean structures, such as graphs and manifolds. Traditional deep learning models, operate on grid-like data (e.g., images, time series, text), but many real-world problems involve more complex, irregular geometries. In particular, the field focuses on analyzing neural networks based on the geometric priors they leverage. Different models combat the curse of dimensionality by modeling signals on domains endowed with symmetry groups, which serve as inductive biases for the network. \marginnote{Imposing inductive biases in learning systems becomes particularly important in data-scarce regimes. While modern Deep Learning is only loosely rooted in biological neural networks, some architectural choices, such as the inductive biases of Convolutional Neural Networks (CNNs), are directly inspired by the workings of the visual cortex.

\includegraphics[width=1\linewidth]{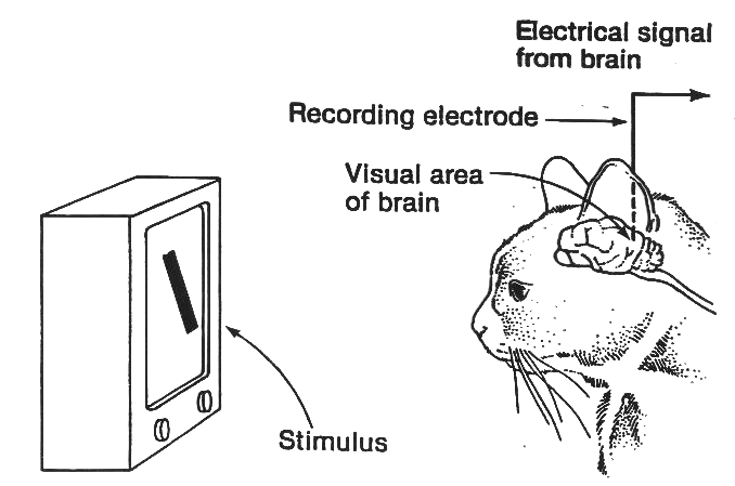}

This inspiration can be traced back to the experiments of Hubel and Wiesel.}  

\begin{tcolorbox}[colback=orange!20, colframe=orange!60]
\textit{Geometric Deep Learning} provides a structured approach to incorporating prior knowledge of physical symmetries into the design of new neural network architectures, while also unifying and understanding successful existing models under a common framework.
\end{tcolorbox}

\begin{figure}[hbpt!]
    \centering  \includegraphics[width=0.7\linewidth]{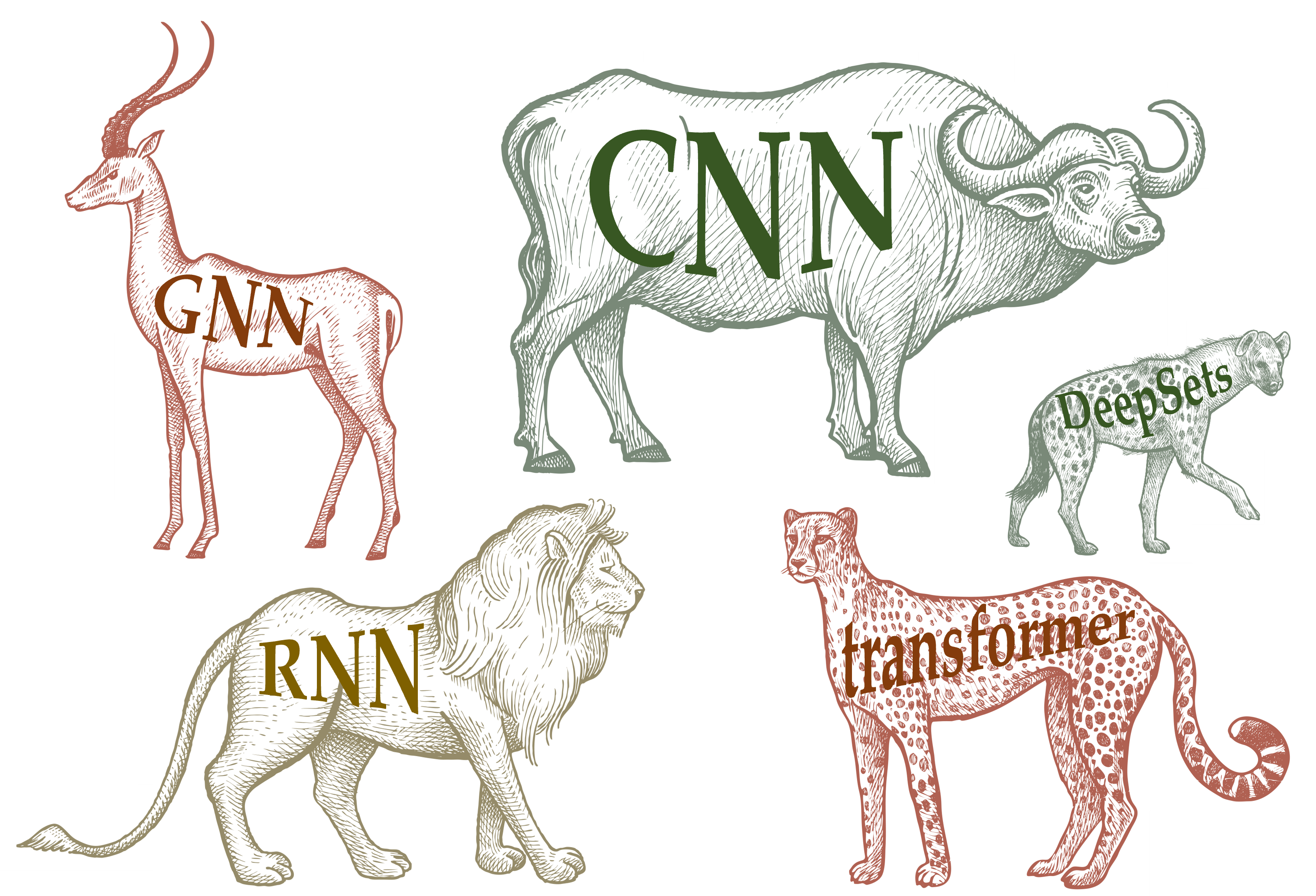}
    \caption{In the spirit of the Erlangen Program, Geometric Deep Learning provides a geometric unification of the zoo of Deep Learning architectures.}
    \label{fig:nn_zoo}
\end{figure}\marginnote{In his landmark work known as the Erlangen Program, Felix Klein proposed that geometry should be approached as the study of invariants or symmetries. His vision offered a unifying framework at a time when the development of various non-Euclidean geometries had led to a fragmented mathematical landscape. Geometric Deep Learning adopts a similar perspective for understanding artificial neural network architectures by analyzing the symmetries and invariances they exploit.

\includegraphics[width=\linewidth]{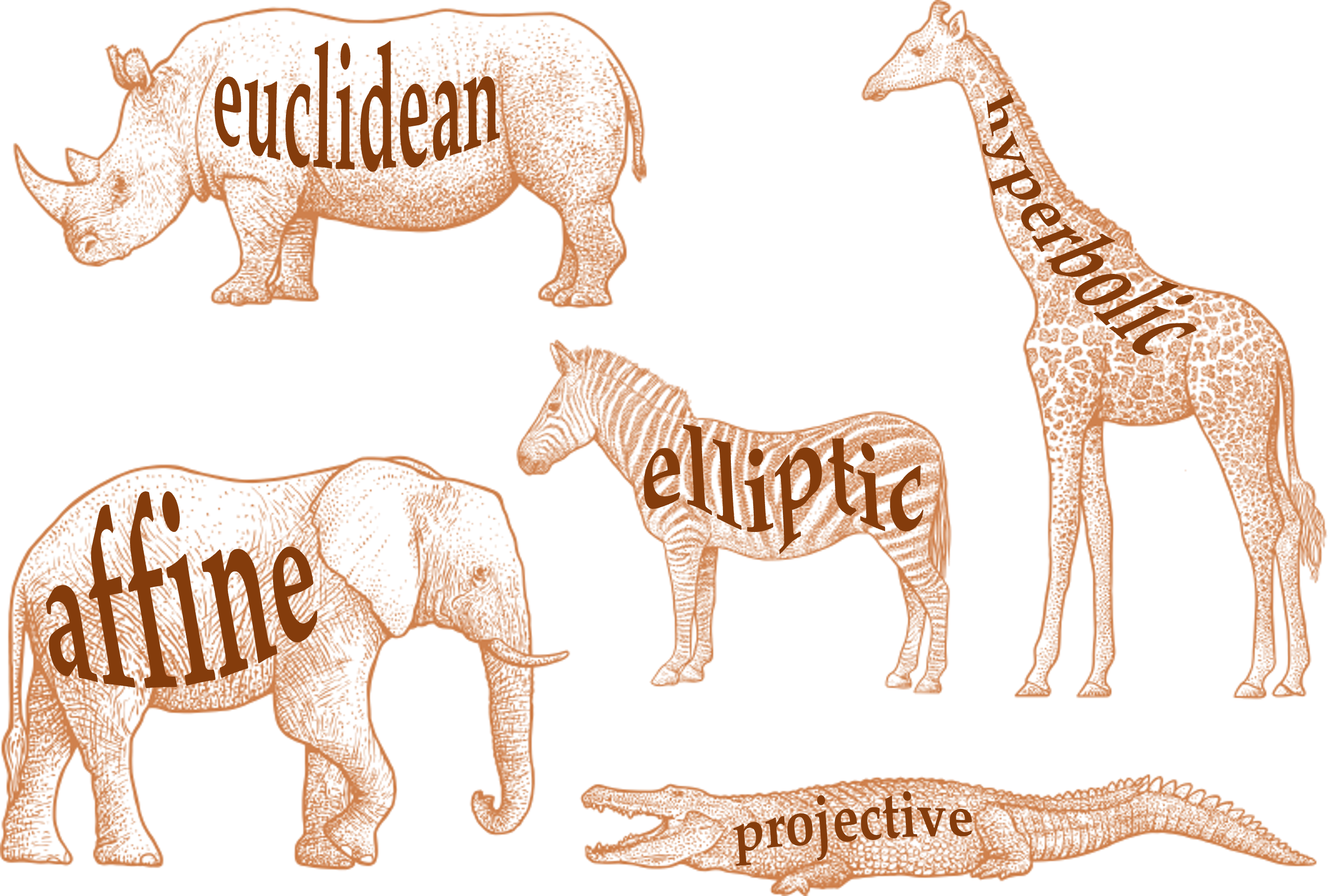}}

In this text, we will not focus directly on (Geometric) Deep Learning and artificial neural networks. Instead, our objective is to provide the necessary preliminary mathematical background often overlooked in standard computer science curricula.

\pagebreak

\tableofcontents

\pagebreak

\section{Algebraic Structures and Mathematics before Numbers}
\label{Algebraic Structures and Mathematics before Numbers}

In this section, we study pre-numerical structures that are fundamental to understanding Geometric Deep Learning. Structures such as sets and maps allow us to mathematically describe collections of objects, the connections between them, and the operations that can be performed on them~\cite{pinter2014bookofsettheory}. A key focus is on groups, which are used in Geometric Deep Learning to model the transformations of the data. 

\greylinelong

\subsection{Sets, Maps, and Functions}

At first glance, numbers may appear as the most elementary objects in mathematics. 
However, it is possible to identify even simpler and more basic structures. 
Indeed, numbers can be added, subtracted, multiplied, and so on, which requires a set of rules defining how these operations are done. But what if we consider just a collection of objects, stripped off any additional assumptions about them?

\begin{tcolorbox}[colback=gray!10, colframe=gray!40]
A {\em set} is a collection of distinct objects, called {\em elements} or {\em members} of the set. \marginnote{The elements of a set are not restricted to being numbers; they could also be English words, for instance: $\{\textrm{cat},\,\textrm{dog}\}$.}
\end{tcolorbox}

These elements can be anything: numbers, symbols, or even other sets. What characterizes a set is that it does not allow for a repetition of elements (i.e., every element appears only once in a set)\marginnote{A {\em multiset} is a set in which elements are allowed to appear more than once. Multisets are common in Geometric Deep Learning in the context of Graph Neural Networks (GNNs)~\cite{gnn_scarselli}, where they are used to model the neighborhood of a node in the graph.}, and the order in which elements appear does not matter (i.e., sets are {\em unordered}). Sets are the basis for defining more complex mathematical structures.

They are typically denoted by capital letters, such as $ A $, $ B $, $ X $, etc. The members of a set are listed inside curly braces $ \{\} $, and if an element $ x $ belongs to a set $ A $, we write $x \in A$, which reads as `$x$ is an element of $A$'. If $ x $ does not belong to $ A $, we write $ x \notin A$. For instance, if $ A = \{1, 2, 3\}, $ then  
$ 2 \in A$, but 
$4 \notin A$.

\greyline

\paragraph*{Examples of Sets} \marginnote{A non-example would be the collection of all sets: there is no set containing all sets.}
\begin{itemize}
    \item $ \emptyset $: The empty set, a set with no elements. It is denoted by $ \emptyset $ or sometimes by $ \{\} $.
    \item {\em Singleton Set}: A set with exactly one element, for example, $ \{1\} $.
    \item $ \mathbb{N} = \{1, 2, 3, \ldots\} $: The set of natural numbers. The ellipsis $ \ldots $ indicates that the set continues indefinitely with positive integers. \marginnote{In some textbooks $\mathbb{N}$ may include 0.}
    \item $ \mathbb{Z} = \{\ldots, -3, -2, -1, 0, 1, 2, 3, \ldots\} $:\marginnote{The notation $\mathbb{Z}$ for integers comes from the German {\em Zahlen}, which means `numbers'.} The set of integers, which includes positive numbers, negative numbers, and zero.
    \item $ \mathbb{Q} = \left\{ \frac{p}{q} \mid p \in \mathbb{Z}, q \in \mathbb{N} \right\} $: The set of rational numbers, which are numbers that can be expressed as a ratio of two integers.
    \item $ \mathbb{R} $: The set of all real numbers, including both rational numbers (e.g., $ 1, 0.75, -3 $) and irrational numbers (e.g., $ \pi, \sqrt{2} $).
    \item $ \mathbb{C} $: The set of all complex numbers, which can be written as $ a + bi $, where $ a $ and $ b $ are real numbers and $ i $ is the imaginary unit with $ i^2 = -1 $.
\end{itemize}

\greyline

\paragraph*{Set Notation and Operations}

\begin{itemize}

\item {\em Set Builder Notation}: Set builder notation is used to describe a set by specifying an expression or the general form of an element, followed by a vertical bar separator $|$\marginnote{Sometimes a colon is used instead of a vertical line:$$\{ x : \mathfrak{f}(x) \}.$$}, and, to its right, a rule that the expression on the left must satisfy.

$$
\{ x | \mathfrak{f}(x) \} = \{ \textrm{expression} | \textrm{rule satisfied by the expression} \}.
$$

In words, it can be read as `$x$ such that (for which) $\mathfrak{f}(x)$'.
    
\item {\em Subset}: A set $ A $ is a subset of a set $ B $, written $ A \subseteq B $, if every element of $ A $ is also an element of $ B $. If $ A \subseteq B $ but $ A \neq B $, we say that $ A $ is a proper subset, written $ A \subset B $.
\item  {\em Union}: The union of two sets $ A $ and $ B $, written $ A \cup B $, is the set of all elements that are in $ A $, in $ B $, or in both.
\item  {\em Intersection}: The intersection of two sets $ A $ and $ B $, written $ A \cap B $, is the set of all elements that are in both $ A $ and $ B $.
\item  {\em Difference}: The difference of two sets $ A $ and $ B $, written $ A \setminus B $, is the set of all elements that are in $ A $ but not in $ B $.
\item  {\em Complement}: The complement of a set $ A $, written $ A^c $, is the set of all elements not in $ A $, assuming a universal set $ U $ that contains all elements under consideration.
\item {\em Power Set}: The power set of a set $ A $, denoted $ \mathcal{P}(A) $, is the set of all subsets of $ A $, including the empty set and $ A $ itself.
\item {\em Cardinality}: \marginnote{The cardinality of $\mathbb{N}$ is denoted by the Hebrew letter $\aleph_0$, which reads as \textit{aleph-nought} or \textit{aleph-zero}. This is the `smallest' type of infinity and represents the size of any countable infinite set, which is a set that can be placed in a one-to-one correspondence (bijection) with $\mathbb{N}$. 
For example, even though they might appear `larger' at first glance, the sets $\mathbb{Z}$ and $\mathbb{Q}$ also have cardinality $\aleph_0$ since they are countably infinite.} The cardinality of a set is the size or number of elements it contains. If a set is finite, its cardinality is a non-negative integer. For infinite sets, cardinality is definited more abstractly: two infinite sets are said to have the same cardinality if there exists a bijection between their elements. The cardinality of a set $A$ is denoted by $|A|$ or sometimes $\#(A).$
\end{itemize}

\greyline

\paragraph{Examples of Set Builder Notation} We provide some examples to build an intuitive understanding. We start with the set builder notation. Below, we show that there are multiple ways to specify a set containing natural even numbers: $$\{ 2x | x \in \mathbb{N} \} = \{ x \in \mathbb{N}| x\,\textrm{is even} \} = \{2,4,6,8,... \}. $$
Alternatively, sometimes the rule that must be satisfied by the elements of the set could be an equation:$$\{x\in\mathbb{Z}| x>0\}=\mathbb{N},$$
$$\{x\in\mathbb{Q}| x^2=2\}=\emptyset.$$

In the last example, the solutions to the equation $x^2=2$ are the roots $x=\pm\sqrt{2}$, which are irrational numbers and, therefore, not elements of $\mathbb{Q}$. Thus, the rule has no satisfying elements, meaning we have found a convoluted way of describing the empty set.

\greyline

\paragraph{Examples of Finite Sets and Simple Operations} Next, let us consider the finite sets $ B = \{1, 2, 3, 4, 5\}, $  $ A = \{1, 2, 3\}, $ $ C = \{1, 2, 3, 4, 5\}, $ then 
$ C \subseteq B$ and $ A \subset B$. 
This is because $A \neq B$, whereas $C = B$. Their cardinalities would be $|A| = 3$, $|B| = 5$, and  $|C|=5$. 
The unions and intersections in this example are 
$ C \cup B = C \cap B = C = B$, 
$ A \cup B = B$, and 
$ A \cap B = A$. Another interesting example is the cardinality of the empty set $|\emptyset| =0$ and the cardinality of the singleton set containing the empty set $|\{\emptyset\}| =1.$

\greyline

\paragraph{Examples of Infinite Sets and Simple Operations} Consider the infinite sets $\mathbb{N} = \{1, 2, 3, 4, 5, \dots\}$ and $\mathbb{E} = \{2, 4, 6, 8, \dots\}$, the set of natural numbers and even natural numbers, respectively. Unsurprisingly, $\mathbb{E} \subset \mathbb{N}$ since every element of $\mathbb{E}$ is an element of $\mathbb{N}$. However, unlike finite sets, the cardinalities of $\mathbb{N}$ and $\mathbb{E}$ are \textit{equal}, denoted as $|\mathbb{N}| = |\mathbb{E}| = \aleph_0$. \marginnote{The Hilbert Hotel with infinitely many rooms that are fully occupied can host an infinite number of new guests by moving the old ones into even-numbered rooms and placing the new ones into odd-numbered rooms.

If there exists a one-to-one correspondence between two infinite sets, although we cannot say that they have the same number of elements, we think of them as having the ``same size''. This intuition is formalized in set theory by defining two sets $A$ and $B$ to be \textit{equipotent} (or having the \textit{same power}), if there is a one-to-one correspondence from $A$ to $B$.} This is due to the fact that there exists a \textit{bijection} between $\mathbb{N}$ and $\mathbb{E}$ (we will explain bijections in more detail soon). One such bijection $f : \mathbb{N} \to \mathbb{E}$ can be defined as $f(n) = 2n$. For every natural number $n \in \mathbb{N}$, $f(n)$ produces a unique element of $\mathbb{E}$, and every element of $\mathbb{E}$ is hit exactly once. For example: $f(1) = 2, \, f(2) = 4, \, f(3) = 6, \, \dots$ Hence, despite $\mathbb{E}$ being a proper subset of $\mathbb{N}$, their infinite cardinality remains the same. 

In terms of other operations: $\mathbb{E} \cup \mathbb{N} = \mathbb{N},\,\mathbb{E} \cap \mathbb{N} = \mathbb{E}$ and $ \mathbb{N} \setminus \mathbb{E} = \{1,3,5,7,\dots\}$. Notably, the cardinality of the set containing, for instance, the infinite sets $\mathbb{R}$ and $\mathbb{N}$ is actually $|\{\mathbb{R},\mathbb{N}\}|=2$, since the set only contains two elements, despite the elements themselves being infinite.

\begin{tcolorbox}[colback=orange!20, colframe=orange!60]
\textbf{Sets in Geometric Deep Learning and Graph Neural Networks.}
In Geometric Deep Learning, we are often interested in modeling signals on collections of nodes, edges, and patches on a manifold, for instance. As we will see later in Section~\ref{Graph Theory}, in the context of GNNs, the geometric domain is defined as a graph \( G = (V, E) \), which is a tuple consisting of a set of nodes \( V \) and a set of edges \( E \). Similarly, to model the neighborhood of a node, multisets (sets that allow repetition of elements) are used.
\end{tcolorbox}

\greyline

\paragraph*{Cartesian Products} After introducing sets and some basic operations, let us define the Cartesian product. Although the concept may initially seem abstract, it plays an important role in discussing manifolds and constructing more complex spaces by combining elements from simpler subspaces.\marginnote{The term \textit{Cartesian product} comes from the Cartesian coordinate system, which in turn is named after the French philosopher and scientist René Descartes. Descartes’s name was Latinized to Renatus Cartesius, hence the adjective \textit{Cartesian}.} The Cartesian product is used to model composite systems and relations between elements of two or more sets. 

\begin{tcolorbox}[colback=gray!10, colframe=gray!40]
The \textit{Cartesian product} of two sets $A$ and $B$, denoted by $A \times B$, is the set of all ordered pairs $(a, b)$ where $a \in A$ and $b \in B$:

$$
A \times B = \{ (a, b) \mid a \in A, b \in B \}.
$$
\end{tcolorbox}

For instance, let $A = \{1, 2\}$ and $B = \{b_1, b_2\}$. Their product $A \times B$ is:
\[
A \times B = \{(1, b_1), (1, b_2), (2, b_1), (2, b_2)\}.
\]

\clearpage

We can also represent it as a table:\marginnote{As we will see in Section~\ref{sec:Manifolds and Differential Geometry}, one application of the Cartesian product is to represent complex manifolds as combinations of simpler ones. For instance, by taking the Cartesian product of multiple 1-spheres (circles), we can define points on a hypertorus. In Geometric Deep Learning, this approach can encode data into complex latent spaces while maintaining a closed-form differentiable representation of the underlying geometry.
\vspace{20pt}
\includegraphics[width=1\linewidth]{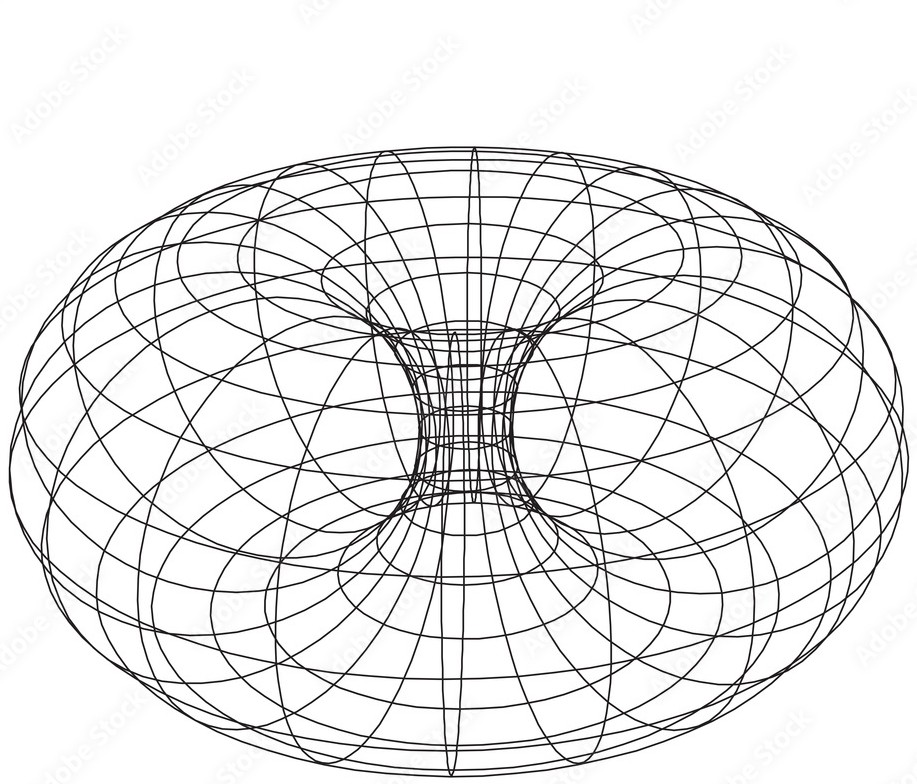}
}

\[
\begin{array}{c|c|c}
    A \times B & b_1 & b_2 \\
    \hline
    1 & (1, b_1) & (1, b_2) \\
    2 & (2, b_1) & (2, b_2) \\
\end{array}
\]

\greyline

\paragraph*{Maps} In many curricula, students are directly introduced to functions. However, before discussing functions, we can explore the more general concept of rules that define \textit{mappings} between elements of different sets.

\begin{tcolorbox}[colback=gray!10, colframe=gray!40]
A {\em map} is a rule $F$ which assigns to each element of a set $A$ another element of a set $B$: $F(a) \equiv b \in B \ \forall a \in A.$
\end{tcolorbox}

In the above expression, we read \(\equiv\) as `is defined as' or `is equivalent to', indicating that \(F(a)\) is explicitly assigned the value \(b\) in the set \(B.\) The symbol \(\forall\) is read as `for all', emphasizing that this rule applies to every element \(a\) in the set \(A.\)  

It is common to use the following notation $F: A \rightarrow B.$ We call $A$ the \textit{domain} and $B$ the \textit{codomain}, the element $a \in A$ fed into the map is the \textit{argument} (or \textit{preimage}), and $F(a)$ its \textit{image}. Note that we use different notations to distinguish a mapping between sets and its behavior on individual elements. For example:

$$
F: \mathbb{N} \rightarrow \mathbb{Z}, \quad x \mapsto F(x) = x^2,
$$

where the expression on the left-hand side focuses on specifying the domain and codomain of $F$, whereas the right-hand side highlights the action of $F$ on individual elements of the domain, that is, on particular inputs.

\begin{tcolorbox}[colback=gray!10, colframe=gray!40]
A \textit{function} is a special type of mapping, which maps a set into the set of numbers.
\end{tcolorbox}

\greyline

\paragraph{Types of Maps} Maps can be \textit{surjective},\marginnote{The terms injection, surjection, and bijection were introduced by a group of French mathematicians publishing under the collective pseudonym Nicholas Bourbaki in 1954, and the adjective forms first used by Claude Chevalley in 1956.
\includegraphics[width=1\linewidth]{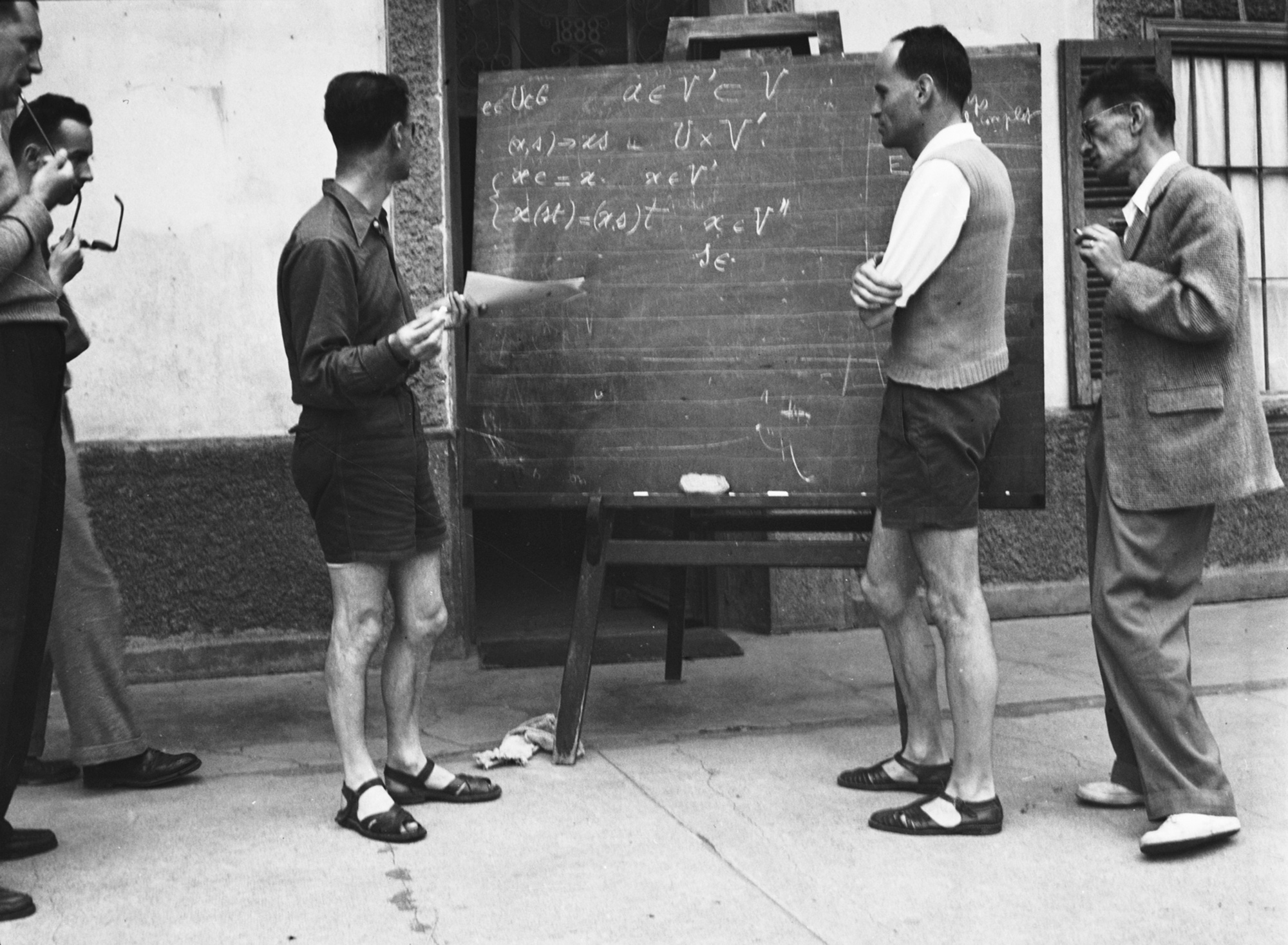}} \textit{injective}, or \textit{bijective}, depending on how they map elements from one set to another. We say that a map between two sets is \textit{bijective} when it is both \textit{injective} and \textit{surjective}.
\vspace{1cm}

\begin{figure}[htbp]
    \centering
    \begin{minipage}{0.3\textwidth}
        \centering
        \begin{tikzpicture}[scale=0.6]
    \tikzset{
        set/.style={draw,circle,minimum size=1.5cm},
        mapping/.style={->,>=stealth,thick}
    }
    \tikzset{
        dotA/.style={circle,fill=blue,fill opacity=0.3,inner sep=2pt},  
        dotB/.style={circle,fill=red,fill opacity=0.3,inner sep=2pt}    
    }

    \node[set] (A1) at (0,0) {};
    \node[dotA] (a1) at (-0.5,0.5) {};
    \node[dotA] (a2) at (-0.5,-0.5) {};

    \node[set] (B1) at (3,0) {};
    \node[dotB] (b1) at (2.5,0.8) {};
    \node[dotB] (b2) at (2.5,0) {};
    \node[dotB] (b3) at (2.5,-0.8) {};
    
    \draw[mapping] (a1) -- (b1);
    \draw[mapping] (a2) -- (b2);
\end{tikzpicture}
        \caption*{(a) Injective}
    \end{minipage}%
    \hfill
    \begin{minipage}{0.3\textwidth}
        \centering
        \begin{tikzpicture}[scale=0.6]
    \tikzset{
        set/.style={draw,circle,minimum size=1.5cm},
        mapping/.style={->,>=stealth,thick}
    }

    \tikzset{
        dotA/.style={circle,fill=blue,fill opacity=0.3,inner sep=2pt},  
        dotB/.style={circle,fill=red,fill opacity=0.3,inner sep=2pt}    
    }

    \node[set] (A2) at (0,0) {};
    \node[dotA] (a3) at (-0.5,0.8) {};
    \node[dotA] (a4) at (-0.5,0) {};
    \node[dotA] (a5) at (-0.5,-0.8) {};

    \node[set] (B2) at (3,0) {};
    \node[dotB] (b4) at (2.5,0.5) {};
    \node[dotB] (b5) at (2.5,-0.5) {};
    
    \draw[mapping] (a3) -- (b4);
    \draw[mapping] (a4) -- (b4);
    \draw[mapping] (a5) -- (b5);
\end{tikzpicture}
        \caption*{(b) Surjective}
    \end{minipage}%
    \hfill
    \begin{minipage}{0.3\textwidth}
        \centering
        \begin{tikzpicture}[scale=0.25]
    \tikzset{
        set/.style={draw,circle,minimum size=1.5cm},
        mapping/.style={->,>=stealth,thick}
    }

    \tikzset{
        dotA/.style={circle,fill=blue,fill opacity=0.3,inner sep=2pt},  
        dotB/.style={circle,fill=red,fill opacity=0.3,inner sep=2pt}   
    }

    \node[set] (A3) at (-3.5,0) {};
    \node[dotA] (a6) at (-3.5,0.8) {};
    \node[dotA] (a7) at (-3.5,0) {};
    \node[dotA] (a8) at (-3.5,-0.8) {};

    \node[set] (B3) at (3.5,0) {};
    \node[dotB] (b6) at (3.5,0.8) {};
    \node[dotB] (b7) at (3.5,0) {};
    \node[dotB] (b8) at (3.5,-0.8) {};
    
    \draw[mapping] (a6) -- (b6);
    \draw[mapping] (a7) -- (b7);
    \draw[mapping] (a8) -- (b8);
\end{tikzpicture}
        \caption*{(c) Bijective}
    \end{minipage}
    \caption{Depiction of injective, surjective, and bijective maps between two sets whose elements are highlighted in blue and red respectively.}
    \label{fig:functions}
\end{figure}

\begin{tcolorbox}[colback=gray!10, colframe=gray!40]
\textit{Injective (One-to-One):}\marginnote{There are other alternative ways of expressing the injectivity property: If $(a_1,b)\in F$ and $(a_2,b)\in F$, then $a_1=a_2$, or $b$ has no more than one pre-image.}
A map $ F: A \to B $ is called {\em injective} (or one-to-one) if different elements in the domain $ A $ map to different elements in the codomain $ B $. That is, for all $ a_1, a_2 \in A $,
$$
F(a_1) = F(a_2) \implies a_1 = a_2.
$$
\end{tcolorbox}
\begin{tcolorbox}[colback=gray!10, colframe=gray!40]
\textit{Surjective (Onto):}\marginnote{$F: A \to B$ is surjective if and only if $\textrm{ran} F = B$, where $\textrm{ran} F = \{ b : \exists a \ni (a,b) \in F \}.$} A map $ F: A \to B $ is called {\em surjective} (or onto) if every element in the codomain $ B $ has at least one preimage in the domain $ A $. That is, for every $ b \in B $, there exists an $ a \in A $ such that
$$
F(a) = b.
$$
\end{tcolorbox}
\begin{tcolorbox}[colback=gray!10, colframe=gray!40]
\textit{Bijective:}
A map $ F: A \to B $ is {\em bijective} if it is both injective and surjective. In other words, each element of $ A $ maps to a unique element of $ B $, and every element of $ B $ has a unique preimage in $ A $. A bijective map has an inverse, denoted $ F^{-1}: B \to A $, such that
$$
F^{-1}(F(a)) = a \quad \forall \quad a \in A, \quad F(F^{-1}(b)) = b \quad \forall \quad b \in B.
$$
\end{tcolorbox}

\greyline

\paragraph{Composition} Maps between different sets can be combined.

\begin{tcolorbox}[colback=gray!10, colframe=gray!40]
Given two maps, \( F_1: A \to B \), and  
\( F_2: B \to C \), the \textit{composition} of \( F_1 \) and \( F_2 \), denoted as \( F_2 \circ F_1 \), is a new map:
\[
F_2 \circ F_1: A \to C.\]
\end{tcolorbox}

Note that when we compose injective maps, the result is also injective. Similarly, when we compose surjective maps or two bijective maps, the resulting maps are also surjective and bijective, respectively.

Like maps\marginnote{Note that composition of functions is associative but not commutative.}, functions can also be composed to create new functions. If \( f: X \to Y \) and \( g: Y \to Z \), their composition, denoted as \( g \circ f \), is a function \( g \circ f: X \to Z \) defined by:

\[
(g \circ f)(x) = g(f(x)).
\]  

For example, let \( f(x) = x^2 \) and \( g(x) = \sin(x) \). Then the composition \( g \circ f \) is:  
\[
(g \circ f)(x) = g(f(x)) = \sin(f(x)) = \sin(x^2).
\]  

Similarly, the reverse composition \( f \circ g \) is:  
\[
(f \circ g)(x) = f(g(x)) = f(\sin(x)) = (\sin(x))^2.
\]  

\begin{tcolorbox}[colback=orange!20, colframe=orange!60]
\textbf{Function Composition and Deep Learning.}
Arguably, the foundation of Deep Learning lies in function composition, where the input undergoes iterative transformations through successive layers. Each layer processes the output (or activations) of the previous one, passing it as input to the next layer in the neural network. Also note that artificial neural networks are generally not bijective, as they are neither guaranteed to be injective nor surjective.
\end{tcolorbox}

\begin{figure}[hbpt!]
    \centering  \includegraphics[width=\linewidth]{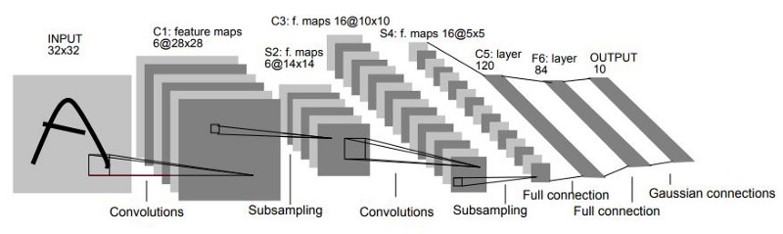}
    \caption{LeNet-5 classical CNN architecture.}
    \label{fig:letnet-5}
\end{figure}

For instance,\marginnote{It is possible to visualize the internal filters learned by deep CNNs. The filters in the initial layers typically capture primitive patterns such as edges, corners, and textures, while the filters in deeper layers learn to compose these primitives into more complex features.

\includegraphics[width=1\linewidth]{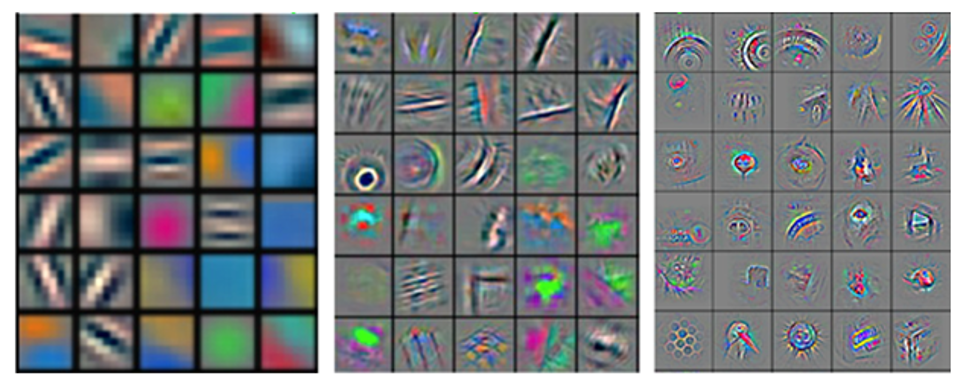}

} Figure~\ref{fig:letnet-5} displays a schematic of a LeNet-5 neural network. We can observe how the input image is processed from left to right. The feature maps (yet another term for layer outputs or activations) are processed by different layers in the architecture and passed as input to the next layer to produce the subsequent set of feature maps. This is an example of function composition.

\greyline
\paragraph{Hypothesis Class} In machine learning it is common to come across the concept of \textit{hypothesis class}.\marginnote{In Machine Learning we want to exploit the underlying low-dimensional structure of the input high
dimensional space \(\mathcal X\). We can expect three sources of error in high-dimensional learning: approximation error, statistical error, and optimization error.}

\begin{tcolorbox}[colback=gray!10, colframe=gray!40]
If \(\mathcal X\) is the input space and \(\mathcal Y\) the label (or output) space, then a \textit{hypothesis class} is any set  
\[
\mathcal F \;\subseteq\; \{\,f:\mathcal X\to\mathcal Y\}
\]  
of functions (hypotheses) from \(\mathcal X\) to \(\mathcal Y\) from which a learning algorithm chooses its prediction rule.
\end{tcolorbox}

For instance, in linear regression the hypothesis class is the set of all possible lines. For the multivariate linear regression case, we have: 
\[
\mathcal F_{\text{lin}} = \bigl\{\,f_{w,b} : \mathbb R^d \to \mathbb R \;\big|\; h_{w,b}(x) = w^\top x + b,\; w\in\mathbb R^d,\;b\in\mathbb R \bigr\},
\]  
i.e. the set of all affine (straight‐line) functions parameterized by \((w,b)\).

\marginnote{An MLP is one of the first neural network architectures. It consists of stacking multiple `perceptrons', which take a multidimensional input, assign a weight to each of its entries, add the results, and apply a non-linear transformation.}In Deep Learning, the hypothesis class is given by the neural network architecture construction we choose to implement. The model then learns to optimize the parameters via gradient descent (Section~\ref{subsec:GradientDescent}) and converges on a particular function given the data used to train it. For example, in the case of a MultiLayer Perceptron (MLP), the hypothesis class would be 
  
\[
\begin{aligned}
\mathcal F_{\rm NN}
&=\;\bigl\{\,f_\theta : \mathcal X \to \mathcal Y \mid \theta\in\Theta\,\},\\
f_\theta(x)
&=\;\sigma_L\Bigl(W^{(L)}\bigl(\cdots\sigma_2\bigl(W^{(2)}(\sigma_1(W^{(1)}x + b^{(1)})) + b^{(2)}\bigr)\cdots\bigr) + b^{(L)}\Bigr).
\end{aligned}
\]

where $L$ is the total number of layers, $\theta = \bigl\{W^{(1)},b^{(1)},\,W^{(2)},b^{(2)},\dots,W^{(L)},b^{(L)}\bigr\}$ is the set of learnable parameters,\marginnote{In the past sigmoid functions were a standard activation function for hidden neural network layers. However, due to the so-called `vanishing gradient problem', sigmoids are currently mainly used as a final non-linear transformation for binary classification problems. Rectified Linear Units (ReLUs) and its variants such as Exponential Linear Units (ELUs) and Leaky ReLUs are a more standard choice in the literature nowadays. For large scale Transformers the Sigmoid Linear Unit (SiLU) (also known as the swish function) is widely used instead. Many other activations functions have been proposed in the literature.} \(\Theta\) is typically \(\mathbb{R}^{\,\sum_i(\dim W^{(i)}+\dim b^{(i)})}\), and $\sigma_1, ... \sigma_L$ are non-linear activation functions. In other words, the MLP architecture is a composition of affine transformations and non-linear functions.

\begin{tcolorbox}[colback=orange!20, colframe=orange!60]
\textbf{Restricting the Hypothesis Class using Symmetries.}
The larger the hypothesis class, the better the best hypothesis models the underlying true function, but the harder it is to find that best hypothesis. In Geometric Deep Learning we often choose to restrict our neural network hypothesis class by embedding symmetry (invariance and equivariant) into our layer transformations. This can lead to more efficient learning in data scarce regimes. This is related to the \textit{bias-variance tradeoff} often mentioned in the literature.
\end{tcolorbox}

\greylinelong
\subsection{Groups}
\label{subsec:Groups}

A group is a way of organizing and understanding how a set of elements interact with one another through a well-defined operation. Groups are used to describe symmetry, structure, and transformations in various mathematical and physical contexts. 

Let us consider a physical example before diving into the formal definition. Think of a square and the \textit{group of rotations of the square}. The set of elements in this group consists of the different rotations $C_4 = \{0^\circ, 90^\circ, 180^\circ, 270^\circ\}$ that can be applied to the square. The operation here is combining rotations. For instance, applying two $90^\circ$ rotations is equivalent to a single $180^\circ$ rotation. Applying a $0^\circ$ rotation followed by a $90^\circ$ rotation results in just a $90^\circ$ rotation. This shows that combining elements of the set results in elements within the same set.\marginnote{Many classes of physical operations can be associated with a group structure. Since Geometric Deep Learning architectures often aim to model such phenomena, groups become essential for designing artificial neural networks whose internal representations align with physical principles.}

\vspace{1cm}

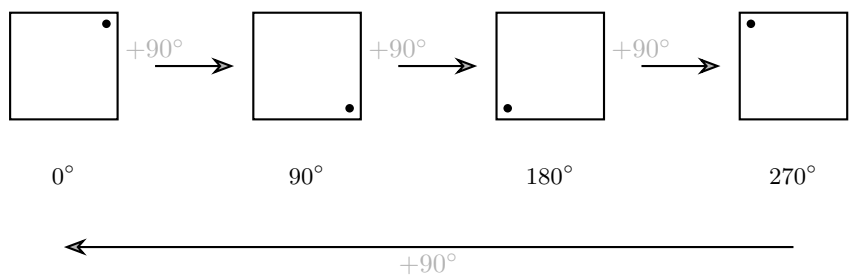
\begin{figure}[hbpt!]
\centering
\begin{tikzpicture}[scale=0.8]
    \definecolor{myblue}{rgb}{0.2, 0.4, 0.8}

    \tikzset{
        dotA/.style={circle,fill=myblue,fill opacity=0.3,inner sep=2pt},  
        dotB/.style={circle,fill=red,fill opacity=0.3,inner sep=2pt},    
        square/.style={regular polygon,regular polygon sides=4,minimum size=2cm,draw,thick},
        arrow/.style={-{Stealth[length=3mm]},thick,fill=myblue,fill opacity=0.3,inner sep=2pt},  
        label/.style={font=\small}
    }
    
    \begin{scope}[xshift=0cm]
        \node[square] (s0) {};
        \node[label] at (0,-1.8) {$0^\circ$};
        \fill (0.7,0.7) circle (2pt);
        \node at (0.9,0.9) {};
    \end{scope}
    
    \begin{scope}[xshift=4cm]
        \node[square,rotate=90] (s90) {};
        \node[label] at (0,-1.8) {$90^\circ$};
        \fill (0.7,-0.7) circle (2pt);
        \node at (0.9,-0.9) {};
    \end{scope}
    
    \begin{scope}[xshift=8cm]
        \node[square,rotate=180] (s180) {};
        \node[label] at (0,-1.8) {$180^\circ$};
        \fill (-0.7,-0.7) circle (2pt);
        \node at (-0.9,-0.9) {};
    \end{scope}
    
    \begin{scope}[xshift=12cm]
        \node[square,rotate=270] (s270) {};
        \node[label] at (0,-1.8) {$270^\circ$};
        \fill (-0.7,0.7) circle (2pt);
        \node at (-0.9,0.9) {};
    \end{scope}
    
    \draw[arrow] (1.5,0) arc (0:0:1cm) node[midway,above] {$+90^\circ$} -- (2.8,0);
    \draw[arrow] (5.5,0) arc (0:0:1cm) node[midway,above] {$+90^\circ$} -- (6.8,0);
    \draw[arrow] (9.5,0) arc (0:0:1cm) node[midway,above] {$+90^\circ$} -- (10.8,0);
    
    \draw[arrow] (12,-3) .. controls (6,-3) and (6,-3) .. (0,-3) 
        node[midway,below] {$+90^\circ$};
\end{tikzpicture}
    \label{fig:square_rotation}
    \caption{Rotational Symmetries of a Square ($C_4$).}
\end{figure}

This situation exemplifies\marginnote{The term \textit{symmetry} has Greek origins `symmetria' literally translates to `same measure'.} symmetry: the square remains unchanged (\textit{invariant}) under these rotations. In mathematics, symmetry refers to a property of an object or system that remains unchanged under specific transformations or operations. 

Similar schematics can be created, for instance, to represent the symmetry of a triangle under both rotations and reflections. More generally, we refer to these as Cayley graphs.

\begin{figure}[hbpt!]
    \centering  \includegraphics[width=0.7\linewidth]{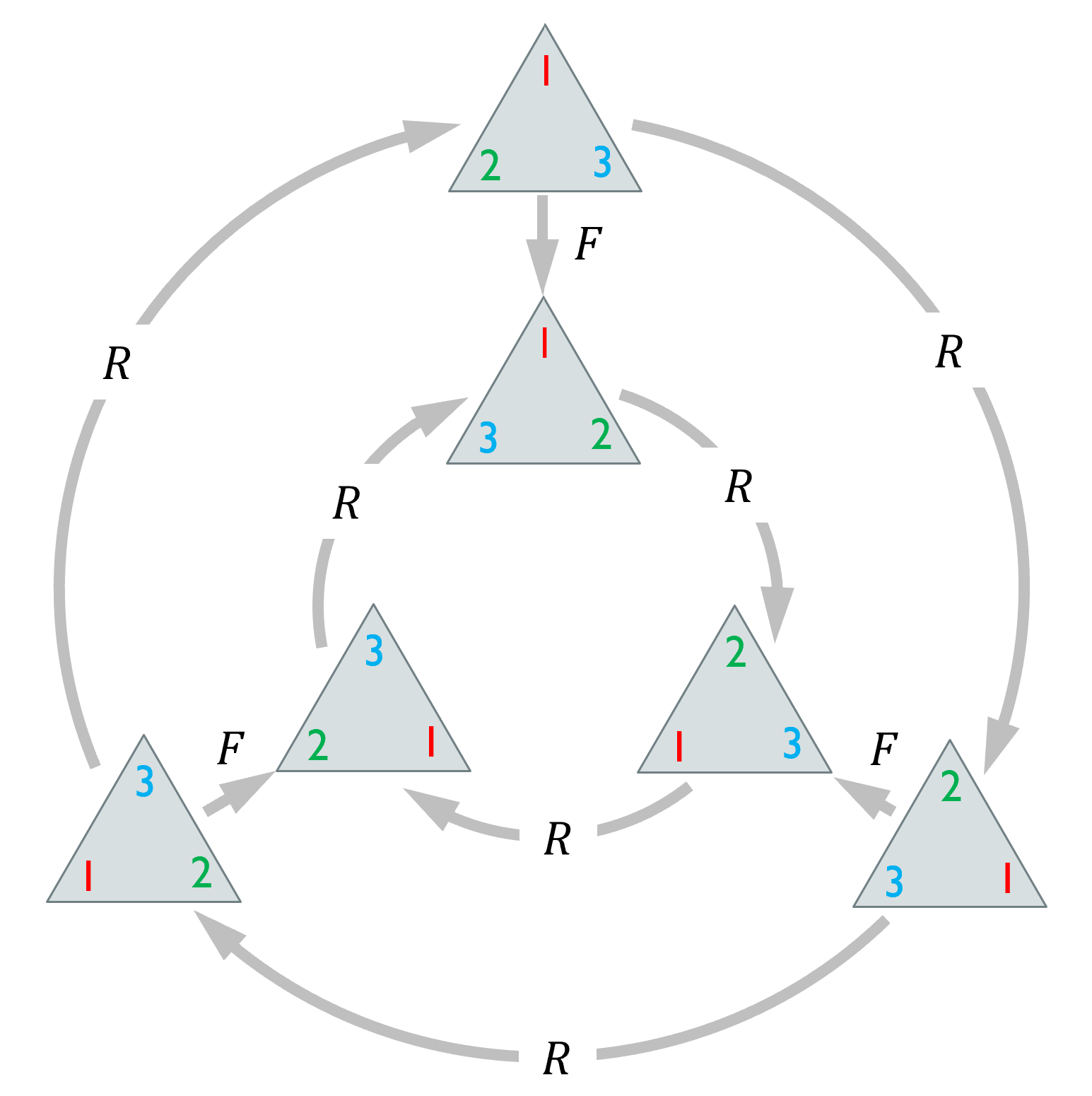}
    \caption{Cayley graph representing the symmetry of a triangle, where $R$ stands for rotation and $F$ for reflection.}
    \label{fig:cayley graph}
\end{figure}

\begin{tcolorbox}[colback=gray!10, colframe=gray!40]
A {\em group} is a set equipped with a binary operation that combines any two elements of the set to form a third element. In a group, the set and the operation can be denoted as $ (G, \circ) $, where $ G $ is the set and $ \circ $ is the binary operation. The operation must satisfy the following fundamental properties, known as the group axioms: \marginnote{$a \circ b$ can be denoted by juxtaposition for brevity: $a \circ b = ab.$ Also, alternatively, one can use the symbol $\ast$.}

\begin{itemize}
    \item {\em Associativity}: For all $ a, b, c \in G $, we have $ (a \circ b) \circ c = a \circ (b \circ c) $.
    \item {\em Identity Element}: There exists an element $ e \in G $ such that for all $ a \in G $, $ e \circ a = a \circ e = a $. This element is called the identity element.
    \item {\em Inverse Element}: For each element $ a \in G $, there exists an element $ b \in G $ such that $ a \circ b = b \circ a = e $, where $ e $ is the identity element. The element $ b $ is called the inverse of $ a $ and is denoted $ a^{-1} $.
\end{itemize}
\end{tcolorbox}

\marginnote{Group theory originated with Galois, who introduced the concept of permutation groups to show that general fifth-degree (quintic) polynomials cannot be solved by radicals. This settled a centuries-old problem that had perplexed mathematicians such as Lagrange and Ruffini. Interestingly, attempts to solve lower-degree equations (like quadratics) date back to ancient Babylonian mathematics.}\textit{Closure} follows from the definition: for all $ a, b \in G $, the result of the operation $ c = a \circ b $ is also in $G$, $ c \in G $, and commutativity does not necessarily apply in general. Groups can be finite, infinite, discrete, or continuous. 

\greyline

\paragraph*{Examples of Groups}
\begin{itemize}
    \item {\em Integers under Addition}: The set of integers $ \mathbb{Z} $ with the operation of addition $ (+) $ forms a group. The identity element is $ 0 $, and each integer $ a $ has an additive inverse $ -a $.
    \item {\em Non-zero Rational Numbers under Multiplication}: The set of non-zero rational numbers $ \mathbb{Q}^* = \mathbb{Q} \setminus \{0\}$ with multiplication $ (\cdot) $ forms a group. The identity element is $ 1 $, and each element $ a $ has a multiplicative inverse $ \frac{1}{a} $.
    \item {\em Symmetric Group}: The symmetric group $ S_N $ consists of all permutations of $ N $ elements. The group operation is the composition of permutations, and it is an example of a finite group.
\end{itemize}

\greyline

\paragraph*{More on Groups}

\begin{itemize}
    \item {\em Abelian Group}: A group $ (G, \circ) $ is called abelian (or commutative) if the operation is commutative, meaning $ a \circ b = b \circ a $ for all $ a, b \in G $.\marginnote{A non-abelian group contains at least some elements for which  $ a \circ b \neq b \circ a $.}
    \item {\em Subgroup}: A subgroup $ H $ of a group $ G $ is a subset of $ G $ that is itself a group under the operation of $ G $. If $ H $ is a subgroup of $ G $, we write $ H \leq G $.
    \item {\em Order of a Group}: The order of a group is the number of elements in the group, $|G|$. 
\end{itemize}

For instance, in our previous example, the group of rotations of a square, $C_4$, is abelian and has an order of 4. \marginnote{Another important abelian group is that formed by all rotations of three-dimensional space.} The group of rotations $C_2=\{0^\circ, 180^\circ\}$ is a subgroup $C_2\leq C_4.$ 

\begin{tcolorbox}[colback=orange!20, colframe=orange!60]
\textbf{Groups and Understanding Data Distributions through the Lens of Geometric Deep Learning.} In Geometric Deep Learning, groups formalize the concept of \textit{symmetry} in data. For instance, in computer vision, the group of translations ensures that object categories remain invariant when their positions shift, a property essential for tasks like visual object classification. In computational chemistry, predicting molecular properties requires outputs invariant to both rotations and translations, achieved through the Euclidean group $E(3)$. \marginnote{Graph Neural Networks~(GNNs) are a type of artificial neural networks designed to process signals over graph structures.} Similarly, for systems with discrete symmetries, such as permutations in graphs, the symmetric group $S_n$ plays a central role. This group underpins transformations where elements (e.g., particles or nodes) can be arbitrarily reordered, a key aspect in GNNs and the message-passing framework (Section~\ref{subsec:Vector Fields on Graphs}).
\end{tcolorbox}

\bigskip

\greyline

\paragraph*{Group Homomorphisms} It is often that we may find groups which are equivalent, or that can be realized in different ways. The essence of a \textit{group homomorphism} lies in preserving structure, rather than focusing solely on particular examples.

\begin{tcolorbox}[colback=gray!10, colframe=gray!40]
A {\em group homomorphism} is a map between two groups that preserves the group structure. Let $ (G, \circ) $ and $ (H, \ast) $ be two groups. A map $ \phi: G \to H $ is called a {\em group homomorphism} if, $ \forall a, b \in G $, the following condition holds:
$$
\phi(a \circ b) = \phi(a) \ast \phi(b).
$$
\end{tcolorbox}

A \textit{group isomorphism} is a bijective homomorphism between two groups $G$ and 
$H$, establishing a perfect identification between them.

\begin{tcolorbox}[colback=gray!10, colframe=gray!40]
Two groups $ (G, \circ) $ and $ (H, \ast) $ are said to be {\em isomorphic}, $(G, \circ) \cong (H, \ast),$ if there exists a bijective map (a one-to-one and onto mapping) $ \phi: G \to H $ such that $ \phi $ is a group homomorphism. 
\end{tcolorbox}

The group\marginnote{The modulo operation (denoted as $ a \mod n $) finds the remainder when $a$ is divided by $n$. Specifically, $ a \mod n $ is the integer remainder $r$ such that $ 0 \leq r < n $ and $ a = n \cdot q + r $ for some integer $q$.} $ C_4 = \{0^\circ, 90^\circ, 180^\circ, 270^\circ\} $ is the group of rotations of a square, where the group operation is addition modulo $ 360^\circ $.  Let the group $ \mathbb{Z}_4 = \{0, 1, 2, 3\} $ be the group of integers under addition modulo 4. These groups are isomorphic, and the isomorphism can be described by a homomorphism.

Define the homomorphism $ \phi: C_4 \to \mathbb{Z}_4 $ as
$$
\phi(0^\circ) = 0, \quad \phi(90^\circ) = 1, \quad \phi(180^\circ) = 2, \quad \phi(270^\circ) = 3.
$$
This mapping respects the group operation. Let us verify the homomorphism property. The group operation in $ C_4 $ is addition modulo $ 360^\circ $, and the group operation in $ \mathbb{Z}_4 $ is addition modulo 4. To verify $ \phi $ is a homomorphism, check that:
$$
\phi(a + b \mod 360^\circ) = \phi(a) + \phi(b) \mod 4, \quad \forall a, b \in C_4.
$$
Some examples include
$$\phi(90^\circ + 180^\circ \mod 360^\circ) = \phi(270^\circ) = 3,$$
$$ \phi(90^\circ) + \phi(180^\circ) \mod 4 = 1 + 2 \mod 4 = 3.$$

Next, let us illustrate a non-isomorphic mapping between $ C_4 $ and $ C_2 = \{0^\circ, 180^\circ\} $. While both $ C_4 $ and $ C_2 $ are cyclic groups, their structures are fundamentally different, and no isomorphism exists between them. However, there are still homomorphisms that preserve the group structure.

Let $ C_2 = \{0^\circ, 180^\circ\} $ where the group operation is addition modulo $ 360^\circ $. Define a homomorphism $ \psi: C_4 \to C_2 $ as:
$$
\psi(0^\circ) = 0^\circ, \quad \psi(90^\circ) = 180^\circ, \quad \psi(180^\circ) = 0^\circ, \quad \psi(270^\circ) = 180^\circ.
$$

This map is not injective (and therefore not bijective), which means that $ C_4 $ and $ C_2 $ are not isomorphic. Let us verify the homomorphism property. The group operation in both $ C_4 $ and $ C_2 $ is addition modulo $ 360^\circ $. To check that $ \psi $ is a homomorphism, we must verify:
$$
\psi(a + b \mod 360^\circ) = \psi(a) + \psi(b) \mod 360^\circ, \quad \forall a, b \in C_4.
$$

Some examples include: let $ a = 90^\circ $ and $ b = 180^\circ $
   $$
   \psi(90^\circ + 180^\circ \mod 360^\circ) = \psi(270^\circ) = 180^\circ,
   $$
   $$
   \psi(90^\circ) + \psi(180^\circ) \mod 360^\circ = 180^\circ + 0^\circ \mod 360^\circ = 180^\circ.
   $$

\greyline

\paragraph*{Group Actions}

A \textit{group action} is a formal way of describing how a group interacts with a set while preserving its structure. It connects abstract group theory to concrete situations where groups \textit{act} on mathematical or physical objects, such as transforming geometric shapes, permuting elements, or applying symmetry operations.

Let us revisit $C_4= \{0^\circ, 90^\circ, 180^\circ, 270^\circ\}$ once more. These rotations act on the set of vertices of the square, $$ V = \{\hat{A}, \hat{B}, \hat{C}, \hat{D}\}, $$ by permuting their positions. For example:
\begin{itemize}
    \item A $90^\circ$ rotation maps $ \hat{A} \to \hat{B} $, $ \hat{B} \to \hat{C} $, $ \hat{C} \to \hat{D} $, $ \hat{D} \to \hat{A} $.
    \item A $180^\circ$ rotation maps $ \hat{A} \to \hat{C} $, $ \hat{B} \to \hat{D} $, $ \hat{C} \to \hat{A} $, $ \hat{D} \to \hat{B} $.
\end{itemize}
This interaction satisfies the structure-preserving properties of a group action.

\begin{tcolorbox}[colback=gray!10, colframe=gray!40]
A \textit{(left) group action} of a group $ G $ on a set $ X $ is a mapping:\marginnote{The group operation vanishes on the right-hand side of the compatibility axiom because it is implicitly handled by the action itself. The key idea is that group actions are associative with respect to the group operation. This means that applying the action of $ a \circ b $ to $ x $ is the same as first applying $ b $ to $ x $ and then applying $ a $ to the result.}
$$
\alpha: G \times X \to X, \quad (g, x) \mapsto \alpha(g, x) = g \cdot x,
$$
satisfying the following axioms:
\begin{itemize}
    \item \textit{Identity}: The identity element $ e \in G $ acts as the identity transformation on $ X $:
    $$
    \alpha(e, x) = e \cdot x = x, \quad \forall x \in X.
    $$
    \item \textit{Compatibility}: $ \forall  g, a \in G $ and $ x \in X $, the action satisfies:
    $$
    (g \circ a) \cdot x = g \cdot (a \cdot x),
    $$
    where $ \circ $ is the group operation in $ G $. 
\end{itemize}
\end{tcolorbox}

\begin{tcolorbox}[colback=orange!20, colframe=orange!60]
\textbf{Groups Actions on Data.}\marginnote{In Geometric Deep Learning, we assume there is a domain underlying our data, which we denote by $\Omega$, and study how groups act on $\Omega$ and how we obtain actions on the same group on the space of signals $\mathcal{X}(\Omega)$.} In Geometric Deep Learning, rather than considering groups as abstract entities, we focus on how different mathematical operations, which we can prescribe for our artificial neural network, transform the input data. This enables us to design our model to perform transformations on the data that respect the structure of its domain.
\end{tcolorbox}

\begin{figure}[hbpt!]
    \centering  \includegraphics[width=\linewidth]{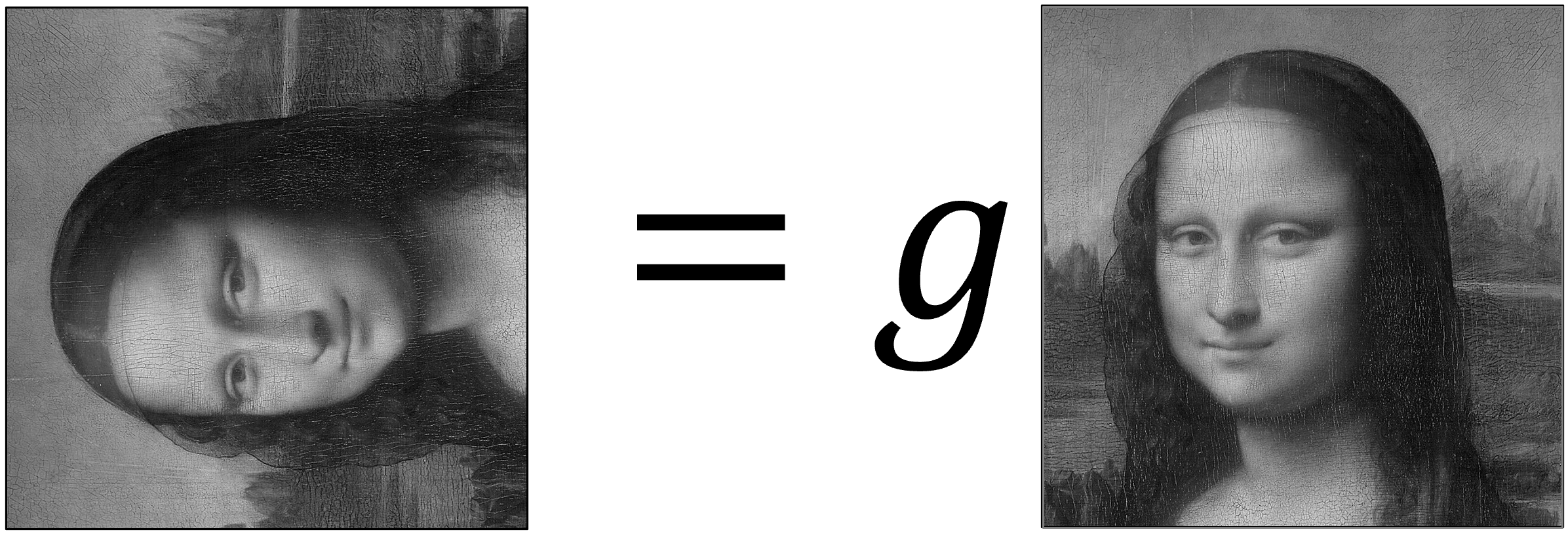}
    \caption{Group action on an image (function). The type of an object can be defined by the way it is transformed by a group.}
    \label{fig:mona_lisa_group_action}
\end{figure}

\greyline

\paragraph*{Group Orbits, Invariance, and Equivariance} We expand on our previous discussion by introducing a few formalisms.

\begin{tcolorbox}[colback=gray!10, colframe=gray!40]
The \textit{orbit} of an element \( x \in X \) under the action of \( G \) is defined as:
\[
\textrm{Orb}(x) = \{ g \cdot x \mid g \in G \}.
\]
\end{tcolorbox} 

That is, the orbit of \(x\) under a group \(G\) is the set of all points one can reach from \(x\) by applying every possible action in \(G\).

Before proceeding further, it is useful to formalize the notions of invariant and equivariant functions. Let \( X \) and \( Y \) be sets on which a group \( G \) acts. 

\begin{tcolorbox}[colback=gray!10, colframe=gray!40]
A function \( f: X \to Y \) is called \emph{\( G \)-invariant} if
\[
f(g \cdot x) = f(x) \quad \forall \, g \in G, \, x \in X.
\]
\end{tcolorbox}

In contrast, 

\begin{tcolorbox}[colback=gray!10, colframe=gray!40]
Let \((X, \cdot_X)\) and \((Y, \cdot_Y)\) be \(G\)-spaces, meaning that the group \(G\) acts on \(X\) via \(\cdot_X\) and on \(Y\) via \(\cdot_Y\). A function \(f: X \to Y\) is said to be \emph{\(G\)-equivariant} if
\[
f(g \cdot_X x) = g \cdot_Y f(x) \quad \forall\, g \in G,\, x \in X.
\]
\end{tcolorbox}

Thus, while an invariant function collapses the entire orbit to a single value, an equivariant function transforms in a predictable way under the group action.

Perhaps somewhat abstractly, one common method to achieve invariance in a neural network is to aggregate over these orbits. For example, a group convolution operator is defined as
\[
(f \star \psi)(x) = \sum_{g \in G} f(g \cdot x)\, \psi(g^{-1}),
\]
or in the continuous setting,
\[
(f \star \psi)(x) = \int_{G} f(g \cdot x)\, \psi(g^{-1})\, dg,
\]
where \( \psi: G \to \mathbb{R} \) is a kernel function and \( dg \) denotes the Haar measure on \( G \).\marginnote{A kernel function \( \psi \) assigns weights to the contributions of different group elements, much like a filter in a convolution, while the Haar measure \( dg \) is a translation-invariant measure on \( G \) that ensures integration over the group is independent of the specific parametrization.} This operator is \( G \)-equivariant, meaning that applying a transformation to the input before the convolution yields the same result as applying it after convolution.

Let us give an intuitive explanation to unravel what our previous mathematical abstraction really means. Consider \( G \) as the group of rotations by \( 90^\circ \), again this is $C_4$, acting on the set \( X \) of images. For a given image \( x \in X \) (for example, the Mona Lisa), its orbit \( \textrm{Orb}(x) \) will contain all four rotated copies: \( x, R_{90}(x), R_{180}(x), R_{270}(x) \in \textrm{Orb}(x) \). An invariant function \( f \) (such as one used for face recognition) would output the same value for each image in the orbit

$$f(x)=f(R_{90}(x)) = f(R_{180}(x))=f(R_{270}(x)),$$ 

recognizing that they all represent the same underlying face despite different orientations.

\begin{tcolorbox}[colback=orange!20, colframe=orange!60]
\textbf{Invariance and Equivariance in Geometric Deep Learning.}\marginnote{The importance of invariance and equivariance came to the forefront much earlier in Physics:

\vspace{10pt}

\textit{“Every [differentiable] 
symmetry of the action of a 
physical system [with 
conservative forces] has a 
corresponding conservation 
law”} -- Emmy Noether, 1918

\vspace{10pt}

\textit{“It is only slightly overstating the case to say that 
Physics is the study of symmetry”} -- Philip Anderson, 1972
} By interleaving transformations that respect the symmetry of the input, Geometric Deep Learning architectures can be made both expressive and robust. This strategy enables the design of models that generalize better and are more interpretable in settings where the data exhibit natural symmetries. More concretely, stacking several equivariant layers enables the network to capture increasingly complex hierarchical patterns while respecting the underlying symmetry. The final invariant operation then distills these symmetry-preserving features into a robust representation suitable for tasks such as classification, segmentation, or regression, where the output should not depend on the particular transformation applied to the input. Note that stacking only invariant transformations would result in a strictly smaller hypothesis class.
\end{tcolorbox}

\greyline

\paragraph*{Fields} Before moving on to discussing vector spaces, let us briefly mention fields. Fields, groups, and vector spaces are interconnected in the hierarchy of algebraic structures. A group has a single binary operation with minimal axioms, while a field has two operations with stringent compatibility conditions. Hence, fields impose more structure than groups and belong to a different class of algebraic objects.

\begin{tcolorbox}[colback=gray!10, colframe=gray!40]
A {\em field} is a set $ \mathbb{F} $ equipped with two binary operations, addition ($+$) and multiplication ($\cdot$), satisfying the following properties:

\begin{itemize}
    \item $ (\mathbb{F}, +) $ forms an abelian group (with identity element $ 0 $).
    \item $ (\mathbb{F} \setminus \{0\}, \cdot) $ forms an abelian group (with identity element $ 1 $).
    \item Multiplication is distributive over addition: $ \forall a, b, c \in \mathbb{F} $, $ a \cdot (b + c) = (a \cdot b) + (a \cdot c) $.
\end{itemize}
\end{tcolorbox}

\greylinelong

\subsection{Vector Spaces}

Now that we have a basic understanding of groups and fields, we can introduce the concept of a \textit{vector space}. 
Vectors are ubiquitous in applications of mathematics, especially in physical sciences. Introductory courses often talk about vectors in geometric terms (`arrows that have direction and length') or computer science terms (`arrays of numbers'). Each of these definitions are a crime against humanity: in order to think of vectors as `arrows', one has to define direction and length by introducing additional structures called inner products and norms; in order to think of vectors as `arrays', one has to define a basis, with respect to which vectors can be represented as ordered sets of coordinates. 
The correct mathematical way of thinking of vectors is as abstract objects that can be scaled and added. 

\begin{tcolorbox}[colback=gray!10, colframe=gray!40]\marginnote{A scalar is a single numerical value, such as a real number, with no direction or dimension. A vector is an ordered array of numbers, representing a point or direction in space, and can be one-dimensional or multi-dimensional. A tensor is a generalization of scalars and vectors to higher dimensions, represented as multi-dimensional arrays. For instance, scalars are 0th-order tensors, vectors are 1st-order tensors, and matrices are 2nd-order tensors. Higher-order tensors extend this concept, representing data with more than two dimensions, such as a sequence of matrices. In Deep Learning we tend to work with high-dimensional tensors.}
$ V $ is a {\em vector space} over a field $ \mathbb{F} $ (typically $ \mathbb{F} = \mathbb{R} $ or $ \mathbb{C} $) with binary operations $ + : V \times V \rightarrow V $ ({\em vector addition}) and $ \cdot : V \times \mathbb{F} \rightarrow V $ ({\em scalar multiplication}) if for any $ u, v, w \in V $ and $ \alpha, \beta \in \mathbb{F} $ we have the following properties:

\begin{itemize}
    \item {\em Associativity} of $+$: $ u + (v + w) = (u + v) + w $
    \item {\em Commutativity} of $+$ : $ u + v = v + u $
    \item {\em Identity element} of $+$: There exists a unique $ 0 \in V $ such that $ u + 0 = u $
    \item {\em Inverse element} of $+$: There exists a unique $ -v \in V $ such that $ v + (-v) = 0 $
    \item {\em Distributivity} of $\cdot$ w.r.t. vector addition : $ \alpha \cdot (u + v) = \alpha \cdot u + \alpha \cdot v $
    \item {\em Distributivity} of $\cdot$ w.r.t. scalar addition: $ (\alpha + \beta) \cdot v = \alpha \cdot v + \beta \cdot v $
    \item {\em Compatibility} of $\cdot$ with scalar multiplication: $ \alpha \cdot (\beta \cdot v) = (\alpha \cdot \beta) \cdot v $
    \item {\em Identity element} of $\cdot$: $\exists ! 1 \in \mathbb{R} \quad \text{s.t.} \quad 1 \cdot u = u$
\end{itemize}
\end{tcolorbox}

Note that notation sometimes can be confusing and therefore should be used with care. The same notation is used for scalar addition $ \alpha + \beta $ and vector addition $ u + v $. It should be understood from context which addition is meant. The same notation is also used for scalar-by-scalar multiplication $ \alpha \cdot \beta $ and vector-by-scalar multiplication $ \alpha \cdot u $. When no confusion arises, the vector-by-scalar multiplication is often denoted as $ \alpha u $ for brevity. The zero vector $ 0 \in V $ (identity element of vector addition) should not be confused with the zero scalar $ 0 \in \mathbb{R} $ (identity element of scalar addition), even though they are often denoted in the same way. Lastly, $ \exists ! u \in V $ means `there exists a unique $ u $ in $ V $', and it implies that there is exactly one element $ u \in V $ such that a particular condition is satisfied.

\greyline

\paragraph*{Examples of Vector Spaces}
\begin{itemize}
    \item {\em Vectors}: $ \mathbb{R}^n = \{ (v_1, \dots, v_n) : v_i \in \mathbb{R}, \forall i = 1, \dots, n \} $ with $ u + v = (u_1 + v_1, \dots, u_n + v_n) $
    \item {\em Functions}: $ \mathcal{F}(\Omega) = \{ f : \Omega \rightarrow \mathbb{R} \} $ with $ (f + g)(x) = f(x) + g(x) $ \marginnote{$\mathcal{F}$ is used to denote a set of functions on the domain $\Omega$. That is, the set $\mathcal{F}(\Omega)$ consists of functions whose domain is $\Omega$. Here, we talk about functions instead of maps, since we are considering special types of maps that map a set $\Omega$ to $\mathbb{R}$, rather than to an arbitrary set.}
\end{itemize}

\greyline

\paragraph*{From Vector Spaces to Tensor Spaces} Although it is less commonly discussed in basic linear algebra than vector spaces, in practice in Deep Learning we work with \textit{tensor spaces}. A \textit{tensor} is a multi-dimensional generalization of vectors and matrices. Tensors are particularly relevant in Deep Learning for parallel data processing.

Next, we discuss some basic examples.  A \textit{scalar} is a tensor of order (or rank) 0, represented by a single value, say
$$
a = 5
$$

A \textit{vector} is a tensor of order 1, represented as a one-dimensional array
$$
v = \begin{bmatrix} 1 \\ 2 \\ 3 \end{bmatrix}.
$$

A \textit{matrix} is a tensor of order 2, represented as a two-dimensional array
$$
M = \begin{bmatrix} 1 & 2 \\ 3 & 4 \\ 5 & 6 \end{bmatrix}.
$$

A higher-order tensor, of order 3 in this example, is a $n$-dimensional array, represented as
$$
T_{ijk} = 
\begin{bmatrix} 
\begin{bmatrix} 1 & 2 \\ 3 & 4 \end{bmatrix}, & 
\begin{bmatrix} 5 & 6 \\ 7 & 8 \end{bmatrix}, &
\begin{bmatrix} 9 & 10 \\ 11 & 12 \end{bmatrix} 
\end{bmatrix}.
$$
$T_{ijk}$ in this particular instance represents a $3\times 2 \times 2$ tensor. Alternatively, we can express the slices more clearly:
$$
T_{ijk} = \left\{ 
\begin{aligned}
T_{1,:,:} &= \begin{bmatrix} 1 & 2 \\ 3 & 4 \end{bmatrix}, \\
T_{2,:,:} &= \begin{bmatrix} 5 & 6 \\ 7 & 8 \end{bmatrix}, \\
T_{3,:,:} &= \begin{bmatrix} 9 & 10 \\ 11 & 12 \end{bmatrix}.
\end{aligned}
\right.
$$

The \textit{Einstein summation convention} is a shorthand for tensor expressions, where repeated indices imply summation over all their possible values. This convention makes it easier to work with high-dimensional tensors. Let us look at some examples of Einstein summation.

The dot product of two vectors $u$ and $v$ in Einstein notation is written as

$$
u_i v^i = \sum_i u_i v^i = a,
$$

where we can omit the summation symbol, and the product results in a scalar, $a$.

For a matrix $M$ and vector $v$, the matrix-vector multiplication in Einstein notation is

$$
M_{ij} v^j = \sum_j M_{ij} v^j = u_i,
$$

which results in another vector, $u$.

Likewise, a tensor contraction, which is a generalization of matrix multiplication, can be written as:
$$
T_{ijk} v^j = \sum_j T_{ijk} v^j = M_{ik}.
$$
This involves summing over the index $j$, since it is the repeated index. Multiplying an order 3 tensor and a vector, results in an order 2 tensor, that is, a matrix.

\begin{tcolorbox}[colback=orange!20, colframe=orange!60]
\textbf{Tensor Spaces in Deep Learning.} A \textit{tensor space} can be thought of as a generalization of vector spaces to higher-dimensional objects, where tensors (multi-dimensional arrays) act as elements in these spaces. More formally, a tensor space can be described as a set of tensors where tensor addition and scalar multiplication follow the usual rules that hold for vector spaces, but are generalized to multi-dimensional arrays. For instance in Computer Vision, we typically process tensors of shape $\texttt{[B,C,H,W]},$ where $\texttt{B}$ stands for batch size, $\texttt{C}$ for channel dimension (RGB channels), and $\texttt{H}$ and $\texttt{W}$ are the height and width of the image. In the context of video we can further include a frames (or time) dimension and the tensor gains an additional dimension, $\texttt{[B,C,F,H,W]}$. However, oftentimes in research articles, transformations are represented in terms of matrices, and additional entries such as those for the batch dimension are omitted for clarity.
\end{tcolorbox}

\clearpage

\section{Geometric and Analytical Structures}

Geometric structures bring life to abstract mathematical objects by introducing familiar concepts like distance, size, and angles. While groups and vector spaces give us powerful ways to study relationships and transformations, they lack the geometric intuition we often need in real-world applications. 

\greylinelong

\subsection{Norms and Normed Vector Spaces}

A norm is a mathematical function that quantifies the size or magnitude of a mathematical object, generalizing our intuitive understanding of length or distance in physical space. Like physical length, a norm assigns a non-negative real number to an object while satisfying specific properties.

\begin{tcolorbox}[colback=gray!10, colframe=gray!40]

\marginnote{The field can also be $\mathbb{F} = \mathbb{C}$, but in the main text we stick to $\mathbb{R}$ for simplicity.} Given a vector space $V$ over a field $\mathbb{F} = \mathbb{R}$, a {\em norm} is a function $\| \, \| : V\rightarrow \mathbb{R}$ satisfying for any $u,v \in V$ and $\alpha \in \mathbb{R}$:

\begin{itemize}

\item \textit{Positive homogeneity:} \itab{$\| \alpha u\| = |\alpha | \| u \|$} 

\item \textit{Triangle inequality:} \itab{$\| u + v \| \leq \| u \| + \| v \|$}

\item \textit{Positive definiteness:} \itab{$\| u \| = 0 \quad \Rightarrow \quad u = 0$}

\end{itemize}

$(V, \| \, \|)$ is called a {\em normed (vector) space}. Intuitively, the norm measures the length of a vector. 

\end{tcolorbox}

The following properties (often listed as part of axiomatic definition of the norm) are in fact consequences of the above definition:\marginnote{The notation  $\| u \|$ refers to the norm of an element in a vector space, where $u$ is a vector. In contrast,  $|\alpha|$ denotes the absolute value of a scalar, which is a special case of a norm when the underlying field is the real or complex numbers. While the norm generalizes the concept of absolute value to vector spaces, the absolute value is specifically used for scalars.}

\begin{itemize}

\item $\| 0 \| = \| 0\cdot u \| \overset{\tiny{(1)}}{=} |0| \| u\| = 0$, i.e. property (3) is iff: $\|u\| =0 \Leftrightarrow u=0$. 

\item $\| u\| \geq 0$, 
\end{itemize}  

where (1) refers to positive homogeneity and (3) to positive definiteness.

\greyline

\paragraph*{Examples of Norms} 

\begin{itemize}

\item {\em $L_p$-norm on $\mathbb{R}^n$:} $\| u \|_p = \left(\sum_{i=1}^n|u_i|^p \right)^{1/p} $, in particular

\begin{itemize}
     \item {\em $L_1$-norm:} $\| u \|_1 = \sum_{i=1}^n |u_i|$ 

     \item {\em $L_2$-norm (Euclidean norm):} $\| u \|_2 = \sqrt{\sum_{i=1}^n |u_i|^2}$\marginnote{The $ L_2 $-norm, also known as the Euclidean norm, is the most commonly used norm, and it provides the notion of the length of a vector.}

\item {\em $L_\infty$-norm:} $\| u \|_\infty = \max\{ |u_1|, \hdots, |u_n|\}$ 

\end{itemize}
\item {\em $L_p$-norm on $\mathcal{F}(\Omega)$:} $\| f \|_p = \left( \int_\Omega |f(x)|^p dx \right)^{1/p} $

\end{itemize}

The summation in the vector case, is replaced by an integral in the function case. This is because functions can be thought of as vectors with infinitely many components, where the integral serves as a continuous analog of the sum.

\begin{tcolorbox}[colback=orange!20, colframe=orange!60]
\textbf{Norms in Geometric Deep Learning.} Norms quantify the magnitude of vectors and are fundamental for enabling invariant feature representations in Geometric Deep Learning architectures, particularly under transformations such as rotations and reflections. Additionally, norms play a key role in regularization in Deep Learning. For example, weight decay penalizes the Euclidean norm of model parameters to prevent overfitting and encourage generalization.
\end{tcolorbox}

\greylinelong

\subsection[Metrics Induced by Norms and Metric Spaces]{Metrics Induced by Norms and Metric Spaces}\marginnote{A metric measures the distance between two elements in a space, generalizing our intuitive notion of distance in physical space. Unlike norms which measure the size of a single vector, metrics quantify the separation between pairs of elements.}

A metric represents a mathematical way to measure distances between elements in a set, with norms being a special case that can generate metrics.
\begin{tcolorbox}[colback=gray!10, colframe=gray!40]
Given a normed vector space $(V, \| \cdot \|)$, a {\em metric} $d: V \times V \to \mathbb{R}$ is naturally defined by:
$$
d(u,v) = \| u - v \|, \quad \forall u,v \in V.
$$
This metric satisfies the following properties, making $(V,d)$ a {\em metric space}:
\begin{itemize}
\item \textit{Non-negativity:} \itab{$d(u,v) \geq 0$} \marginnote{While normed vector spaces are inherently metric spaces, not all metric spaces have the additional algebraic structure of a vector space. A vector space requires operations like vector addition and scalar multiplication that satisfy specific axioms. Many metric spaces lack these operations or do not satisfy the vector space axioms. For instance, in $\mathbb{R}^n$, the metric $d(u, v) = \sqrt{|u_1 - v_1|} + \cdots + \sqrt{|u_n - v_n|}$ is a valid metric but cannot be derived from a norm.}
\item \textit{Identity of indiscernibles:} \itab{$d(u,v) = 0 \Leftrightarrow  u = v$}
\item \textit{Symmetry:} \itab{$d(u,v) = d(v,u)$}
\item \textit{Triangle inequality:} \itab{$d(u,w) \leq d(u,v) + d(v,w)$}
\end{itemize}

\end{tcolorbox}

Note that every normed vector space is also a metric space with a metric induced by its norm. However, not all metric spaces are normed vector spaces. 

\greyline

\paragraph*{Examples of Metrics Induced by Norms}
\begin{itemize}
\item {\em $L_p$ distance in $\mathbb{R}^n$:} $d_p(u,v) = \| u - v \|_p = \left(\sum_{i=1}^n |u_i - v_i|^p \right)^{1/p}$, in particular
\begin{itemize}
\item {\em $L_1$ distance:} $d_1(u,v) = \| u - v \|_1 = \sum_{i=1}^n |u_i - v_i|$
\item {\em $L_2$ distance (Euclidean distance):} $d_2(u,v) = \| u - v \|_2 = \sqrt{\sum_{i=1}^n |u_i - v_i|^2}$
\item {\em $L_\infty$ distance:} \marginnote{The Euclidean distance is the most intuitive metric, corresponding to the physical distance between points in space.}$d_\infty(u,v) = \| u - v \|_\infty = \max\{ |u_1 - v_1|, \hdots, |u_n - v_n| \}$
\end{itemize}
\item {\em $L_p$ distance for functions:} $d_p(f,g) = \| f - g \|_p = \left( \int_\Omega |f(x) - g(x)|^p dx \right)^{1/p}$
\end{itemize}

\greyline

\paragraph*{Generalizations of Metrics} The following are important generalizations of metrics:

\begin{itemize}
\item A {\em pseudo-metric}\marginnote{For instance, in the context of general relativity, the term pseudo-metric often refers to the metric tensor of spacetime, which is actually a pseudo-Riemannian metric.} is a function $d: V \times V \to \mathbb{R}$ satisfying all properties of a metric except the identity of indiscernibles. That is, $d(u,v) = 0$ does not necessarily imply $u = v.$ 
\item A {\em quasi-metric} also satisfies all properties of a metric space, but it relaxes the triangle inequality to:
$$
d(u,w) \leq \mathcal{C}(d(u,v) + d(v,w)),
$$
known as the {\em $\mathcal{C}$-relaxed triangle inequality}. When $\mathcal{C}=1$, this reduces to a standard metric space.
\end{itemize}

\greyline

\paragraph*{Hausdorff Distance}\marginnote{The Hausdorff distance is particularly useful in comparing shapes, curves, or other geometric objects in applications such as computer vision, shape analysis, and geometric deep learning. It is closely related to the Chamfer distance, which computes the average closest point distance instead. Furthermore, the Hausdorff distance can be generalized into the Gromov-Hausdorff distance, which is used to compare metric spaces rather than subsets of a fixed metric space. It provides a way to measure how `far apart' two metric spaces are, considering their intrinsic geometry rather than their embedding into a common space.}

The Hausdorff distance provides a way to measure how far apart two subsets of a metric space are. 

\begin{tcolorbox}[colback=gray!10, colframe=gray!40]
Given two non-empty subsets $A, B \subset V$ in a metric space $(V, d)$, the Hausdorff distance $d_H$ is defined as:
$$
d_H(A, B) = \max \left\{ \sup_{a \in A} \inf_{b \in B} d(a, b), \sup_{b \in B} \inf_{a \in A} d(b, a) \right\}.
$$
\end{tcolorbox}

Here, $d(a, b)$ is the distance between points $a \in A$ and $b \in B$ as defined by the metric $d$ on $V$. The Hausdorff distance satisfies the following properties:
\begin{itemize}
\item \textit{Non-negativity:} $d_H(A, B) \geq 0$, and $d_H(A, B) = 0$ if and only if $A = B$ (when $A$ and $B$ are closed sets).
\item \textit{Symmetry:} $d_H(A, B) = d_H(B, A)$.
\item \textit{Triangle inequality:} $d_H(A, C) \leq d_H(A, B) + d_H(B, C)$ for any subsets $A, B, C \subset V$.
\end{itemize}

In $\mathbb{R}^n$ with the Euclidean distance, the Hausdorff distance is often used to compare geometric objects such as polygons or point clouds.

\begin{tcolorbox}[colback=orange!20, colframe=orange!60]
\textbf{Metrics in Geometric Deep Learning.} Metrics define distance measures for comparing data points across graph, manifold, and point cloud representations, as well as in neural latent (embedding) spaces. In particular, the Euclidean distance $d_2(u,v) = \| u - v \|_2 = \sqrt{\sum_{i=1}^n |u_i - v_i|^2}$ is a natural choice in many Deep Learning implementations. For instance, in Geometric Deep Learning and computational biology, Euclidean distance is commonly used to construct unit disk graphs or k-nearest neighbor graphs in $\mathbb{R}^3$. This approach allows to define the connectivity structure of atomic point clouds, such as those derived from protein structures resolved via X-ray crystallography or cryo-Electron Microscopy, where nodes correspond to atoms and edges represent proximity-based interactions. Another notable example is vector quantization methods for neural discrete representation learning developed in the late 2010s, which use Euclidean distance to compare continuous latent embeddings with entries in a learned codebook. Moreover, beyond continuous metric spaces, we often leverage metrics induced by discrete structures such as graph geodesic distances to compute, for example, optimal commute times in transportation networks or information flow in social graphs.
\end{tcolorbox}

\greylinelong
\subsection[The Inner Product and Inner Product Spaces]{The Inner Product and Inner Product Spaces}\label{subsec:Inner Product}

In terms of hierarchy, metric spaces form the foundational mathematical structure defining distance, with normed vector spaces and inner product spaces representing progressively more specialized and structured mathematical environments. Normed vector spaces extend metric spaces by integrating a norm that naturally induces a metric, while inner product spaces further enhance this structure by introducing an inner product that generates a norm.

\begin{tcolorbox}[colback=gray!10, colframe=gray!40]\marginnote{The field can also be $\mathbb{F} = \mathbb{C}.$}
Given a vector space $V$ over a field $\mathbb{F}=\mathbb{R}$, an {\em inner product} is a function $\langle \, , \,\rangle : V\times V\rightarrow \mathbb{R}$ satisfying for any $u,v,w \in V$ and $\alpha \in \mathbb{R}$:

\begin{itemize}

\item \textit{Conjugate (Hermitian) Symmetry:} $\langle u, v\rangle  = \overline{\langle v, u\rangle} $\marginnote{The overline $\overline{(\cdot)}$ is used to denote the complex conjugate. For $z = a + bi,$ then its complex conjugate is: $\overline{z} = a - bi.$ Note that the complex conjugate of a real number is itself.}

\item \textit{Linearity:} $\langle \alpha u, v\rangle = \alpha \langle u, v\rangle,\,\langle u+w, v\rangle = \langle u, v\rangle + \langle w, v\rangle$ 

\item \textit{Positive Semi-Definiteness:} $\langle u, u\rangle \geq 0,\,\langle u, u\rangle = 0 \Leftrightarrow u=0$

\end{itemize}

\noindent $(V, \langle \, , \,\rangle)$ is called an {\em inner product space}. 
\end{tcolorbox}

The following additional property, called \textit{conjugate linearity} in the second argument, is a consequence of the above definition (considering the field to be $\mathbb{F} = \mathbb{C}$ for more generality):\marginnote{Here, we have applied in order: conjugate symmetry, linearity in the second argument of the inner product, the distributive property of complex conjugation, and substitution from the conjugate symmetry.}

$$\langle  u,\alpha v\rangle = \overline{\langle \alpha v, u\rangle} = \overline{\alpha \langle v, u\rangle} = \overline{\alpha} \cdot \overline{\langle v, u\rangle} = \overline{\alpha} \langle u, v\rangle.$$ 
Also, note that as previously discussed, in Einstein summation convention, repeated indices are implicitly summed over. For example, in the case of real vectors, we can write the inner product as:
    $$\langle u, v \rangle = u_i v_i,$$
\noindent
where the repeated index $i$ is implicitly summed over from $1$ to $n$.\marginnote{Gram-Schmidt orthogonalization is a method to transform a set of linearly independent vectors into an orthogonal (or orthonormal) set of vectors; eigenvalue decomposition factors a square matrix into a product involving its eigenvalues and eigenvectors; and principal component analysis is used to reduce the dimensionality of a dataset while retaining as much variance as possible.}

Inner products provide additional structure beyond what a norm alone can offer. In particular, they enable definitions of angles, orthogonality, and support advanced computational techniques like Gram-Schmidt orthogonalization, eigenvalue decomposition, and principal component analysis. These operations leverage the geometric insights intrinsic to inner product structures. Also, norms derived from inner products often have smoother behavior compared to arbitrary norms. This characteristic makes inner product spaces especially valuable in optimization contexts, where they facilitate natural gradient calculations and provide well-defined curvature representations. Finally note that inner products induce norms, but not vice versa.

\greyline

\paragraph*{Examples of Inner Products}

\begin{itemize}\marginnote{A square-integrable function is a function $ f $ defined on a domain $ \Omega $ such that the square of its absolute value is integrable over $ \Omega $. Specifically, a function $ f(x) $ belongs to the space $ L^2(\Omega) $ if:
$$
\int_\Omega |f(x)|^2 \, dx < \infty.
$$}

\item {\em Real vectors $\mathbb{R}^n$:} $\langle u, v \rangle =  \sum_{i=1}^n u_i v_i = u_i v_i = v^\top u$

\item {\em Complex vectors $\mathbb{C}^n$:} $\langle u, v \rangle =  \sum_{i=1}^n u_i \overline{v}_i = u_i \overline{v}_i = v^* u$

\item {\em Real matrices:} $\langle A, B \rangle =  \mathrm{trace(AB^\top)}$

\item {\em Square-integrable functions $L^2(\Omega)$:} $\langle f, g \rangle =  \int_\Omega f(x) \overline{g(x)} dx$ 

\item {\em Square-summable real sequences $\ell^2$:} $\langle x, y \rangle =  \sum_{i\geq 1} x_i y_i $\marginnote{A square-summable real sequence is a sequence of real numbers $\{a_n\}_{n=1}^\infty$ such that the sum of the squares of its elements is finite:
$$
\sum_{n=1}^\infty a_n^2 < \infty.
$$
}

\end{itemize}

\greyline

\paragraph*{Relation to Norms}

The inner product naturally defines a norm, given by 
$$
\| u \| = \left( \langle u, u \rangle \right)^{1/2}.
$$ 

This norm satisfies the \textit{Cauchy-Schwarz (Bunyakovsky) inequality:}
$$
|  \langle u, v \rangle | \leq \| u \| \cdot \| v \|.
$$
This inequality is crucial because it provides an upper bound on the inner product in terms of the magnitudes (norms) of the vectors, ensuring that the inner product cannot exceed the product of the norms of the vectors.

The cosine of the angle between two vectors is given by
$$
\cos \angle (u,v) = \frac{\langle u, v \rangle}{ \| u \| \cdot \| v \| },
$$
which expresses the relationship between the vectors in terms of their geometric angle. When $\langle u, v \rangle = 0$, the vectors are said to be orthogonal, meaning the angle between them is $90^\circ$ (i.e., $u \perp v$). This condition is essential for understanding orthogonality in inner product spaces.

Not every norm defines an inner product! A norm that satisfies the \textit{parallelogram law}:
$$
2\|u\|^2 + 2\|v\|^2 = \|u+v\|^2 + \|u-v\|^2,
$$
can be used to define an inner product via the \textit{polarization identity}:
$$
 \langle u, v \rangle = \frac{1}{4} \left( \|u+v\|^2 - \|u-v\|^2 \right).
$$
The parallelogram law provides a critical condition for determining whether a norm arises from an inner product. It describes how the lengths of vectors behave geometrically when combined through addition or subtraction. Specifically, it expresses a relationship between the squares of the lengths of the vectors and their sums and differences, mirroring the geometry of inner product spaces.

If the parallelogram law is not satisfied, then the norm cannot be derived from an inner product. Without this structure, we lose important geometric concepts like orthogonality, angles, and projections, which are fundamental to understanding the behavior of vectors in the space. For example, spaces with norms that do not satisfy the parallelogram law, such as the $ L_1 $ norm, do not allow for meaningful definitions of orthogonality or angles.

\begin{tcolorbox}[colback=blue!10, colframe=blue!60]
\begin{theorem}[Generalized Pythagorean Theorem]
For a set of pairwise orthogonal vectors $ v_1, v_2, \dots, v_n \in V $ (i.e., $ \langle v_i, v_j \rangle = 0 $ for $ i \neq j $), we have the following property:
$$
\left\| \sum_{i=1}^n v_i \right\|^2 = \sum_{i=1}^n \| v_i \|^2.
$$
\end{theorem}
\label{theorem:pythagoras}
\end{tcolorbox}
This result directly generalizes the \textit{Pythagorean theorem} from Euclidean geometry: when vectors are orthogonal, the square of the norm of their sum is equal to the sum of the squares of their individual norms.\marginnote{The original Pythagorean theorem states that in a right triangle with legs of length $a$ and $b$, and hypotenuse of length $c$, the relation $a^2 + b^2 = c^2$ holds. This theorem can be interpreted geometrically in Euclidean space as the sum of the squares of the orthogonal components of a vector.} For non-orthogonal vectors, the sum will be less than or equal to the square of the norm of the sum, by virtue of the triangle inequality.

\begin{tcolorbox}[colback=orange!20, colframe=orange!60]
\textbf{Inner products in Deep Learning.} The inner product between two vectors encodes similarity, but is not invariant to scale since large magnitudes can dominate even if directions differ. To mitigate this, the cosine similarity, defined as the normalized inner product, captures the directional alignment between vectors while discarding scale information. \textit{Scale invariance} can be particularly useful in Deep Learning, where activations can vary in norm due to factors such as network depth, normalization, or noise, but their direction in latent space often encodes semantic content. Inner products (and their normalized counterparts) are smooth, linear functions that provide more stable comparisons than raw norms, such as \( L_p \) distances. They underpin attention mechanisms in Transformers (the ubiquitous neural network architecture that has impregnated all realms of Deep Learning), where scaled dot-product attention is used. While the dot-product itself is not scale-invariant, scaling by \( \frac{1}{\sqrt{d}} \) reduces the sensitivity to vector norm and makes the softmax activation more numerically stable. Interestingly, in high-dimensional latent spaces the curse of dimensionality can become a blessing: random vectors are almost always nearly orthogonal which allows neural networks to store a large number of features in directions that do not interfere with each other. In short, high-dimensional latent spaces can pack more information than their dimension may initially suggest, thanks to near-orthogonality.
\end{tcolorbox}

\clearpage

\section{Vector calculus}
\label{sec: Vector calculus}

Scalar and vector fields represent quantities that vary across space. These concepts differ from the abstract notion of a vector space, which is purely an algebraic structure. In this section, we examine scalar fields, vector fields, and calculus, which provides essential tools for quantifying variations across space. The latter enables the description of scalar and vector field behavior through operations like differentiation and integration. Differentiation is used to quantify local field behavior, while integral operators establish relationships between infinitesimal variations and macroscopic field properties.

\greylinelong

\subsection{(Lipschitz) Continuity, Differentiability, and Smoothness}

In practice, modeling scalar and vector fields is common in Geometric Deep Learning, particularly in applications such as data-driven physics simulations and 3D graphics. These fields are often represented as, or assumed to be, continuous functions that can be approximated using artificial neural networks.

\greyline 

\paragraph*{Continuity} For a function to be continuous at a point, the limit of the function as we approach that point must exist and be equal to the function's value at that point. In simpler terms, a continuous function has no abrupt jumps or breaks and `can be drawing without lifting your pen from the page'.

\begin{tcolorbox}[colback=gray!10, colframe=gray!40]
\textit{Continuity} of a function $f$ at a point $x_0$ requires:
\begin{itemize}
\item The limit $\lim_{x \to x_0} f(x)$ exists,
\item $\lim_{x \to x_0^-} f(x) = \lim_{x \to x_0^+} f(x)$ (the limit is independent of the direction from which $x$ approaches $x_0$),
\item and the limit and function value must be equal $f(x_0) = \lim_{x \to x_0} f(x)$.
\end{itemize}
\end{tcolorbox}

Note that the mention of one-sided limits (\( \lim_{x \to x_0^-} \) and \( \lim_{x \to x_0^+} \)) is specific to functions on \( \mathbb{R} \), where continuity is analyzed along a single dimension. For higher dimensions, this concept generalizes to approaching \( x_0 \) from any direction. If the requirements above are satisfied we say that $f$ is a \textit{continuous function}.

\begin{tcolorbox}[colback=gray!10, colframe=gray!40]
A function $f$ is \textit{Lipschitz continuous} with Lipschitz constant $L$ if for all $x, y \in \mathbb{R}^n$:
$$
|f(x) - f(y)| \leq L |x - y|
$$
\end{tcolorbox}
\marginnote{In optimization, the notion of Lipschitz continuity is sometimes used to provide guarantees regarding the convergence of algorithms based on iterative methods.}Lipschitz continuity bounds the rate of change of a function and ensures that it does not change too rapidly between any two points. The Lipschitz constant $L$ provides an upper bound on the function's local slope or steepness. Functions that are Lipschitz continuous are always continuous but not vice versa.

\greyline 

\paragraph*{Differentiability and Smoothness} Differentiability is a stronger condition than continuity. While a continuous function ensures smooth variation, a differentiable function provides additional information about the rate of change. The existence of derivatives at every point implies that the function can be well-approximated by its tangent line or hyperplane locally.

\begin{tcolorbox}[colback=gray!10, colframe=gray!40]
A function $f$ is said to be \textit{smooth} when its derivatives exist up to a certain order and are continuous. We denote this using $\mathcal{C}^k$ notation:
\begin{itemize}
\item $\mathcal{C}^0$: Continuous function
\item $\mathcal{C}^1$: Continuously differentiable (first derivatives are continuous)
\item $\mathcal{C}^k$: $k$ times continuously differentiable
\item $\mathcal{C}^{\infty}$: Infinitely differentiable (derivatives of all orders exist and are continuous)
\end{itemize}
\end{tcolorbox}
Smoothness represents progressively stronger conditions on a function's differentiability. As the smoothness class increases from $\mathcal{C}^0$ to $\mathcal{C}^{\infty}$, the function becomes increasingly well-behaved. Note that being \textit{continuously differentiable} is a stronger condition that being \textit{differentiable} alone, since it implies that the derivative does not only exist but it is also continuous.

\greylinelong

\subsection{Scalar Fields, Vector Fields, and Signals}
\label{subsec:Scalar Fields, Vector Fields and Signals}

\begin{tcolorbox}[colback=gray!10, colframe=gray!40]
A \textit{scalar field} is a function $f: \mathbb{R}^n \to \mathbb{R}$ that assigns a single scalar value to every point in $n$-dimensional space, $f(x) = f(x_1, \hdots, x_n).$
\end{tcolorbox}

In $\mathbb{R}^3$, $f(x, y, z)$ could represent the temperature at a specific point $(x, y, z)$ in a room. The value of $f(x)$ at each point is a scalar, meaning it has magnitude but no direction.

\begin{tcolorbox}[colback=gray!10, colframe=gray!40]
A \textit{vector field} is a function $F: \mathbb{R}^n \to \mathbb{R}^m$ that assigns a vector to each point in space. 
\end{tcolorbox}

For instance, in $\mathbb{R}^3$, $F(x, y, z) = (F_1(x, y, z), F_2(x, y, z), F_3(x, y, z))$ might represent the velocity of a fluid or the direction and magnitude of a force at each point in space. In this physical example, the value of $F(x)$ at each point has both magnitude and direction, distinguishing it from a scalar field. Note, however, that in the mathematical sense, a vector field is simply a function that assigns a vector to each point in some domain, hence, strictly speaking each of the vector field components can be an independent scalar function.

While the definitions above assume the domain is Euclidean $\mathbb{R}^n$, they extend naturally to more general domains $\Omega$, such as graphs or manifolds. In such cases, derivatives are interpreted using the domain's intrinsic structure (e.g., graph gradients or Laplacians for graphs, and covariant derivatives on manifolds). We will discuss this in more depth in Section~\ref{Graph Theory}.

\begin{tcolorbox}[colback=orange!20, colframe=orange!60]
\textbf{What do we mean by Signals.} We define signals as mappings from a domain $\Omega$ to a vector space $\mathcal{C}$, whose dimensions are referred to as `channels' in Deep Learning terminology. In the most general case, $\Omega$ does not necessarily possess a vector space structure. Therefore, when we use the term `signal', we are referring to a vector field $F: \Omega \to \mathcal{C}$, where $\mathcal{C} = \mathbb{R}^m$ and $m$ denotes the number of channel dimensions. In physics, $\Omega$ is often Euclidean space, but in Geometric Deep Learning, it could be another non-Euclidean structure, such as a graph. If $m=1$ this would be a scalar field instead. We often can vectors in $\mathcal{C}$ `feature vectors'. 
\end{tcolorbox}

\greylinelong

\subsection{Derivatives and Gradients}
\label{subsec:Derivatives and Gradients}

A derivative captures how a function changes with respect to a change in its input. More concretely, it quantifies the rate of change or the slope of the function at a given point.

\begin{tcolorbox}[colback=gray!10, colframe=gray!40]
\marginnote{In this context, by smoothness we imply being at least twice continuously differentiable (often denoted as $\mathcal{C}^2$), i.e., having continuous second-order derivatives.}  Let $f: \mathbb{R}^n\rightarrow \mathbb{R}$ be a smooth scalar field. A {\em directional derivative} of $f$ at $x$ in direction $d \in \mathbb{R}^n$ is given by 
$$
\partial_d f(x) = f_{x_i}(x) = \lim_{\epsilon \rightarrow 0}\, \frac{f(x+\epsilon d) - f(x)}{\epsilon}.
$$ 
\end{tcolorbox}

\begin{tcolorbox}[colback=gray!10, colframe=gray!40]\marginnote{The directional derivative quantifies how the function $f$ changes as one moves from the point $x$ in the direction specified by the vector $d$.}
A {\em partial derivative} of $f$ at $x$ w.r.t. coordinate $x_i$ is given by 
$$
\frac{\partial}{\partial x_i} f(x) = f_{x_i}(x) = \lim_{\epsilon \rightarrow 0}\, \frac{f(x_1, \hdots, x_i + \epsilon, \hdots, x_n) - f(x_1, \hdots, x_n)}{\epsilon},
$$ 
and is thus a directional derivative in the direction $x_i$. 
\end{tcolorbox}

Hence, partial derivatives are special cases of directional derivatives, where the direction aligns with the unit vector along the $i$-th coordinate axis.

In its simplest form, when the scalar field has a single input dimension $ f: \mathbb{R} \to \mathbb{R}$, the derivative $f'(x)$ measures the rate of change of $f$ with respect to the single variable $x$, and we can simply right $f'(x) = \frac{d}{dx}f(x)$, instead of using the $\partial$ notation.

\greyline

\paragraph*{Numerical Methods and Approximations of the Derivative}

To compute derivatives in practical settings, especially when analytical expressions are unavailable, numerical methods are used. These approximations leverage finite differences to estimate derivatives.

For a scalar field $f: \mathbb{R} \to \mathbb{R}$, the derivative $f'(x)$ at a point $x$ can be approximated using finite differences:  
\begin{itemize}
    \item \textit{Forward Difference:} 
    $$
    f'(x) \approx \frac{f(x+h) - f(x)}{h},
    $$
    where $h > 0$ is a small step size.
    \item \textit{Backward Difference:}
    $$
    f'(x) \approx \frac{f(x) - f(x-h)}{h}.
    $$
    \item \textit{Central Difference:}
    $$
    f'(x) \approx \frac{f(x+h) - f(x-h)}{2h}.
    $$
\end{itemize}

Central differences are generally more accurate, as they reduce the truncation error to $\mathcal{O}(h^2)$. \marginnote{The notation, $\mathcal{O}(h^2)$, is called `Big-O' notation, and it indicates that the leading term of the truncation error is proportional to $h^2$. This effectively means that the error increases quadratically as a function of the step size.}

Finite difference methods introduce truncation errors due to the approximation of the limit. The magnitude of the error depends on the choice of $h$.

\greyline

\paragraph*{The Gradient} The gradient is a linear functional assigning to each direction how much the function $f$ changes in that direction.

\begin{tcolorbox}[colback=gray!10, colframe=gray!40]
The {\em gradient} of $f$ is a vector-valued function ({\em vector field}) $\nabla f : \mathbb{R}^n \rightarrow \mathbb{R}^n$ satisfying $\langle \nabla f(x), d\rangle = \partial_d f(x)$ for all $x, d\in \mathbb{R}^n$. 
\end{tcolorbox}

We stress that vectors should be correctly treated as \textit{abstract objects} rather than their coordinates in some basis. However,  if one wishes to express the gradient w.r.t. to the standard basis of unit vectors $\{e_1, \hdots, e_n \}$ on $\mathbb{R}^n$, this is possible by applying $\langle \nabla f(x), e_i\rangle = \frac{\partial}{\partial x_i} f(x)$. This leads to the usual (somewhat primitive) way of thinking of the gradient as a vector of partial derivatives, 
$$
\nabla f(x) = \left (\frac{\partial}{\partial x_1} f(x), \hdots, \frac{\partial}{\partial x_n} f(x) \right ).
$$

Using the gradient, one can provide a linear approximation (first-order {\em Taylor expansion}) of $f$ around $x$,\marginnote{The Taylor series expansion provides a polynomial approximation of the smooth function $f$.}
$$
f(x+dx) = f(x) + \langle \nabla f(x), dx\rangle + \mathcal{O}(\|dx\|^2),
$$
where $dx$ is some infinitesimal displacement. Note the direct relation to numerical methods and the forward difference.

\greyline

\paragraph*{The Jacobian Matrix}

The Jacobian matrix generalizes the gradient to vector fields. 

\begin{tcolorbox}[colback=gray!10, colframe=gray!40]
For a vector-valued function $F: \mathbb{R}^n \to \mathbb{R}^m$, the \textit{Jacobian matrix} $J_F(x)$ at a point $x \in \mathbb{R}^n$ is defined as the matrix of all first-order partial derivatives of the components of $F$. That is,
$$
J_F(x) = \left[ \frac{\partial F_i}{\partial x_j} \right]_{i=1, \dots, m, j=1, \dots, n}=\begin{bmatrix}
\frac{\partial F_1}{\partial x_1} & \frac{\partial F_1}{\partial x_2} & \cdots & \frac{\partial F_1}{\partial x_n} \\
\frac{\partial F_2}{\partial x_1} & \frac{\partial F_2}{\partial x_2} & \cdots & \frac{\partial F_2}{\partial x_n} \\
\vdots & \vdots & \ddots & \vdots \\
\frac{\partial F_m}{\partial x_1} & \frac{\partial F_m}{\partial x_2} & \cdots & \frac{\partial F_m}{\partial x_n}
\end{bmatrix}.
$$
\end{tcolorbox}

Each element of the Jacobian represents how a single component of the vector field $F$ changes in response to a change in one of the coordinates of the domain. The Jacobian provides valuable information about the local behavior of the function, such as how the function stretches or compresses space.

\greylinelong

\subsection{Integrals}

\begin{tcolorbox}[colback=gray!10, colframe=gray!40]
The {\em integral} of a function $f$ over a domain $\Omega$ is a value that represents the total accumulation of $f$ across $\Omega$. For functions $f: \mathbb{R}^n \to \mathbb{R}$, the integral is formally defined as 
$$
\int_{\Omega} f(x) \, dV,
$$ 
where $dV$ denotes the infinitesimal volume element.
\end{tcolorbox}

Integration generalizes the notion of summation to continuous domains. For scalar functions $f$, the integral provides a measure of how $f$ `adds up' across the domain $\Omega$. For instance, in the case of $n=1$, integration corresponds to calculating the signed area under the curve $f(x)$ over an interval. In higher \marginnote{When the domain $\Omega$ is defined by bounds on individual coordinates, the multi-dimensional integral can be split into a series of one-dimensional integrals. This is known as Fubini's theorem.}dimensions, the infinitesimal volume element $dV$ depends on the coordinate system used. For Cartesian coordinates in $\mathbb{R}^n$, $dV = dx_1 dx_2 \cdots dx_n$. In polar, cylindrical, or spherical coordinates, $dV$ includes factors to account for the geometry of the domain.

\greyline

\paragraph*{Riemann Integral} 

The Riemann integral is one of the foundational approaches to defining integration. 

\begin{tcolorbox}[colback=gray!10, colframe=gray!40]
For a bounded function $f: [a, b] \to \mathbb{R}$, its \textit{Riemann integral} is defined as the limit of Riemann sums: \marginnote{If the function is not bounded or if it presents severe discontinuities, the Riemann integral fails. We say that such functions are not Riemann integrable. Alternatives like the Lebesgue integral can handle such cases.}
$$
\int_a^b f(x) \, dx = \lim_{n \to \infty} \sum_{i=1}^n f(x_i^*) \Delta x_i,
$$
where $[a, b]$ is divided into $n$ subintervals of width $\Delta x_i$, and $x_i^*$ is a chosen point in each subinterval. 
\end{tcolorbox}

This approach intuitively captures the idea of summing up small contributions $f(x_i^*) \Delta x_i$. Similar to the forward difference method for approximating derivatives, when the closed-form solution to an integral is unknown, the Riemann sum is often used as a numerical approximation in computational methods.

\greyline

\paragraph*{Line and Surface Integrals}

Integration extends beyond volumes to lower-dimensional objects, such as curves and surfaces. 

\begin{tcolorbox}[colback=gray!10, colframe=gray!40]
A \textit{line integral} accumulates a function $f$ along a curve $C$:
    $$
    \int_C f(x) \, ds,
    $$
where $ds$ is the infinitesimal arc length.
\end{tcolorbox}

\begin{tcolorbox}[colback=gray!10, colframe=gray!40]
A \textit{surface integral} accumulates a function $f$ on a $S$, with the infinitesimal area element $dA$:
    $$
    \int_S f(x) \, dA.
    $$
\end{tcolorbox}

\greyline

\paragraph*{Fundamental Theorem of Calculus (FTC)}

The Fundamental Theorem of Calculus bridges the concepts of integration and differentiation. 

\begin{tcolorbox}[colback=blue!10, colframe=blue!60]
\begin{theorem}[Fundamental Theorem of Calculus]
In one dimension, for a function $f$ with antiderivative $F$:
$$
\int_a^b f(x) \, dx = F(b) - F(a),
$$

where $b>a$.
\end{theorem}
\end{tcolorbox}

An anti-derivative of a function $f$ is a function $F$ such that $F'=f$. Note that a given function can have infinite many anti-derivatives. For instance, if $F'(x)=f(x)$ then $F(x)+C$ for any constant $C$ is also an anti-derivative of $f(x)$.

\greylinelong

\subsection{Divergence}

Let $F:\mathbb{R}^n \rightarrow \mathbb{R}^m$ be a smooth {\em vector field}, $F(x) = (F_1(x), \hdots, F_n(x))$. 

\begin{tcolorbox}[colback=gray!10, colframe=gray!40]
The {\em divergence } of $F$ is a scalar field $\mathrm{div}F: \mathbb{R}^n \rightarrow \mathbb{R}$, satisfying 
$$
\mathrm{div} F(x) = \sum_{i=1}^n \frac{\partial}{\partial x_i} F_i(x) \,\,\equiv \,\,\, \nabla \cdot F.
$$
\end{tcolorbox}

\noindent Thinking of $F(x)$ as a flow around $x$, the divergence can be given the interpretation of the density of an outward flux from an infinitesimal volume around $x$. 

\begin{tcolorbox}[colback=blue!10, colframe=blue!60]\marginnote{The unit normal vector $\hat{n}(x)$ is a vector of length 1 that is perpendicular to the tangent plane of the boundary $\partial \Omega$ at point $x$. Its direction is chosen conventionally to point outward from $\Omega$ unless stated otherwise. The boundary integral $\int_{\partial \Omega}$ represents integration over the boundary surface $\partial \Omega$. The scalar product $\langle F, \hat{n}\rangle$ measures how the vector field $F$ aligns with the normal direction, while $dS$ indicates the infinitesimal surface area element on $\partial \Omega$.}
\begin{theorem}[Gauss-(Ostrogradsky-Stokes) or simply Divergence theorem]
\label{theorem:divergence}
Let $\Omega \subseteq \mathbb{R}^n$ be a region in space with boundary $\partial \Omega$. Then, 
$$\int_\Omega \mathrm{div}F dV = \int_{\partial \Omega} \langle F, \hat{n}\rangle dS,  
$$
where $\hat{n}(x)$ denotes the unit normal vector to the boundary surface $\partial \Omega$ at point $x$ on thereon.
\end{theorem}
\end{tcolorbox}

Note that in the above theorem one assumes that $\Omega$ is a smooth region and likewise $F$ is a sufficiently smooth vector field (at least continuously differentiable).

The divergence theorem is a mathematical statement of the physical conservation law that, in the absence of the creation or destruction of matter, the density within a region of space can change only by having it flow into or away from the region through its boundary.

In a sense, the divergence does an operation `opposite' to that of the gradient; in fact, the two operators are adjoint w.r.t. the appropriate inner products defined on the spaces of scalar and vector fields: \marginnote{The negative sign in the adjoint relationship does not prevent them from being adjoint operators; however, we sometimes refer to such operators as skew-adjoint operators to distinguish them from the perhaps more standard positive case.}
$$
\langle \nabla f, F \rangle = -\langle f, \mathrm{div} F \rangle.
$$

More concretely, let $\Omega \subset \mathbb{R}^n$ be a bounded domain with smooth boundary $\partial \Omega$. Define inner products, for scalar fields $f, g \in C^\infty(\Omega)$:
$$\langle f, g \rangle_{L^2(\Omega)} = \int_\Omega fg \, dx,$$

and or vector fields $F, G \in [C^\infty(\Omega)]^n$:

$$\langle F, G \rangle_{L^2(\Omega)} = \int_\Omega F \cdot G \, dx.$$

The left side of the original expression expands as:
$$\langle \nabla f, F \rangle_{L^2(\Omega)} = \int_\Omega \nabla f \cdot F \, dx = \int_\Omega \sum_{i=1}^n \frac{\partial f}{\partial x_i}F_i \, dx $$

Let us apply integration by parts to each term in the summation above:
   
$$\int_\Omega \frac{\partial f}{\partial x_i}F_i \, dx =  \int_{\Omega} fF_i dx - \int_\Omega f\frac{\partial F_i}{\partial x_i} \, dx = \int_{\partial \Omega} fF_i \hat{n}_i \, dS - \int_\Omega f\frac{\partial F_i}{\partial x_i} \, dx,$$

where the boundary terms comes from the divergence theorem (Theorem~\ref{theorem:divergence}) and we transition from the volume element $dx$ to the surface element $dS$. Summing over $i$ from 1 to~$n$:
$$\langle \nabla f, F \rangle_{L^2(\Omega)} = \int_{\partial \Omega} f(F \cdot \hat{n}) \, dS - \int_\Omega f\sum_{i=1}^n \frac{\partial F_i}{\partial x_i} \, dx.$$

Since $\mathrm{div} F = \nabla \cdot F = \sum_{i=1}^n \frac{\partial F_i}{\partial x_i}$:
$$\langle \nabla f, F \rangle_{L^2(\Omega)} = \int_{\partial \Omega} f(F \cdot \hat{n}) \, dS - \int_\Omega f(\mathrm{div} F) \, dx.$$

The boundary term vanishes under any of these conditions:

\begin{itemize}
    \item Dirichlet boundary condition: $f|_{\partial \Omega} = 0$
    \item $F|_{\partial \Omega} = 0$
    \item Normal component vanishes: $F \cdot n|_{\partial \Omega} = 0$
    \item If $\Omega = \mathbb{R}^n$ and $F$ decays faster than $\|x\|^{-n}$ as $\|x\| \to \infty$
\end{itemize}

Adopting any of the above: 

$$\langle \nabla f, F \rangle_{L^2(\Omega)} = \int_{\partial \Omega} f(F \cdot \hat{n}) \, dS - \int_\Omega f(\mathrm{div} F) \, dx = 0 \, -  \int_\Omega f(\mathrm{div} F) = -\langle f, \mathrm{div} F \rangle_{L^2(\Omega)}.$$

\greylinelong

\subsection{Laplacian}

The Laplacian operator is a measure of how a function behaves locally in terms of its rate of change. 

\begin{tcolorbox}[colback=gray!10, colframe=gray!40]\marginnote{It is common to define the Laplacian as $-\mathrm{div} \nabla f$, to make it a positive-semidefinite operator.}
The Laplacian of a scalar field $f$ is given  by 
$$
\Delta f(x) = \mathrm{div} \nabla f. 
$$
\end{tcolorbox}

The quadratic functional $\langle f, \Delta f\rangle = \langle \nabla f, \nabla f \rangle $,  known in physics as the {\em Dirichlet energy}, is a measure of how variable the function $f$ is.

\begin{tcolorbox}[colback=blue!10, colframe=blue!60]
\begin{theorem} The Laplacian is rotation-invariant. 
\end{theorem}
\end{tcolorbox}

\begin{proof}
Write the Laplacian as the trace of the Hessian, $\Delta f(x) = \mathrm{tr}(\nabla^2 f(x))$. Note that when representing the Hessian as a matrix w.r.t. the standard basis, its diagonal contains second order derivatives $\frac{\partial^2 }{\partial x_i^2} f(x)$:
$$\nabla^2 f(x) =\begin{bmatrix}
\frac{\partial^2 f(x)}{\partial x_1^2} & \cdots & \frac{\partial^2 f(x)}{\partial x_1 \partial x_n} \\
\vdots & \ddots & \vdots \\
\frac{\partial^2 f(x)}{\partial x_n \partial x_1} & \cdots & \frac{\partial^2 f(x)}{\partial x_n^2}
\end{bmatrix}$$
Let $Ax$ be some transformation of coordinates. Then, applying the chain rule, we have
\begin{eqnarray*}
\nabla_x f(Ax) &=& A^\top \nabla_{Ax} f(Ax) \\
\nabla^2_x f(Ax) &=& A^\top \nabla^2_{Ax} f(Ax)  A. 
\end{eqnarray*}
Assuming $A$ is an orthogonal matrix ($A A^\top =  A^\top A = I$) and using matrix commutativity\marginnote{The trace of a product of matrices has the property $\text{tr}(X Y) = \text{tr}(Y X).$} under trace we get 
\begin{eqnarray*}
\Delta_x f(Ax) &=& \mathrm{tr}(A^\top \nabla^2_{Ax} f(Ax)  A) \\ 
&=& \mathrm{tr}( \nabla^2_{Ax} f(Ax)  A A^\top) \\
&=& \mathrm{tr}( \nabla^2_{Ax} f(Ax)) = \Delta_{Ax} f(Ax)
\end{eqnarray*}
\end{proof}

This invariance \marginnote{When transforming coordinates, a change of basis can be represented by multiplying by a matrix $A$. If $A$ is an orthogonal matrix, the transformation does not distort the geometry of the space, that is, distances and angles remain unchanged. This is a necessary condition for the invariance of the Laplacian under rotation.} suggests that the behavior of the Laplacian does not depend on the specific orientation of the coordinate system, but rather on the intrinsic geometry of the scalar field itself.

\greylinelong
\subsection[Gradient Descent Optimization in DL]{Gradient Descent Optimization in Deep Learning}
\label{subsec:GradientDescent}

In Deep Learning, gradients play a central role in training models, that is, in optimizing the parameters of artificial neural networks. Although we have not yet introduced artificial neural networks properly, we can think of them as mapping functions (vector fields) $F(x;w): \mathbb{R}^n \to \mathbb{R}^m$ parametrized by a set of weights (and biases) $w$.

\greyline
\paragraph*{Loss Functions as Scalar Fields} A loss function can be thought of as a scalar field, $\mathcal{L}(w)$, where $w$ represents the model parameters. The loss function assigns a scalar value that indicates how well the model performs. In \textit{supervised learning}, this is generally computed with respect to some reference \textit{ground truth} prediction

$$\mathcal{L}(w) = \mathcal{L}(F(x;w),\hat{y}),$$

where $\hat{y}$ is the ground truth (or the label), $F(x;w)$ represents the artificial neural network output (or prediction), and the loss is, for instance, the mean squared error loss in some regression tasks. Note, however, that the exact setup is task dependent, and more generally we can think of the loss function as returning a scalar based on the artificial neural network parameters $w$.

\begin{tcolorbox}[colback=gray!10, colframe=gray!40]
A loss function $\mathcal{L}(w)$ is a scalar field that assigns a scalar value to each set of parameters $w$, quantifying the model’s error.
\end{tcolorbox}

Just like in vector calculus, we are interested in how $\mathcal{L}(w)$ changes with respect to small changes in the parameters $w$. This is captured by the gradient of $\mathcal{L}(w)$, denoted as $\nabla \mathcal{L}(w)$. The gradient tells us the direction and rate at which the loss function increases most rapidly. By adjusting the parameters in the opposite direction of the gradient (steepest descent), we can minimize the loss. \marginnote{Loss plot credits to the research paper \textit{'Visualizing the Loss Landscape of Neural Nets'}.}

\begin{figure}[hbpt!]
\centering
\includegraphics[width=0.45\linewidth]{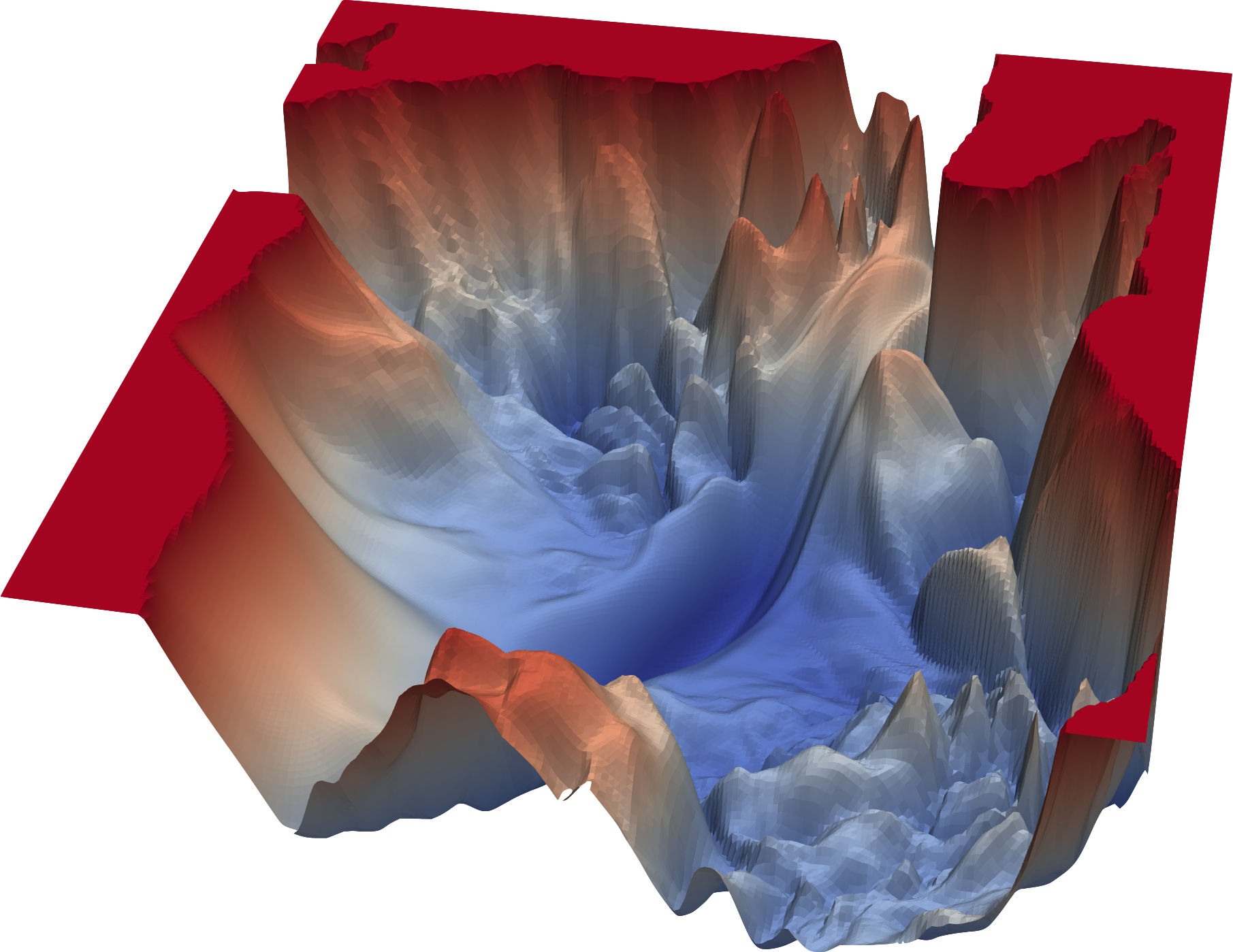}
    \caption{Example loss landscape visualization for a neural network.}
    \label{fig:loss-land}
\end{figure}

\greyline
\paragraph*{Gradient Descent Optimization} Gradient descent is the most common optimization method used in Deep Learning\marginnote{Typically stochastic gradient descent (SGD) is mentioned as the optimization technique of choice in most textbooks. However, in contemporary Deep Learning more modern variations of SGD are used, such as the AdamW optimizer.}.
\begin{tcolorbox}[colback=gray!10, colframe=gray!40]
\textit{Gradient descent} leverages the gradient $\nabla \mathcal{L}(w)$ of the loss function to adjust the parameters of a parametrized model in order to minimize the loss,

$$
w_{t+1} = w_t - \eta \nabla \mathcal{L}(w_t),
$$

where $w_t$ are the model parameters at iteration (or time step) $t$ , $\eta$ is the learning rate, a scalar that controls the step size, and $\nabla \mathcal{L}(w_t)$ is the gradient of the loss function with respect to the parameters at $w_t$.
\end{tcolorbox}

The gradient guides the model parameters toward a local minimum of the loss. Using this procedure we `translate the weights in space', from an initial random configuration to a suitable location that is able to model the data with low error. That is, the final weight configuration is a able to mimic the patterns present in the data.

\begin{tcolorbox}[colback=orange!20, colframe=orange!60]
\textbf{The Curse of Optimization.} Finding global optima of generic high
dimensional functions is NP-hard. Then, one may ask: How can we overcome this curse in optimization? In Geometric Deep Learning we argue that we can try to leverage the underlying low-dimensional structure of the input high-dimensional space. In particular, the geometric domain in which the signal lives provides new notions of regularity that can be exploited for more efficient learning.
\end{tcolorbox}

\greyline
\paragraph*{Backpropagation and the Chain Rule} The gradient of the loss function with respect to the model parameters is typically computed using \textit{backpropagation}. This method relies on the chain rule of calculus to propagate gradients through the network. 

Given a point $ x \in \mathbb{R}^n $, the composition of two vector fields $ f: \mathbb{R}^n \to \mathbb{R}^m $ and $ g: \mathbb{R}^m \to \mathbb{R}^p $ is written as $g \circ f(x) = g(f(x)),$ which represents the transformation of $ x $ through both functions $ f $ and $ g $.

\begin{tcolorbox}[colback=gray!10, colframe=gray!40]
The \textit{chain rule} states that given two vector fields  $ f: \mathbb{R}^n \to \mathbb{R}^m $ and $ g: \mathbb{R}^m \to \mathbb{R}^p $ and their respective Jacobian matrices $ J_f $ and $ J_g $, the derivative of their composition is given by the matrix product:

$$
\frac{d}{dx} \left( g \circ f(x) \right) = J_g(f(x)) \cdot J_f(x)
$$
\end{tcolorbox}

Artificial neural networks are composed of multiple layers, which can be understood in terms of function composition. The gradient of the loss function $\mathcal{L}$ with respect to each layer’s weights is computed iteratively:

$$
\nabla_{w^{(l)}} \mathcal{L} = J_L(a^{(L)}) \cdot J_{a^{(L)}}(a^{(L-1)}) \cdot ... \cdot J_{a^{(l+1)}}(w^{(l)})
$$
where $a^{(l)}$ is the activation (the output of an intermediate transformation) of the $l$-th layer. 

The Jacobian-based representation can handle cases where activations or transformations are vector-valued, which is typically the case in Deep Learning (technically we work with tensors which becomes even more complex). In the scalar or element-wise gradient context, we can rewrite the expression above as 

$$
\nabla_{w^{(l)}} \mathcal{L} = \frac{\partial \mathcal{L}}{\partial a^{(L)}} \cdot \frac{\partial a^{(L)}}{\partial a^{(L-1)}} \cdot \dots \cdot \frac{\partial a^{(l+1)}}{\partial w^{(l)}},
$$

which may be more accessible to readers less familiar with matrix calculus.

\begin{tcolorbox}[colback=gray!10, colframe=gray!40]
\textit{Backpropagation} uses the chain rule to compute gradients of the loss function with respect to each layer’s weights, which are then used to update the weights of artificial neural networks in an iterative fashion.
\end{tcolorbox}

\begin{tcolorbox}[colback=orange!20, colframe=orange!60]
\textbf{Vector Calculus and the Laplacian in Geometric Deep Learning.} Beyond other use cases of the gradient such as in gradient descent optimization, in Geometric Deep Learning, the Laplacian is often used to understand the smoothness of functions defined on graphs or manifolds. These structures, such as the vertices and edges of a graph, or the points on a surface, require modifications of traditional calculus tools to account for the inherent irregularities of the data. Hence, vector calculus is not only foundational in classical analysis but are also key components in the development of algorithms for learning over non-Euclidean data. 
\end{tcolorbox}

\clearpage

\section{Topological Foundations and Differential Geometry}

As we have seen so far, normed spaces add the ability to measure the length or magnitude of vectors. Metric spaces then enter the picture, with their additional structure allowing us to measure how far apart elements are, just as we measure distances in everyday space. And finally, inner products complete this geometric toolkit by defining angles between elements, enabling us to determine when vectors are perpendicular or parallel, for instance.

Students are often first introduced to this geometric foundations rather than topology~\cite{mendelson1975introduction,munkres}, because the former deal with tangible aspects of space, which are familiar in our everyday lives. However, this focus on geometry can sometimes overshadow topology, a more abstract field that underpins many concepts in geometry and other areas of mathematics. In essence, topology is concerned with connectivity and studies properties of space that remain unchanged under continuous deformations, such as stretching and bending. Topological spaces can later be augmented with additional structures to measure geometric quantities, such as a metric.

A solid understanding of topology provides deeper insights into the nature of space and is fundamental for grasping more advanced mathematical and scientific concepts. Therefore, in this section, we take a step back to introduce the reader to basic concepts in topology, with a particular focus on manifolds, which are central to many Geometric Deep Learning generalizations of traditional neural network models. We then complement this discussion by presenting key ideas from differential~\cite{doCarmo1976} and Riemannian geometry~\cite{Lee1997}. The text is kept succinct, with the goal of familiarizing the reader with the main concepts without going into excessive depth.\marginnote{Differential geometry and Riemannian geometry are closely related, but they are not the same. Differential geometry is the general study of geometry using calculus and linear algebra. It deals with smooth manifolds and smooth maps between them. On the other hand, Riemannian geometry is a special case of differential geometry where the manifold is equipped with a Riemannian metric.} 

\greylinelong

\subsection{A Brief Introduction to Topology}

\paragraph{Historical Context} The word \textit{topology} was coined by Johann Benedict Listing, a German mathematician, in his 1847 book \textit{Vorstudien zur Topologie}, although he used the word as early as 1836 in correspondence. The etymology of the word stems from the Greek `topos', meaning `location' or `place', and the suffix `-logy' for `study of'. Topology was initially conceived as a type of geometry that focused on properties preserved under much more flexible transformations than those allowed in Euclidean or other specific geometries. It is often called `rubber-sheet geometry' to illustrate this idea: one can stretch and deform the `rubber sheet', but you cannot tear it or glue parts together. The French Henri Poincaré, with his \textit{Analysis Situs} series of papers starting in 1895, is largely credited with establishing topology as a coherent and independent field. Indeed, nowadays geometry and topology are considered two separate branches of mathematics, concerned with measurement and connectedness, respectively, as we have repeatedly emphasized in this text.

\greyline

\paragraph{Sets as a Collection of Objects with no Connectedness} As previously discussed in Section~\ref{Algebraic Structures and Mathematics before Numbers}, a set is a collection of distinct elements and has no structure beyond membership. For example, the set of points in the plane \( \mathbb{R}^2 \) can be written using the set builder notation as follows:
\[
\mathbb{R}^2 = \{(x, y) \mid x, y \in \mathbb{R}\}.
\]
This is simply a collection of points. Thus, there is no notion of `closeness' or `nearness' between points: the points are not connected in any way. For instance, the point $(0,0)$ is neither closer to $(0,1)$ nor to $(10,10)$ because the elements of the set are considered unordered and unrelated beyond membership.

When we introduce a \textit{structure} to this set, such as \textit{connectedness} or \textit{topology}, we begin to impose rules on how the points are related. For example, we can define which sets of points are considered `close' to each other or which subsets of \( \mathbb{R}^2 \) are `open'. This, in turn, leads to a concept of continuous connection among points.

\greyline

\paragraph{Open Intervals and Open Sets} In the context of the real line \( \mathbb{R} \), an \textit{open interval} is a set of points that does not include its boundary points. For example, the open interval \( (a, b) \) is the set of points \( x \) such that:
\[
a < x < b.
\]
This interval contains all points between \( a \) and \( b \), but does not include \( a \) and \( b \) themselves.

In a more general setting, a set \( U \) is called \textit{open} if it contains a `neighborhood' around each of its points. This means that for every point \( x \in U \), there is a small region around \( x \) that is entirely contained within \( U \).

\begin{figure}[hbpt!]
\centering
\includegraphics[width=0.5\linewidth]{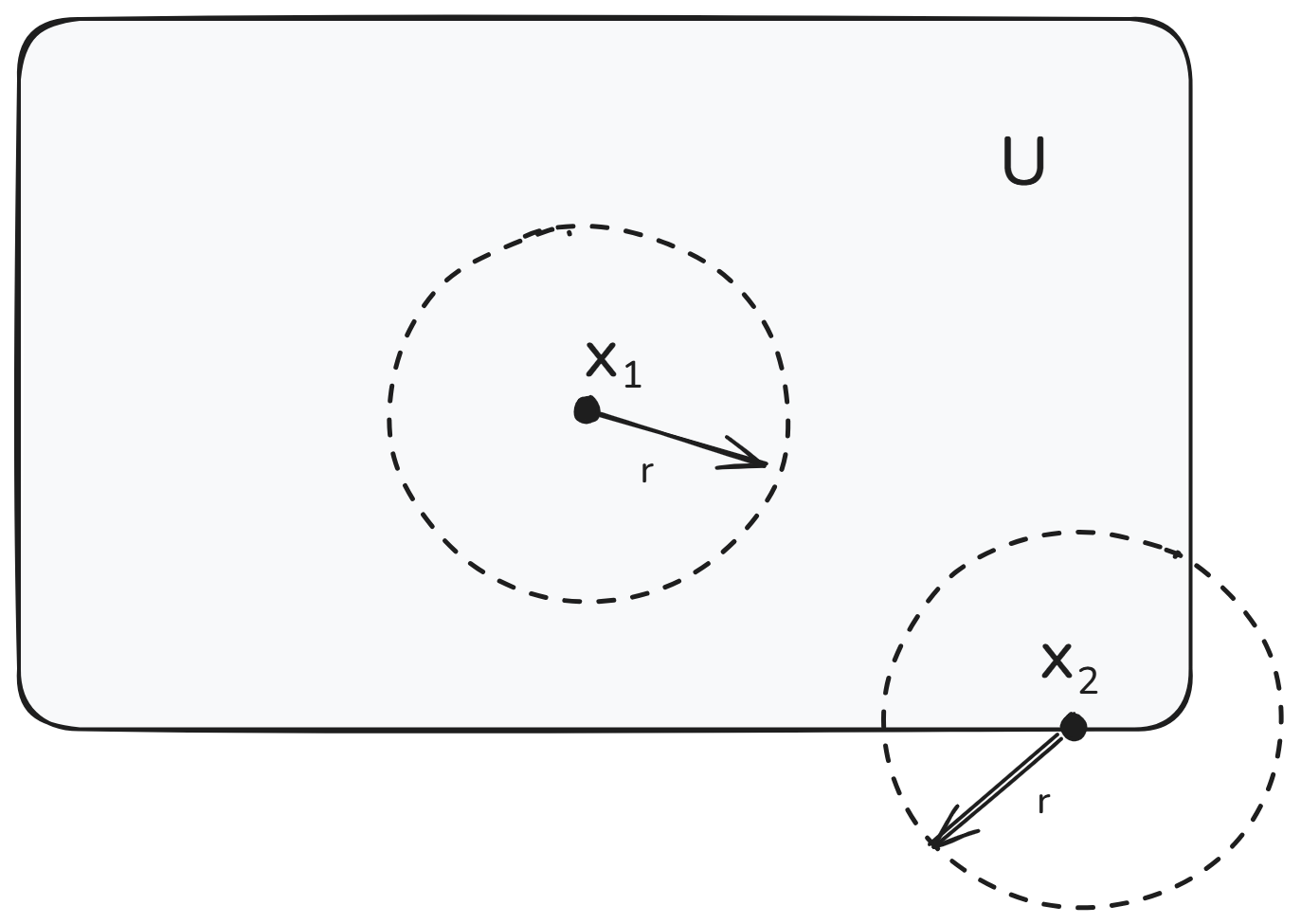}
    \caption{Point $x_1$ has an open neighborhood fully contained in $U$, while point $x_2$, located on the boundary, does not.}
    \label{fig:openset_yes_no}
\end{figure}

In the context of metric spaces, this is formalized as follows:

\begin{tcolorbox}[colback=gray!10, colframe=gray!40]
Let \( (X, d) \) be a metric space, where \( d \) is the distance function. A subset \( U \subseteq X \) is \textit{open} if, for every point \( x \in U \), there exists a radius \( r > 0 \) such that the \textit{open ball} \( B(x, r) = \{y \in X \mid d(x, y) < r\} \) is entirely contained within \( U \).
\end{tcolorbox}

Although the above definition is perhaps intuitive, it relies on a distance function. Actually, open sets can also be defined without relying on a metric space, and purely in terms of set theory, as we will see next.

\greyline

\paragraph{Topological Spaces} The concept of open sets can be generalized in the context of topological spaces. A topological space is defined as a \textit{set} \( X \) together with a collection of subsets \( \mathcal{T} \) (called \textit{open sets}) that satisfy certain properties. These properties ensure that the notion of `openness' is well-behaved.

\begin{tcolorbox}[colback=gray!10, colframe=gray!40]
Let \( X \) be a set, and \( \mathcal{T} \subseteq \mathcal{P}(X) \) the power set of \( X \). Then \( \mathcal{T} \) is a topology on \( X \) if:\marginnote{The symbols \( A \cup B \) and \( A \cap B \) refer specifically to the union and intersection of two sets, \( A \) and \( B \). In contrast, \( \bigcup_{\alpha \in A} U_\alpha \) and \( \bigcap_{\alpha \in A} U_\alpha \) are more general notations used to describe the union or intersection of a collection of sets \( \{U_\alpha\}_{\alpha \in A} \), where the index \( \alpha \) ranges over some set \( A \). Also, note that the notation in the definition differs to highlight the axioms: any union $\bigcup_{\alpha \in A}$ of open sets is open, but only finite intersections $\bigcap_{i=1}^n$ are required to be open. The indices reflect this arbitrary vs. finite condition.}

\begin{itemize}
    \item \( \emptyset, X \in \mathcal{T} \),
    \item \( \bigcup_{\alpha \in A} U_\alpha \in \mathcal{T}, \text{ for any collection } \{U_\alpha\}_{\alpha \in A} \subseteq \mathcal{T} \),
    \item \( \bigcap_{i=1}^n U_i \in \mathcal{T}, \text{ for any finite collection } \{U_i\}_{i=1}^n \subseteq \mathcal{T} \). 
\end{itemize}
\end{tcolorbox}

These conditions specify the following: the empty set and the entire set \( X \) must be included in \( \mathcal{T} \), arbitrary unions of open sets must be open, and finite intersections of open sets must be open. Note that $\mathcal{T}$ is a set of subsets.

\begin{tcolorbox}[colback=gray!10, colframe=gray!40]
The pair \( (X, \mathcal{T}) \) is called a \textit{topological space}. Elements of \( X \) are referred to as \textit{points}, and elements of \( \mathcal{T} \) are called \textit{open sets}.
\end{tcolorbox} 

\begin{tcolorbox}[colback=gray!10, colframe=gray!40]
A subset \( U \subseteq X \) is called \textit{open} if \( U \in \mathcal{T} \).
\end{tcolorbox}

Open sets are a generalization of intervals in \( \mathbb{R} \), which are open in the sense that they do not include their boundary points. Metric spaces are specific examples of topological spaces, and, similarly, open balls in a metric space are examples of open sets.

\greyline
\paragraph*{Examples of Topological Spaces}

\begin{itemize}
    \item \textit{Euclidean Topology:} For \( X = \mathbb{R}^n \), the standard topology is generated by open balls. An open ball in \( \mathbb{R}^n \) centered at \( x \in \mathbb{R}^n \) with radius \( r > 0 \) is defined as
    \[
    B(x, r) = \{ y \in \mathbb{R}^n : \|x - y\| < r \}.
    \]
    The topology \( \mathcal{T} \) in this case is the collection of all open sets that can be expressed as arbitrary unions of open balls. That is, 
    \[
    \mathcal{T} = \left\{ U \subseteq \mathbb{R}^n : U = \bigcup_{\alpha \in A} B(x_\alpha, r_\alpha) \text{ for some index set } A \right\},
    \]
    where each \( B(x_\alpha, r_\alpha) = B_{\alpha} \) is an open ball.

    \begin{figure}[hbpt!]
    \centering
    \includegraphics[width=0.65\linewidth]{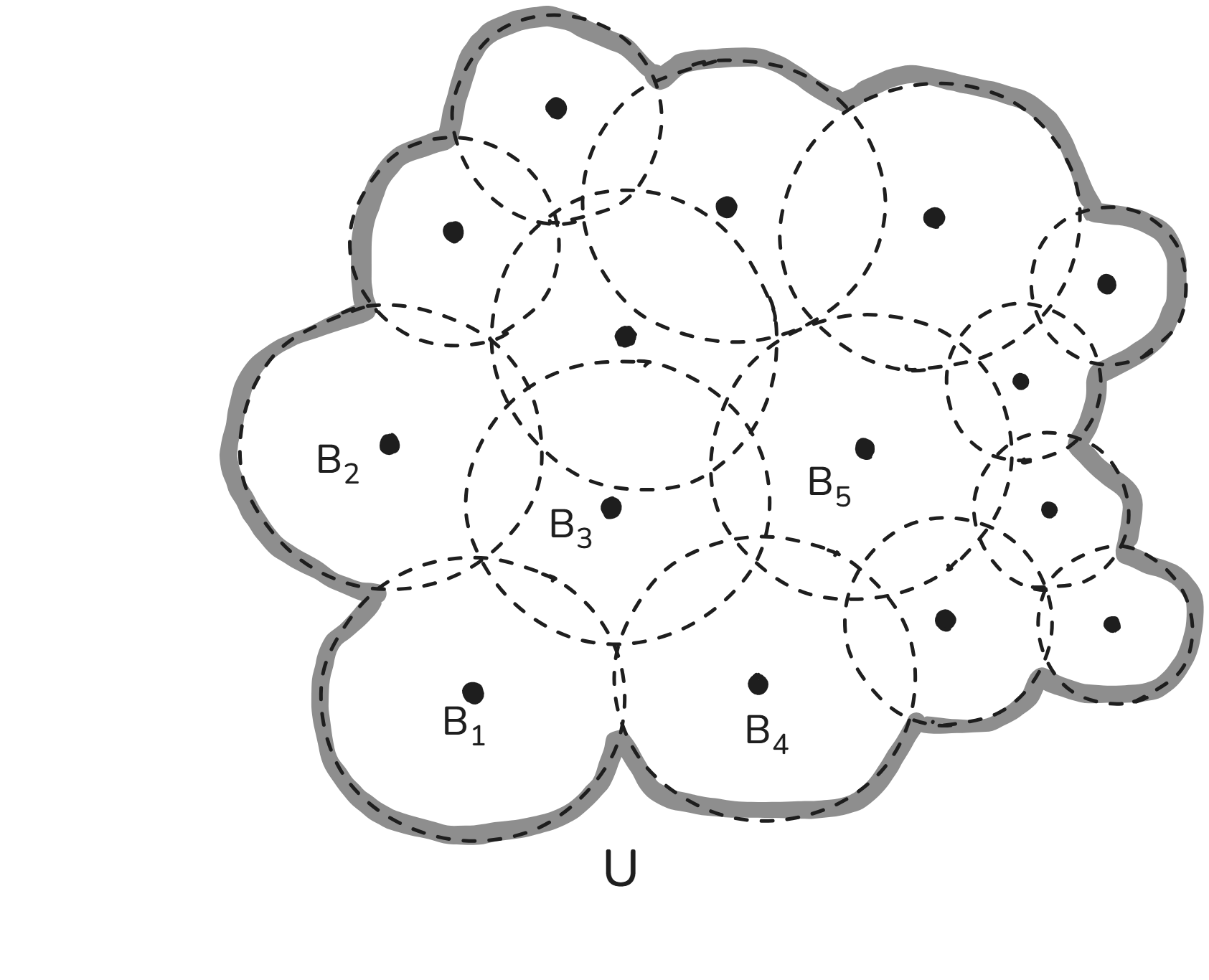}
        \caption{Illustration of an open set $U$ defined as the union of open balls, $U = B_1 \cup B_2 \cup B_2 \cup B_3 \cup B_4 \cup B_5 \cup \dots$. Each dashed circle represents an open ball $B_\alpha$, demonstrating how open sets in Euclidean topology are constructed.}
        \label{fig:euclidean_topology_u}
\end{figure}
    
    \item \textit{Discrete Topology:} In the discrete topology, every subset of \( X \) is open. Therefore, for any set \( X \), the topology \( \mathcal{T} \) is the power set of \( X \), i.e., 
    \[
    \mathcal{T} = \mathcal{P}(X) = \{ U \subseteq X : U \text{ is a subset of } X \}.
    \]
    
    \item \textit{Trivial Topology:} In the trivial topology, only the empty set \( \emptyset \) and the entire set \( X \) are open. Therefore, the topology \( \mathcal{T} \) is 
    \[
    \mathcal{T} = \{\emptyset, X\}.
    \]
\end{itemize}

The discrete topology is the finest topology because every subset of the space is an open set, making it the topology with the most open sets. In contrast, the trivial topology is the coarsest possible topology, as it contains the fewest open sets.

\greylinelong

\subsection{Topological Equivalences}

\begin{tcolorbox}[colback=gray!10, colframe=gray!40]
\textit{Topology} studies properties of spaces that are invariant under any continuous deformation.
\end{tcolorbox}

\paragraph{Continuity} Continuous maps between topological spaces do not `break' the space, meaning that small changes in the input correspond to small changes in the output, without any sudden jumps or gaps. In other words, the map allows the space to be deformed without tearing it and it preserves the structure of the space, enabling smooth transitions from one point to another. \marginnote{This definition of continuity does not require the notion of limits, as in the classical sense, but instead relies purely on the topological structure of the spaces involved.}

\begin{tcolorbox}[colback=gray!10, colframe=gray!40]
A map \( F: X \to Y \) between topological spaces is continuous if for every open set \( U \in \mathcal{T}_Y \), the preimage \( F^{-1}(U) \) is an open set in \( X \), i.e., \( F^{-1}(U) \in \mathcal{T}_X \).
\end{tcolorbox}

\greyline

\paragraph{Homeomorphisms and Homotopy} A \textit{homeomorphism} is a special type of continuous map that has a continuous inverse. \marginnote{It is quite common to confuse homeomorphisms with homomorphisms. A homomorphism is a structure-preserving map between two algebraic structures of the same type, as we saw earlier for groups. In contrast, a homeomorphism is a bijective map between two topological spaces that is continuous and has a continuous inverse. In short, homomorphisms pertain to algebra, while homeomorphisms arise in the context of topology.}

\begin{tcolorbox}[colback=gray!10, colframe=gray!40]
A map \( F: X \to Y \) is a \textit{homeomorphism} if it is bijective, continuous, and its inverse \( F^{-1}: Y \to X \) is also continuous.
\end{tcolorbox}

When such a map between two topological spaces exists, we say that \( X \) and \( Y \) are \textit{homeomorphic}, meaning they are topologically equivalent. For example, the surface of a sphere and that of a cube are homeomorphic, as one can be continuously deformed into the other without tearing or gluing. Note that a homeomorphism is a strong equivalence and denotes that there is a one-to-one correspondence between points in the spaces.

\begin{figure}[hbpt!]
\centering
        \includegraphics[width=\linewidth]{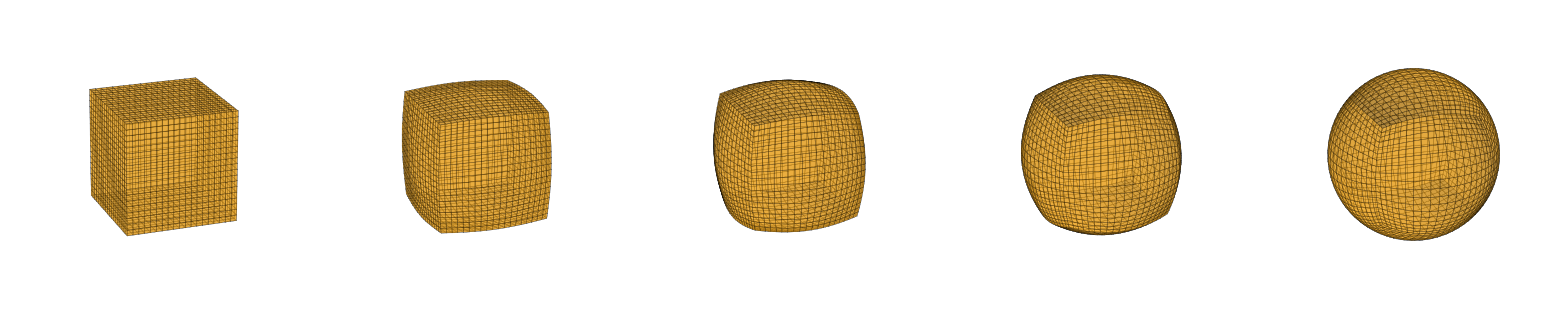}
    \caption{The cube and the sphere are homeomorphic: they are both simply connected (no holes) and can be continuously deformed into each other.}
    \label{fig:cube_sphere_homotopy}
\end{figure}

On the other hand, two spaces  are \textit{homotopic} (or \textit{homotopy equivalent}) if one can be continuously deformed into the other through a process called \textit{homotopy}. This is a weaker equivalence than homeomorphism since it allows for more general deformations such as collapsing or stretching parts of the space.\marginnote{The collapsing of a circle to a point is an example of a homotopy, not a homeomorphism, precisely because the inverse operation is not continuous. While the forward map (circle to point) can be considered continuous, if one were to start from a single point and try to map it back to a circle, it would require `expanding' that single point into an entire circle. A single point has dimension 0, while a circle has dimension 1. A continuous inverse would imply that a point is topologically equivalent to a circle, which is not true.} For example, a circle and a point are homotopy equivalent because the circle can be continuously shrunk to a single point.

\begin{figure}[hbpt!]
\centering
\includegraphics[width=\linewidth]{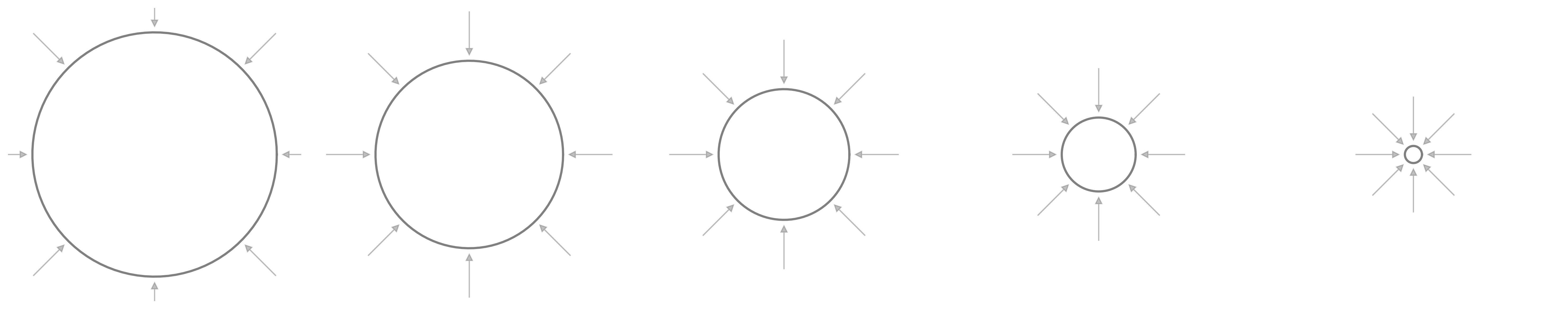}
    \caption{Visualization of circle shrinking into a point. The map between them is surjective, but not injective since all points on the circle are mapped (or collapsed) into the same single point.}
    \label{fig:circle_to_point}
\end{figure}

Discrete geometric representations, such as meshes or graphs, can also approximate topological features like homotopy, allowing us to reason about how shapes deform, connect, or contain loops, even in combinatorial settings. We will look at discrete representations later, in Section~\ref{Graph Theory}.

\greyline

\marginnote{A polyhedron (plural: polyhedra) is a three-dimensional solid whose boundary consists of polygons.
\includegraphics[width=0.9\linewidth]{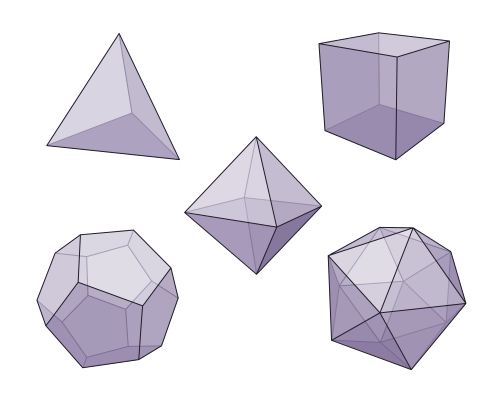}
For instance, the Platonic solids displayed above are a special, highly symmetric class of convex polyhedra characterized by faces that are all congruent regular polygons, with the same number of faces meeting at each vertex. All of them have an Euler characteristic of 2. Interestingly, they are named `Platonic' after the ancient Greek philosopher Plato (despite also being studied by Theaetetus and Euclid), due to his role in associating them with the classical elements of fire, earth, air, and water in his cosmological theories.}\paragraph{Euler Characteristic} The Euler characteristic is a topological invariant that assigns a numerical value to a topological space. It is defined for a variety of spaces, both continuous (like surfaces) and discrete (like polyhedra), and remains unchanged under homeomorphisms (but not necessarily under homotopy equivalence). That is, if two spaces are topologically equivalent, they share the same Euler characteristic, regardless of differences in their geometric shape or size. For a closed surface (compact surfaces without boundary), the Euler characteristic can be computed using the \textit{genus}, which represents the number of `holes' (or `handles') in the surface. For example, a sphere and a cube both have an Euler characteristic of 2, even though their geometric structures are quite different. On the other hand, a sphere and a point are homotopy equivalent (in the weak sense), but their Euler characteristics are different (2 and 1, respectively). While this text does not delve into the detailed calculation of the Euler characteristic (often done via homology), it is worth highlighting that there are methods for quantifying the equivalence of spaces based solely on their connectivity, entirely disregarding their geometric details.

\greylinelong

\subsection{Manifolds and Differential Geometry}
\label{sec:Manifolds and Differential Geometry}

Manifolds are mathematical objects used to describe and generalize to spaces that may not have a simple, flat, Euclidean structure. Indeed, many natural phenomena occur in spaces (or \textit{domains}) that are curved.

\greyline

\paragraph{Non-Euclidean Geometry and Historical Background}\marginnote{Euclid's monopoly came to an end in the 19th century, with a remarkable burst of creativity that made geometry arguably the most exciting field of mathematics, primarily through the work of pioneers like Gauss, Bolyai, Lobachevsky, Riemann, and Beltrami. However, one of the first attempts at questioning Euclid's fifth postulates dates back to the Italian mathematician Girolamo Saccheri (1667-1733) in his work \textit{Euclides ab omni naevo vindicatus}.

\includegraphics[width=1\linewidth]{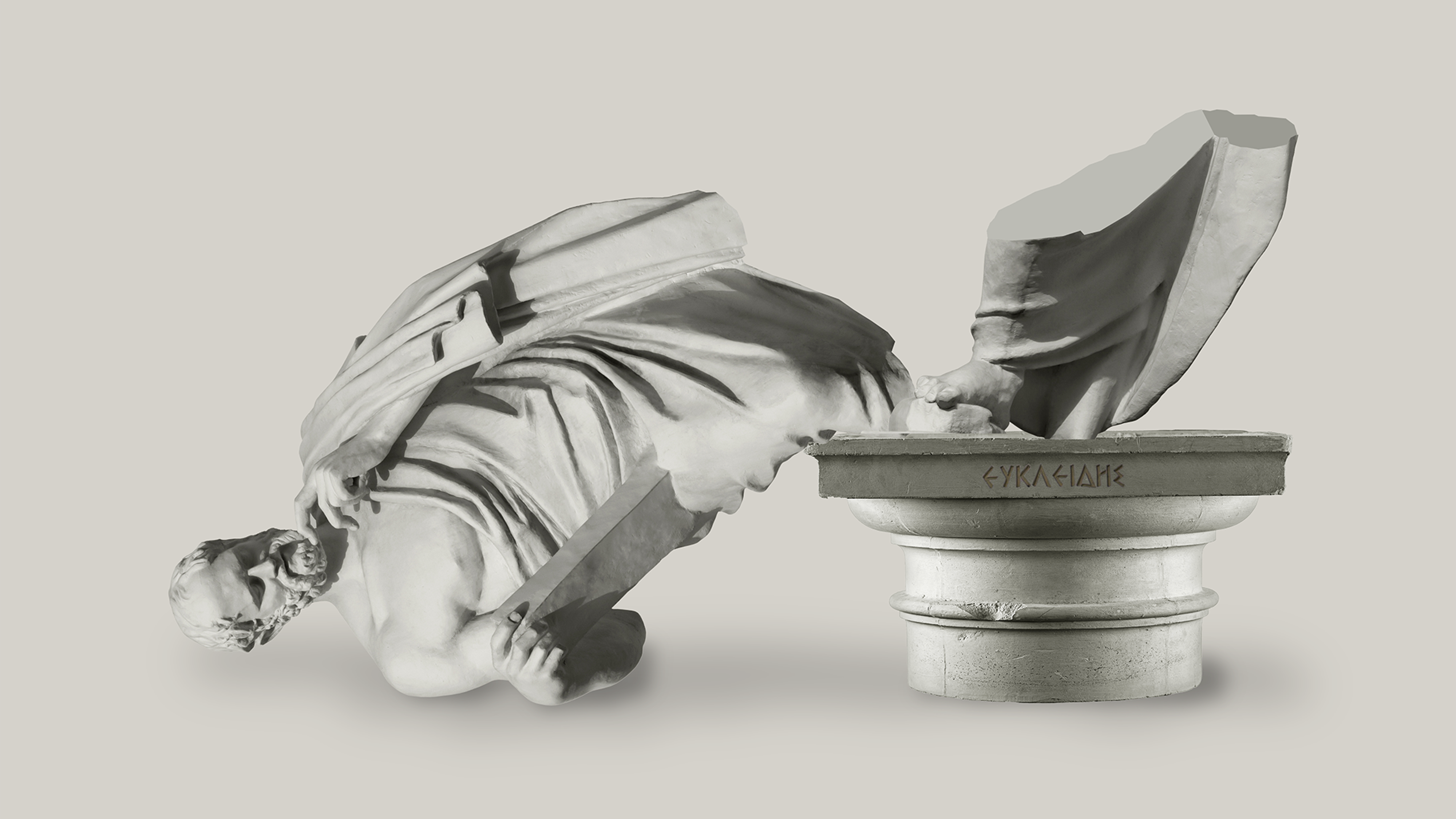}
} Since Euclid of Alexandria (c.~300~BC) stated the fifth postulate—later renamed the parallel postulate or axiom of parallels—in his famous work \textit{Elements}, it was accepted for two thousand years that \textit{through a point exterior to a given line, one and only one parallel line could be drawn}, and that no logically consistent alternative to his geometric framework could exist. After numerous failed attempts at deriving this postulate from the previous four, mathematicians started exploring geometries for which the fifth postulate did not hold. They found that it was possible to construct logically consistent frameworks that did not satisfy the postulate, giving rise to, for instance, elliptical (or spherical; the differences are subtle and outside the scope of this text) and hyperbolic geometry. In these frameworks, our common notion of a line is replaced by the shortest path between two points while remaining on the surface of the space at hand: the geodesic. The aforementioned geometries are characterized by their geodesic dispersion: in elliptical geometry, initially parallel geodesics eventually converge, whereas in hyperbolic geometry, they diverge exponentially, unlike in Euclidean geometry where they remain equidistant (the space is flat).

But these were not the only examples. For instance, projective geometry, which formalizes the principles of perspective projection, notably treating parallel lines as meeting at `points at infinity' and focusing on properties invariant under projection, was inspired by the arts.\marginnote{Renaissance artists, such as the Florentine Leonardo da Vinci and the German Albrecht Dürer, sought techniques to represent the three-dimensional world realistically on a two-dimensional surface like the canvas. Dürer, in particular, not only produced a large body of paintings but also authored written treatises on geometry related to this challenge. Below we display da Vinci's \textit{The Last Supper}, a classic example.

\includegraphics[width=1\linewidth]{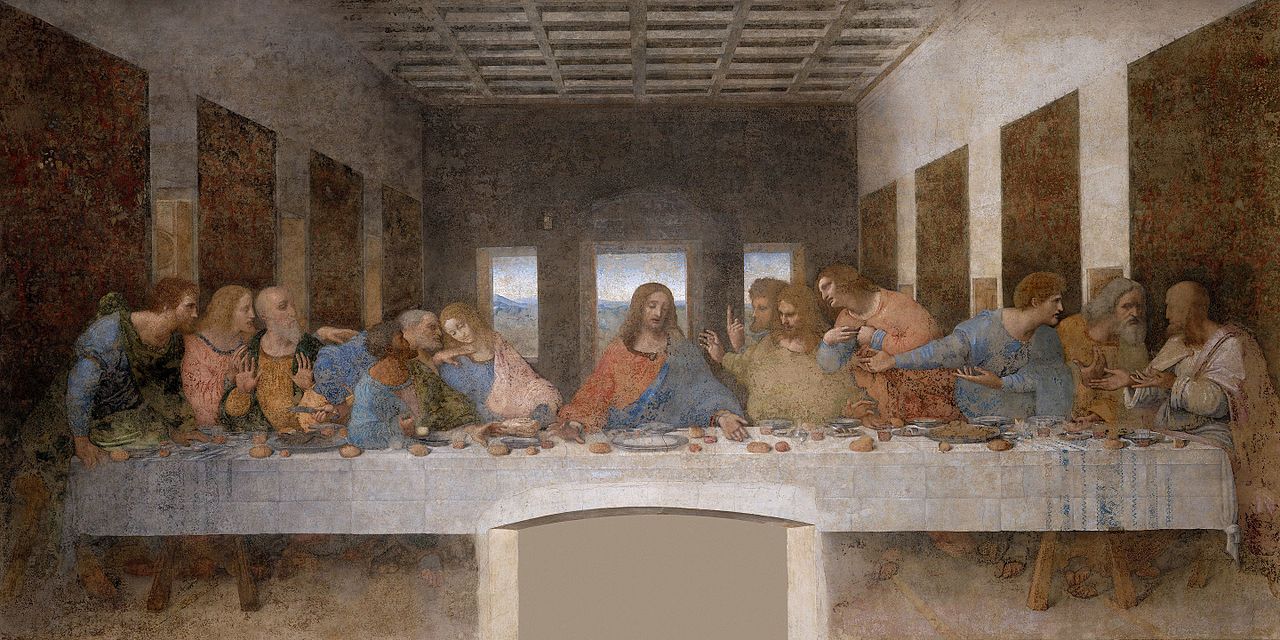}} The emergence of these varied and equally consistent geometric systems prompted a fundamental question: what truly defines `geometry'? It was not until Felix Klein (aided by insights from his discussions with Sophus Lie) proposed a unifying framework in his Erlangen Program (1872) that geometry came to be understood not merely by its objects (points, lines) but as the study of properties that remain invariant under a specified group of transformations.

\begin{figure}[hbpt!]
    \centering
    \includegraphics[width=0.5\linewidth]{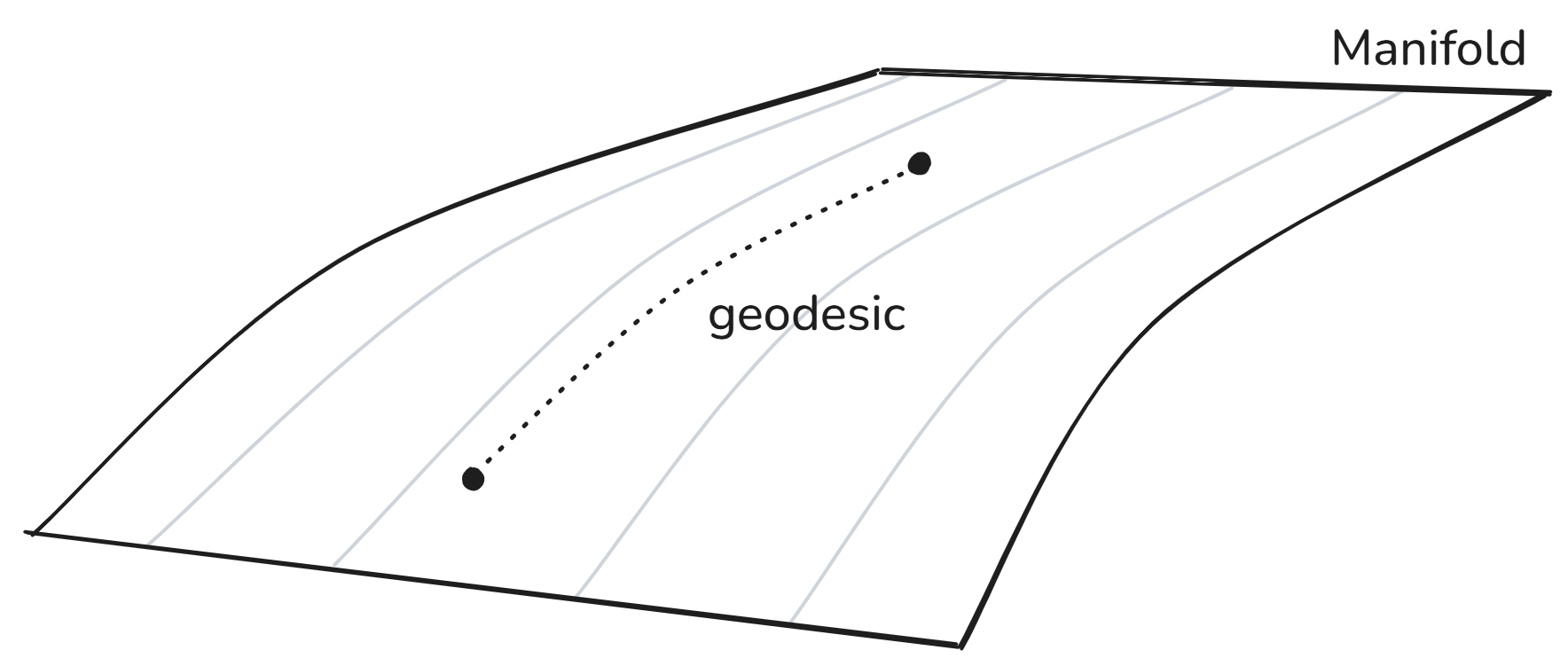}
        \caption{The geodesic is the shortest path between the two points, while staying on the curved 2-dimensional surface: it is not a straight line.}
        \label{fig:geodesic_manifold}
\end{figure}

\greyline

\paragraph{Topological Manifolds}\marginnote{Note that a `basic' manifold (topological or even smooth) does not inherently come equipped with a way to measure distances, angles, or curvature. Here, we are primarily concerned about the structure and connectivity of the space and its local resemblance to Euclidean (flat) space. On the other hand, geometry is about the study of measurement and properties on the space.} To understand manifolds, we begin with the simplest notion of a topological manifold, which captures the idea of spaces that locally resemble Euclidean space. From there, we can progressively add more structure to these spaces, eventually obtaining smooth manifolds, which allow for calculus and differential geometry, and Riemannian manifolds, which introduce a way to measure distances and angles.

A \textit{manifold} is a topological space that locally resembles Euclidean space.

\begin{tcolorbox}[colback=gray!10, colframe=gray!40]
A topological space \( \mathcal{M} \) is an \( n \)-dimensional \textit{(topological) manifold} if for every point \( p \in \mathcal{M} \), there exists an open neighborhood \( U \subseteq \mathcal{M} \) and a homeomorphism \( \varphi: U \to \mathbb{R}^n \).
\end{tcolorbox}

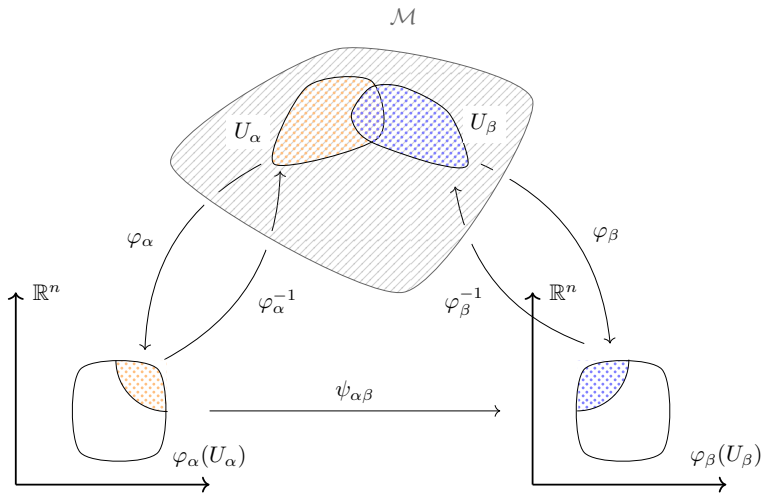
\begin{figure}[hbpt!]
    \centering
\scalebox{0.85}{\begin{tikzpicture}

    \path[->] (0.8, 0) edge [bend right] node[left, xshift=-2mm] {$\varphi_\alpha$} (-1, -2.9);
    \draw[white,fill=white] (0.06,-0.57) circle (.15cm);
    \path[->] (-0.7, -3.05) edge [bend right] node [right, yshift=-3mm] {$\varphi^{-1}_\alpha$} (1.093, -0.11);
    \draw[white, fill=white] (0.95,-1.2) circle (.15cm);

    \path[->] (5.8, -2.8) edge [bend left] node[midway, xshift=-5mm, yshift=-3mm] {$\varphi^{-1}_\beta$} (3.8, -0.35);
    \draw[white, fill=white] (4,-1.1) circle (.15cm);
    \path[->] (4.2, 0) edge [bend left] node[right, xshift=2mm] {$\varphi_\beta$} (6.2, -2.8);
    \draw[white, fill=white] (4.54,-0.12) circle (.15cm);

    \draw[smooth cycle, tension=0.4, fill=white, pattern color=gray, pattern=north east lines, opacity=0.6] plot coordinates{(2,1.8) (-0.6,0) (3,-2) (5,1)} node at (3,2.3) {$\mathcal{M}$};


    \draw[smooth cycle, pattern color=orange!50, pattern=crosshatch dots] 
        plot coordinates {(1,0) (1.5, 1.2) (2.5,1.3) (2.6, 0.4)} 
        node [label={[label distance=-0.3cm, xshift=-2cm, fill=white]:$U_\alpha$}] {};
    \draw[smooth cycle, pattern color=blue!50, pattern=crosshatch dots] 
        plot coordinates {(4, 0) (3.7, 0.8) (3.0, 1.2) (2.5, 1.2) (2.2, 0.8) (2.3, 0.5) (2.6, 0.3) (3.5, 0.0)} 
        node [label={[label distance=-0.8cm, xshift=.75cm, yshift=1cm, fill=white]:$U_\beta$}] {};

    \draw[thick, ->] (-3,-5) -- (0, -5) node [label=above:$\varphi_\alpha(U_\alpha)$] {};
    \draw[thick, ->] (-3,-5) -- (-3, -2) node [label=right:$\mathbb{R}^n$] {};

    \draw[->] (0, -3.85) -- node[midway, above]{$\psi_{\alpha\beta}$} (4.5, -3.85);

    \draw[thick, ->] (5, -5) -- (8, -5) node [label=above:$\varphi_\beta(U_\beta)$] {};
    \draw[thick, ->] (5, -5) -- (5, -2) node [label=right:$\mathbb{R}^n$] {};

    \draw[white, pattern color=orange!50, pattern=crosshatch dots] (-0.67, -3.06) -- +(180:0.8) arc (180:270:0.8);
    \fill[even odd rule, white] [smooth cycle] plot coordinates{(-2, -4.5) (-2, -3.2) (-0.8, -3.2) (-0.8, -4.5)} (-0.67, -3.06) -- +(180:0.8) arc (180:270:0.8);
    \draw[smooth cycle] plot coordinates{(-2, -4.5) (-2, -3.2) (-0.8, -3.2) (-0.8, -4.5)};
    \draw (-1.45, -3.06) arc (180:270:0.8);

    \draw[white, pattern color=blue!50, pattern=crosshatch dots] (5.7, -3.06) -- +(-90:0.8) arc (-90:0:0.8);
    \fill[even odd rule, white] [smooth cycle] plot coordinates{(7, -4.5) (7, -3.2) (5.8, -3.2) (5.8, -4.5)} (5.7, -3.06) -- +(-90:0.8) arc (-90:0:0.8);
    \draw[smooth cycle] plot coordinates{(7, -4.5) (7, -3.2) (5.8, -3.2) (5.8, -4.5)};
    \draw (5.69, -3.85) arc (-90:0:0.8);

\end{tikzpicture}}
    \caption{Illustration of a manifold $\mathcal{M}$ with overlapping open subsets $U_\alpha$ and $U_\beta$. Each has a corresponding chart, represented by a homeomorphism $\varphi_\alpha$ and $\varphi_\beta$, mapping it onto an open subset of the Euclidean space, $\mathbb{R}^n$. The transition map $\psi_{\alpha\beta} = \varphi_\beta \circ \varphi_\alpha^{-1}$ describes how these charts relate to each other on their overlapping regions.}
    \label{fig:manifold}
\end{figure}

\marginnote{In relativity, the manifold used to model the universe is a 4-dimensional Lorentzian manifold, which is commonly referred to as spacetime.} Manifolds can be classified based on their dimensionality, such as curves (1-dimensional manifolds), surfaces (2-dimensional manifolds), and higher-dimensional manifolds. They are the central objects in differential geometry and are fundamental in the study of geometry and physics, particularly in general relativity.

The local homeomorphisms between a manifold and Euclidean space are called \textit{charts}. A collection of charts that cover the entire manifold is called an \textit{atlas}.\marginnote{The term `atlas' in mathematics draws an analogy to a collection of maps used in geography. Just as a geographic atlas contains individual maps that collectively describe different regions of the Earth's surface, an atlas on a manifold consists of charts that collectively describe the manifold's structure.

\includegraphics[width=1\linewidth]{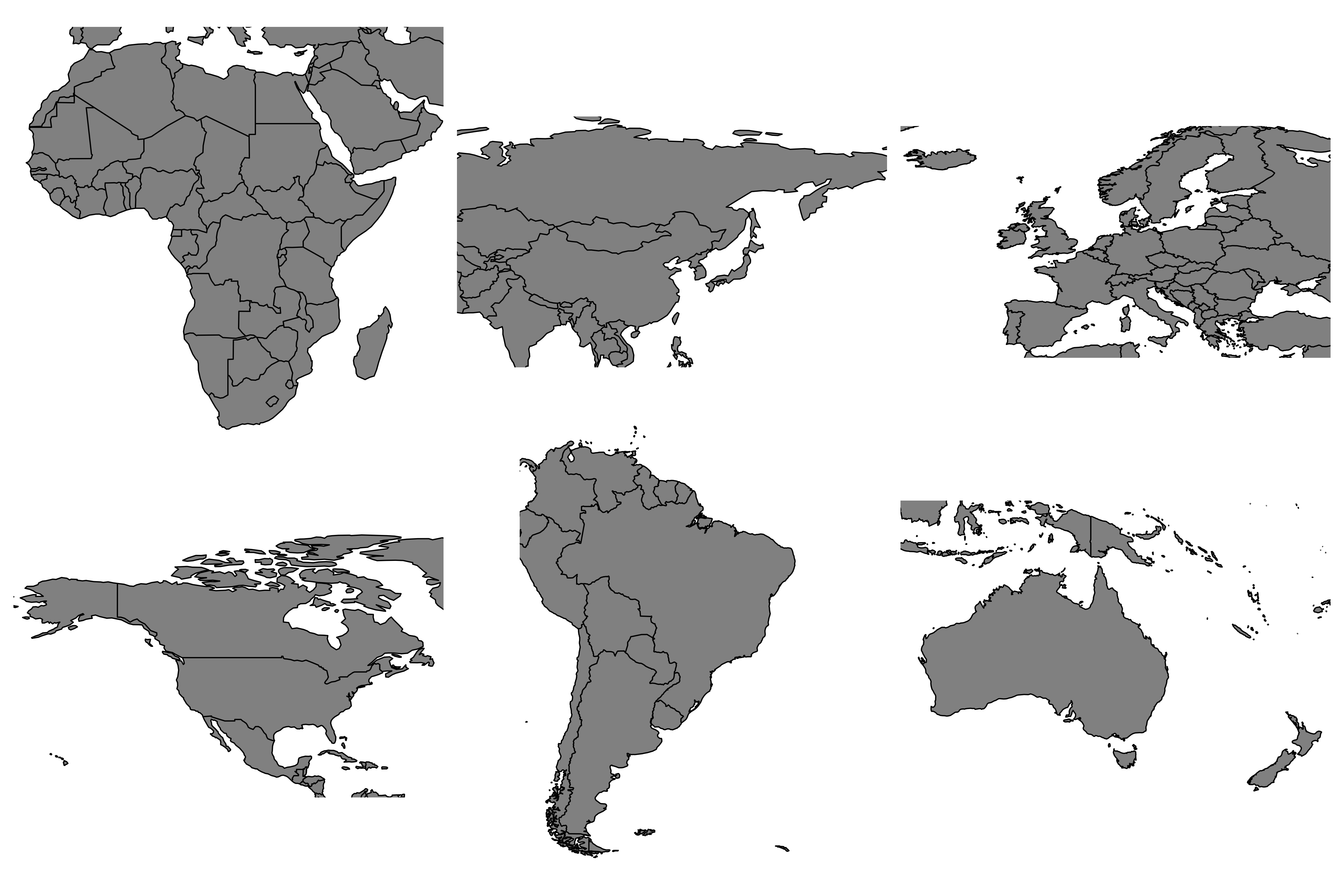}}

\begin{tcolorbox}[colback=gray!10, colframe=gray!40]
An \textit{atlas} for a manifold \( \mathcal{M} \) is a collection of \textit{charts} \( \{(U_\alpha, \varphi_\alpha)\} \), where \( U_\alpha \) is an open subset of \( \mathcal{M} \) and \( \varphi_\alpha: U_\alpha \to \mathbb{R}^n \) is a homeomorphism. The charts must be compatible, meaning that the transition maps \( \psi_{\alpha\beta} =\varphi_\beta \circ \varphi_\alpha^{-1} \) are homeomorphisms on their domains of overlap.
\end{tcolorbox}

\greyline

\paragraph{Smooth Manifolds} A smooth manifold is a topological manifold equipped with a smooth structure. This means that, in addition to the local homeomorphisms to Euclidean space, the transition maps between overlapping neighborhoods are differentiable. More formally:

\begin{tcolorbox}[colback=gray!10, colframe=gray!40]
A topological space \( \mathcal{M} \) is an \( n \)-dimensional \textit{smooth manifold} if for every pair of points \( p, q \in \mathcal{M} \), there exist open neighborhoods \( U_\alpha \subseteq M \) around \( p \) and \( U_\beta \subseteq \mathcal{M} \) around \( q \) such that the transition map between the homeomorphisms \( \varphi_\alpha: U_\alpha \to \mathbb{R}^n \) and \( \varphi_\beta: U_\beta \to \mathbb{R}^n \) is a smooth (infinitely differentiable) map.
\end{tcolorbox}

\marginnote{Lie groups are both groups and (smooth) manifolds.} The smooth structure of these manifolds allows for the definition of smooth functions, smooth curves, and other objects in differential geometry, making them central to the study of calculus on manifolds.

\greyline

\paragraph{Diffeomorphisms} Diffeomorphisms allow for the transfer of geometric and differential properties between manifolds that share similar local structures.

\begin{tcolorbox}[colback=gray!10, colframe=gray!40]
A map between two manifolds \( \varphi: \mathcal{M} \to \mathcal{N} \) is a \textit{diffeomorphism} if: $\varphi$ is smooth (infinitely differentiable), $\varphi$ is bijective, and $\varphi^{-1}: \mathcal{N} \to \mathcal{M} $ is also smooth.
\end{tcolorbox}

While both homeomorphisms and diffeomorphisms are bijections that preserve certain structures, homeomorphisms preserve topological properties (such as continuity and connectedness), whereas diffeomorphisms preserve smooth (differentiable) structures. 

\greyline

\paragraph{Tangent Spaces and Bundles} The tangent space is a key concept for understanding the local geometry of the manifold.

\begin{tcolorbox}[colback=gray!10, colframe=gray!40]
\marginnote{To project points from the tangent space to the manifold and back we use exponential and logarithmic maps.}Given a smooth manifold \( \mathcal{M} \) and a point \( p \in \mathcal{M} \), the \textit{tangent space} at \( p \), denoted \( T_p\mathcal{M} \), is a vector space that represents the possible directions in which one can move away from \( p \). Formally, it is the space of equivalence classes of smooth curves passing through \( p \).
\end{tcolorbox}

\marginnote{$\bigsqcup$ refers to the disjoint union, whereas $\bigcup$ is used to denote the regular union. The former preserves the identity of the original sets, treating overlapping elements as distinct. On the other hand, the latter merges sets, discarding duplicate elements. In the context of the definition of tangent bundles, $\bigsqcup$ is used to emphasize that the tangent spaces at different points of the manifold are distinct and should be treated as separate entities, even if they may have overlapping elements.}\begin{tcolorbox}[colback=gray!10, colframe=gray!40]
The \textit{tangent bundle} of a smooth manifold \( \mathcal{M} \), denoted \( T\mathcal{M} \), is the disjoint union of all tangent spaces of \( \mathcal{M} \):
\[
T\mathcal{M} = \bigsqcup_{p \in \mathcal{M}} T_p\mathcal{M}.
\]
Each point \( (p, v) \in T\mathcal{M} \) consists of a point \( p \in \mathcal{M} \) and a tangent vector \( v \in T_p\mathcal{M} \). 
\end{tcolorbox}

\begin{figure}[hbpt!]
    \centering
\includegraphics[width=0.5\linewidth, trim={19cm 0 0 0}]{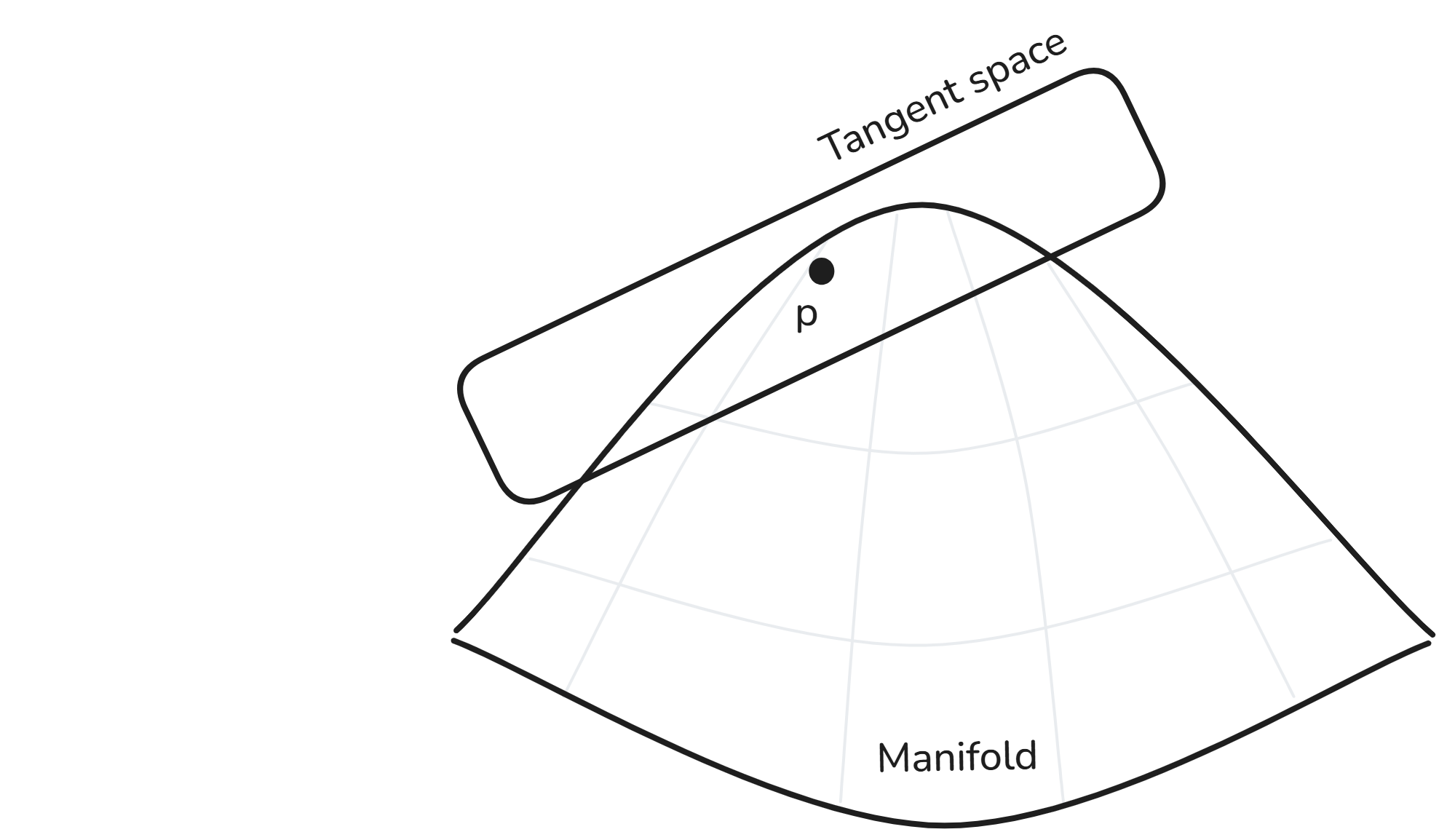}
        \caption{Illustration of the tangent space $T_p \mathcal{M}$ at a point $p$ on the manifold $\mathcal{M}$. The tangent space is a flat, vector-space approximation of the manifold at $p$.}
\label{fig:tangent_space}
\end{figure}\marginnote{\textit{“Manifolds in which, as in the plane and in space, 
the line-element may be reduced to the form $\sqrt{\sum dx^2}$, 
are therefore only a particular case of the manifolds 
to be here investigated; they require a special name, 
and therefore these manifolds in which the square of 
the line-element may be expressed as the sum of the 
squares of complete differentials I will call flat.”} -- Bernhard Riemann, 1856}

\greyline

\paragraph{Riemannian Manifolds} A Riemannian manifold is a smooth manifold equipped with a Riemannian metric, which is a smoothly varying inner product on the tangent spaces of the manifold. Formally:

\begin{tcolorbox}[colback=gray!10, colframe=gray!40]\marginnote{The sphere in particular is both a \textit{homogeneous manifold} and has \textit{constant curvature}. Without getting into formal definitions, a homogeneous manifold is a manifold with a high degree of symmetry, where the manifold looks the same at every point. A manifold is a \textit{constant curvature manifold} if its curvature (a measure of how the manifold bends in space) is the same at every point. Note that, more generally, manifolds can have variable curvature and very intricate structures, and that homogeneous manifolds with constant curvature, as well as products thereof, are simply easier-to-study special cases. See below an example of variable-curvature Riemannian geometry on manifolds.

\includegraphics[width=0.8\linewidth]{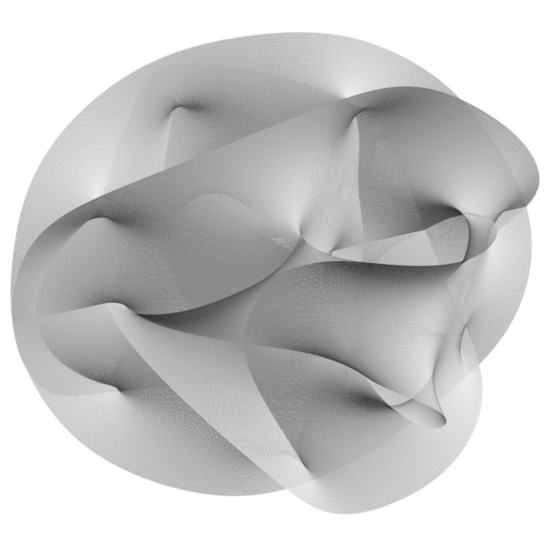}}
A smooth manifold \( \mathcal{M} \) is a \textit{Riemannian manifold} if it is equipped with a Riemannian metric, which is a smooth assignment of an inner product on the tangent space at each point \( p \in \mathcal{M} \), i.e., a map \( g_p: T_p\mathcal{M} \times T_p\mathcal{M} \to \mathbb{R} \) that is smooth in \( p \), where \( T_p\mathcal{M} \) is the tangent space at \( p \). We typically denote the Riemannian manifold as a tuple $(\mathcal{M},g).$
\end{tcolorbox}

\begin{figure}[hbpt!]
\centering
\includegraphics[width=0.45\linewidth]{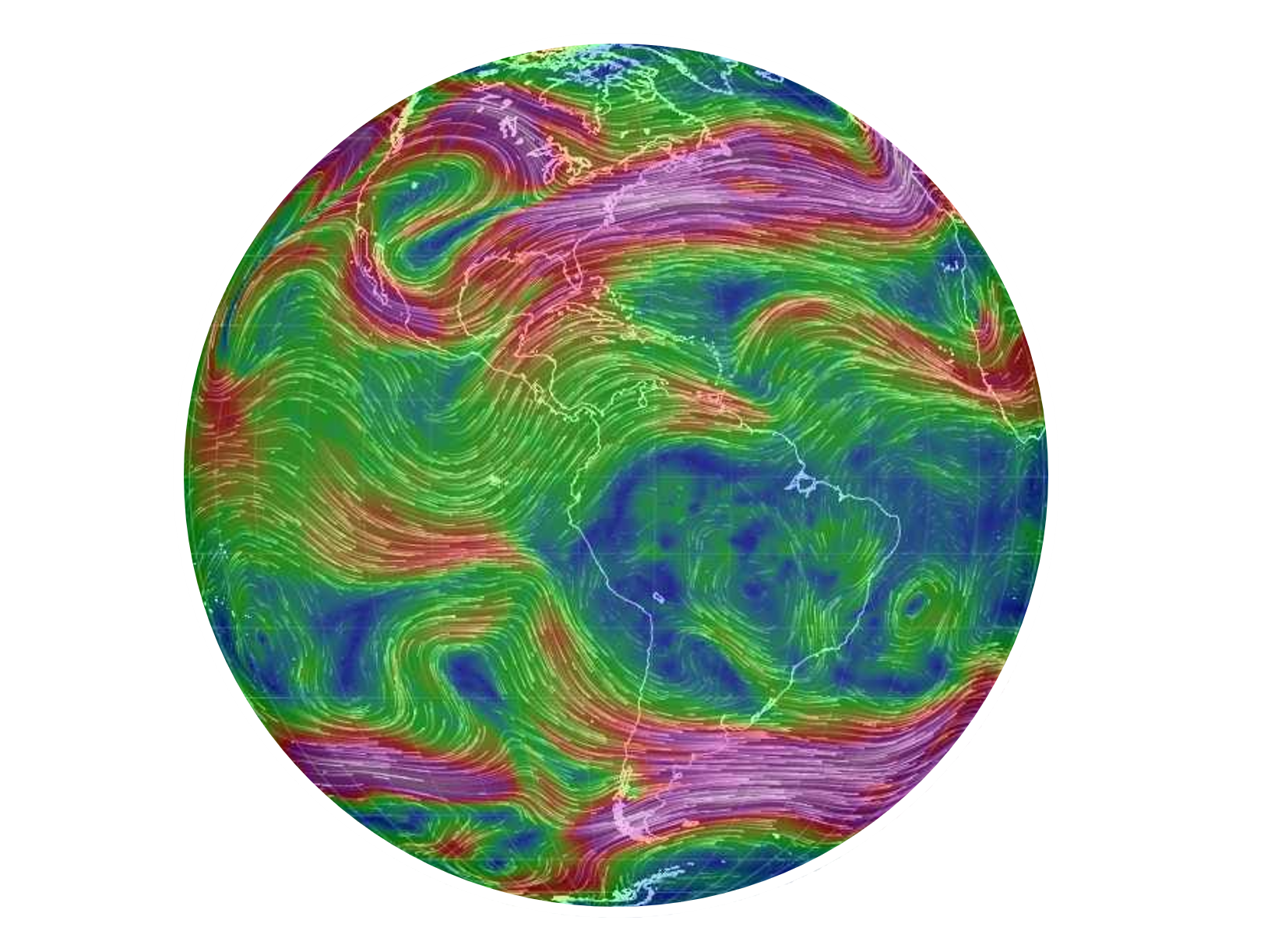}
    \caption{The sphere is an example of a Riemannian manifold, locally resembling Euclidean space. Indeed, when walking on the surface of the Earth, it appears flat. We can define functions on this manifold to characterize various phenomena, such as the distribution of atmospheric pressure or the velocity of the wind.}
    \label{fig:earth_sphere}
\end{figure} 

The Riemannian metric enables the measurement of distances between points and the definition of geodesics.

\begin{tcolorbox}[colback=gray!10, colframe=gray!40]
A \textit{geodesic} on a Riemannian manifold is a curve which locally minimizes the distance between points.
\end{tcolorbox}

In our day-to-day, we often say that `the shortest path between two points is always a straight line', and this is true for flat Euclidean space. However, in more general spaces, geodesics may not be straight lines. For example, when connecting two points on the surface of a sphere, the shortest path is an arc of a great circle.\marginnote{Here, we only aim to provide the intuitive idea behind the concept of geodesics. For a more mathematically rigorous understanding of geodesics one would need to introduce the geodesic equation, which is derived from the principle of least action applied to the length of a curve. This relies on presenting concepts such as metric tensors, Euler-Lagrange equations, and Christoffel symbols, which we omit for simplicity.}

\greyline

\paragraph{Exponential and Logarithmic Maps}  
For any point \( p \) on a Riemannian manifold \( (\mathcal{M},g) \), as previously discussed, the \textit{tangent space} \( T_p\mathcal{M} \) is a vector space that locally approximates the manifold. A fundamental tool in differential geometry is the exponential map at \( p \):

\begin{tcolorbox}[colback=gray!10, colframe=gray!40]
The \textit{exponential map}, 
\[
\exp_p : T_p\mathcal{M} \to \mathcal{M},
\]
takes a tangent vector \( v\in T_p\mathcal{M} \) and returns a point on the manifold reached by following the unique geodesic starting at \( p \) in the direction \( v \) for a distance equal to the norm \( \|v\| \). 
\end{tcolorbox}

\begin{figure}[hbpt!]
    \centering
\includegraphics[width=0.5\linewidth, trim={19cm 0 0 0}]{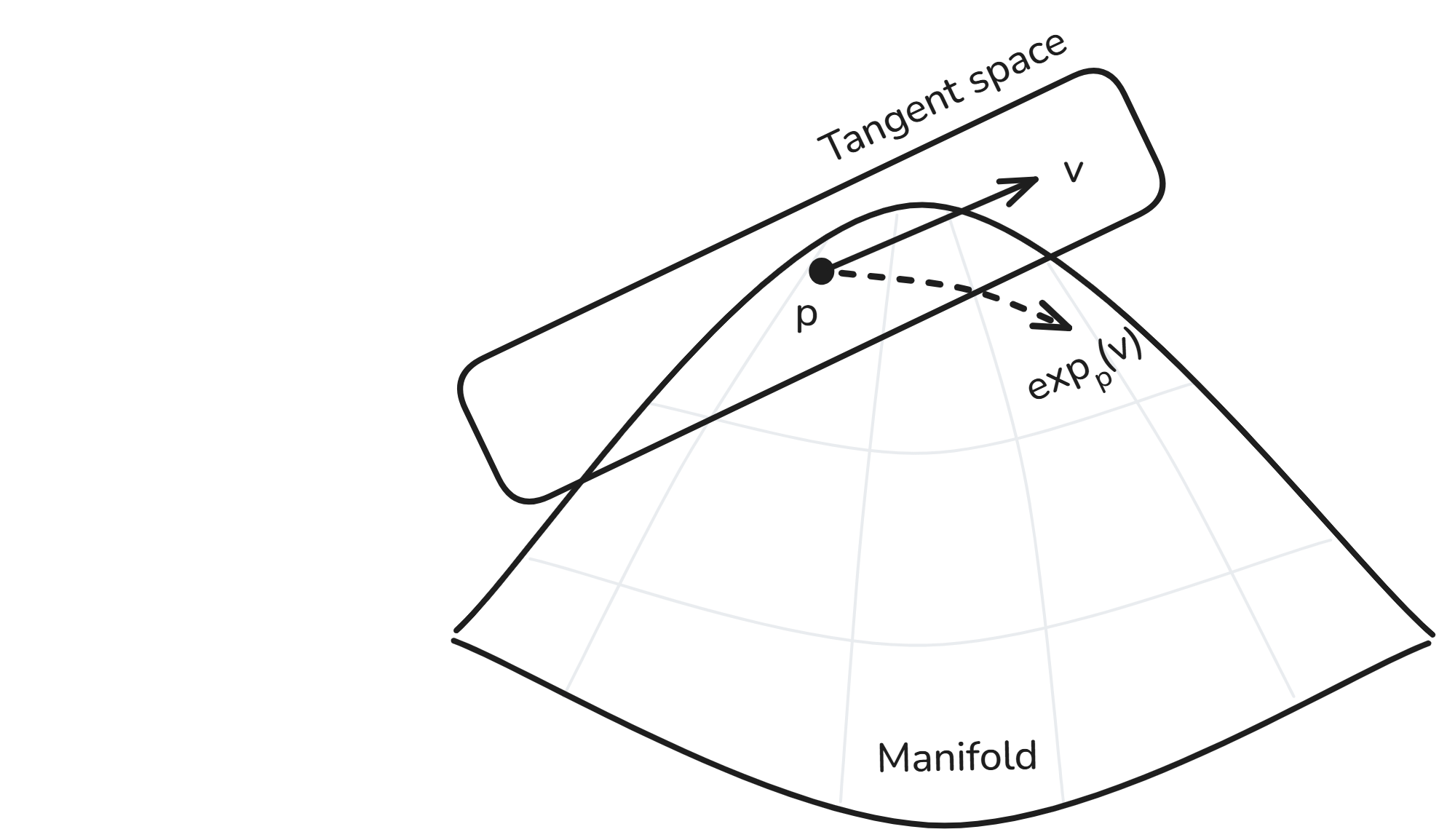}
        \caption{Illustration of the exponential map $\exp_p$, mapping a tangent vector $v$ at point $p$ in the tangent space $T_p\mathcal{M}$ to a point $\exp_p(v)$ on the manifold $\mathcal{M}$. This mapping is realized by following the geodesic starting at $p$ in the direction of $v$ for a distance equal to $\|v\|$.}
\label{fig:tangent_space2}
\end{figure}

Next, we provide an example. Let \( p = (0, 0, 1) \in S^2 \subset \mathbb{R}^3 \) be the north pole of the unit sphere, and let \( v = (\epsilon, 0, 0) \in T_p S^2 \) be a small tangent vector. Note that \( T_p S^2 \) consists of all vectors in \( \mathbb{R}^3 \) that are perpendicular to \( p \):
\[
T_p S^2 = \{ v \in \mathbb{R}^3 \mid v \cdot p = 0 \} = \{ (x, y, 0) \in \mathbb{R}^3 \}.
\]

Since geodesics on \( S^2 \) are great circles, the geodesic starting at \( p \) in the direction of \( v \) can be expressed as: $\gamma(t) = \cos(t)\,p + \sin(t)\,\hat{v},$
where \( \hat{v} = \frac{v}{\|v\|} = (1, 0, 0) \). Evaluating this at \( t = \|v\| = \epsilon \) gives:
\[
\exp_p(v) = \cos(\epsilon)\,(0, 0, 1) + \sin(\epsilon)\,(1, 0, 0) = (\sin(\epsilon), 0, \cos(\epsilon)).
\]

For small \( \epsilon \), this is approximately:
\[
\exp_p(v) \approx (\epsilon, 0, 1 - \tfrac{\epsilon^2}{2}),
\]
which captures the fact that the sphere is locally well-approximated by a flat plane.

\begin{tcolorbox}[colback=gray!10, colframe=gray!40]
The \textit{logarithmic map}, 
\[
\log_p : \mathcal{M} \to T_p\mathcal{M},
\]
is the local inverse of the exponential map. It maps a point \( q \in \mathcal{M} \) (sufficiently close to \( p \)) to the tangent vector \( v \in T_p\mathcal{M} \) such that \( \exp_p(v) = q \). In other words, it returns the initial velocity of the geodesic starting at \( p \) and reaching \( q \).
\end{tcolorbox}

These tools allow us to conduct operation in the locally flat tangent space and to project point from and back to it.

\begin{tcolorbox}[colback=orange!20, colframe=orange!60]
\textbf{Embedding Latent Representations into non-Euclidean Manifolds using the Exponential Map.}\marginnote{The Poincaré ball is a model of hyperbolic geometry represented as a unit ball, where distances grow infinitely as one approaches the boundary. In 2D we often refer to it as the Poincaré disk instead.} For instance, in the context of embedding hierarchical representations, the exponential map can be employed to project the output of an encoder onto a specific manifold, such as the Poincaré ball. Initially, one applies multiple non-linear transformations to the input (encoder), obtaining latent representations that (are assumed to) reside in a Euclidean space. Subsequently, these representations are mapped onto the desired manifold via the exponential map.
\end{tcolorbox}

\greyline

\clearpage

\paragraph{Tangent Vector Fields} In the context of Geometric Deep Learning, tangent vector fields defined smoothly across a manifold can be used to encode local geometric information at each point, see Figure~\ref{fig:vectorfield_manifold} below.

\begin{tcolorbox}[colback=gray!10, colframe=gray!40]
A \textit{tangent vector field} on a smooth manifold \( \mathcal{M} \) is a smooth assignment of a tangent vector \( v_p \in T_p\mathcal{M} \) to each point \( p \in \mathcal{M} \). Formally, it is a smooth mapping:
\[
V : \mathcal{M} \rightarrow T\mathcal{M}, \quad p \mapsto V(p) = v_p \in T_p\mathcal{M}.
\]
This mapping ensures continuity and differentiability, allowing for consistent geometric analysis across the manifold.
\end{tcolorbox}

A natural question is: \textit{what is the difference between a tangent bundle and a tangent vector field?} In short, the tangent bundle is the space of all possible tangent vectors at all points, while a tangent vector field is a smooth assignment of one specific tangent vector to each point on the manifold. For clarity, let us discuss an intuitive example. Imagine the Earth's surface as your manifold, that is, a sphere. A tangent vector field is a specific weather map showing the wind direction and speed at every single point on the Earth right now. It is one specific wind pattern selected from all possibilities throughout the day, months, years, decades, etc. On the other hand, the tangent bundle would correspond to the entire collection of all possible wind arrows you could ever imagine drawing at every single point on the Earth, no matter the direction or speed (representing every conceivable instantaneous motion at that point).

\begin{figure}[hbpt!]
\centering
        \includegraphics[width=0.6\linewidth]{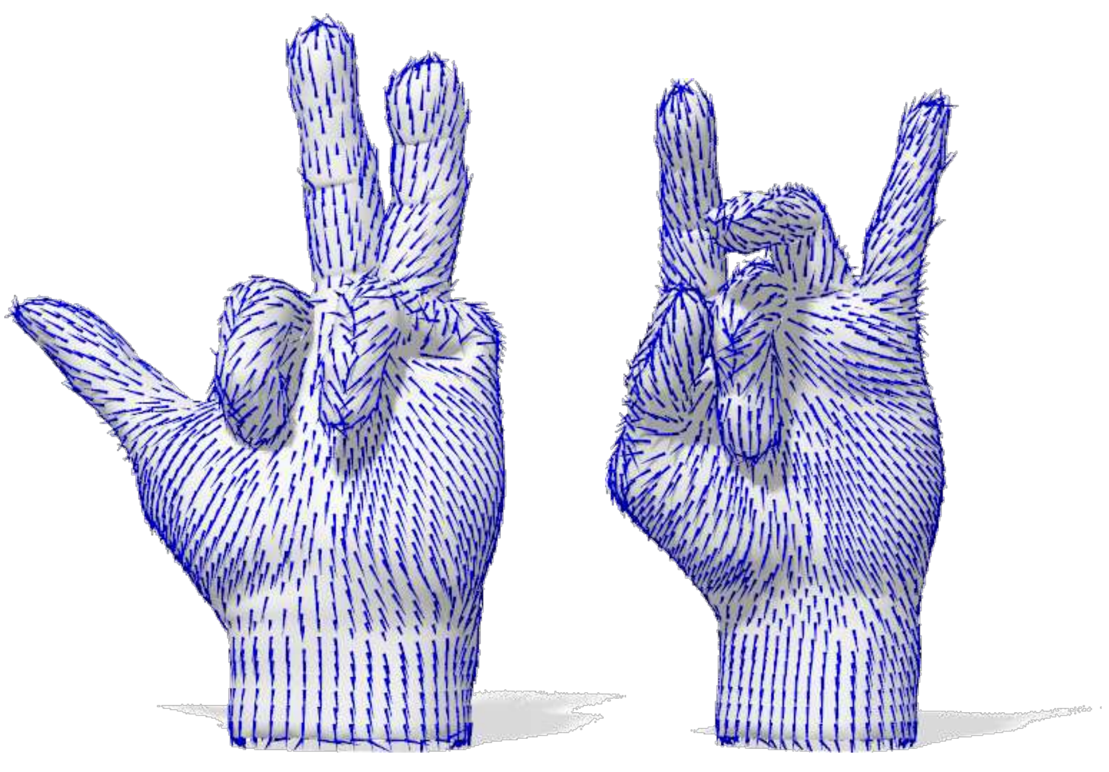}
    \caption{In many problems in Geometric Deep Learning and Geometric Data Processing we work with tangent vector fields.}
    \label{fig:vectorfield_manifold}
\end{figure} 

\greyline

\paragraph{Gauges and Gauge Transformations} In practical scenarios, for instance when we want to process a signal by applying a filter, we often need to select a local coordinate system, known as a \textit{gauge}, at each point \(p\) on the manifold. 

\begin{tcolorbox}[colback=gray!10, colframe=gray!40]
Given a manifold \(\mathcal{M}\) and a point \(p \in \mathcal{M}\), a \textit{gauge} at \(p\) is a local isomorphism \(\omega_p: \mathbb{R}^n \rightarrow T_p\mathcal{M}\).
\end{tcolorbox}\marginnote{Here we refer to isomorphisms: diffeomorphisms that preserve the relevant algebraic structure, like linearity. Note that a homeomorphism is not sufficient for the definition of a gauge.}

In the above definition, \(\mathbb{R}^n\) is an \(n\)-dimensional vector space (the model space); \(T_p\mathcal{M}\) is the tangent space of \(\mathcal{M}\) at the point \(p\); and a local isomorphism is a linear mapping that preserves the structure and is invertible. 

\clearpage
However, the choice of this local coordinate system is not unique. We can choose different, equally valid gauges. A \textit{gauge transformation} describes how to switch between these different, equally valid local coordinate systems. 

\begin{tcolorbox}[colback=gray!10, colframe=gray!40]
Given a manifold \( \mathcal{M} \) and a point \( p \in \mathcal{M} \), a \textit{gauge transformation} between two gauges \( \omega_p: \mathbb{R}^n \rightarrow T_p\mathcal{M} \) and \( \omega'_p: \mathbb{R}^n \rightarrow T_p\mathcal{M} \) at \( p \) is an isomorphism \( \tau: \mathbb{R}^n \rightarrow \mathbb{R}^n \) such that:
\[
\omega'_p = \omega_p \circ \tau.
\]
\end{tcolorbox}\marginnote{\(\tau\) has type \(\mathbb{R}^n\!\to\!\mathbb{R}^n\) and it encodes how the two gauges differ, while \(\omega\) has type \(\mathbb{R}^n\!\to\!T_p\mathcal{M}\) sending the standard basis to that of the tangent space. Thus the only way to form a composite \(\mathbb{R}^n\to T_p\mathcal{M}\) is \(\omega\circ\tau\), not \(\tau\circ\omega\).
}

\greyline

\paragraph{Gauge Equivariance in Convolution Operators and Signal Processing} When designing operators (e.g., convolution) on manifolds, we typically define a filter function \( \psi \) on the tangent space that acts on features from a function \( f:\mathcal{M}\to\mathbb{R}^C \). To ensure the operation is independent of the arbitrary choice of gauge, the filter must be \textit{gauge equivariant}. That is, under a gauge transformation \( \tau \), the filter satisfies
\[
\psi\bigl(\tau\,(v)\bigr) \;=\; \tau\,(\psi(v)),
\]
\marginnote{Traditional neural network architectures can be adapted to work on manifolds, meshes, and geometric graphs by focusing on local neighborhoods.

\includegraphics[width=\linewidth]{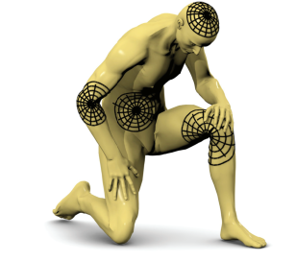}}where the action of \( \tau \) on the feature vector \( \psi(v) \) is defined by the same representation. Consequently, a gauge-equivariant convolution is defined as
\[
(f\star\psi)(p) \;=\; \int_{T_p\mathcal{M}} \psi(v)\, f\bigl(\exp_p(v)\bigr)\, dv,
\]
which ensures that any change in the local gauge is appropriately counteracted by the corresponding transformation of the filter. This property is essential in applications, as it guarantees that learned features and convolutions are intrinsic to the manifold and not contingent on an arbitrary choice of coordinates.

\greyline

\paragraph{Product Manifolds} Moreover, similarly to Cartesian products between sets, it is also possible to define product manifolds based on the Cartesian product of two subspaces. For example, taking the Cartesian product of two circles (1-spheres) yields a torus. Product manifolds are useful for building more complex yet computationally tractable and interpretable spaces from simpler, well-understood components. Mathematically, the product of two manifolds \( \mathcal{M} \) and \( \mathcal{N} \) is a new manifold \( \mathcal{M} \times \mathcal{N} \). The tangent space at a point \( (p, q) \in \mathcal{M} \times \mathcal{N} \) is the direct sum of the tangent spaces at \( p \in \mathcal{M} \) and \( q \in \mathcal{N} \), i.e., 
\[
T_{(p, q)}(\mathcal{M} \times \mathcal{N}) = T_p\mathcal{M} \oplus T_q\mathcal{N}.
\]
Here, the direct sum \( \oplus \) refers to the combination of two vector spaces (or tangent spaces) such that each element of the resulting space is uniquely a pair consisting of one element from each of the original spaces. A Riemannian metric on the product manifold is then defined as the sum of the individual metrics on \( \mathcal{M} \) and \( \mathcal{N} \). In certain applications, especially in machine learning models using latent spaces composed of constant-curvature manifolds such as spheres or hyperbolic spaces, it is common to define a distance on the product space by combining the individual geodesic distances as:
\[
d((x_1, x_2), (y_1, y_2)) := \sqrt{d_{\mathcal{M}}(x_1, y_1)^2 + d_{\mathcal{N}}(x_2, y_2)^2}.
\]
This distance function does not generally coincide with the geodesic distance of the Riemannian product manifold but is instead a modeling choice that simplifies computations and leverages closed-form geodesics in the component spaces.

\begin{tcolorbox}[colback=orange!20, colframe=orange!60]
\textbf{Manifolds in Geometric Deep Learning.} Geometric Deep Learning aims to extend neural network architectures to effectively handle data defined on general non-Euclidean domains, including manifolds such as surfaces in 3D space or more abstract, higher-dimensional spaces. When we talk about data lying on a manifold, we often implicitly assume that this manifold has some geometric structure that we want our models to understand and leverage. This structure usually involves notions of distance or similarity, which falls under the umbrella of `geometry'. The manifold provides the framework, and the geometry provides the rules for measurement and relationships on that framework. This involves leveraging tools from differential geometry, like geodesics, curvature, and local charts, to design models that respect the manifold's intrinsic geometry. For example, convolution-like operations on manifolds may be defined in terms of local neighborhoods, where the neighborhood structure is governed by the manifold's geometry rather than a regular grid.
\end{tcolorbox}

\greylinelong

\subsection{The Manifold Hypothesis}

Many ML and AI algorithms rely on the \textit{manifold hypothesis}~\cite{10.1109/TPAMI.2013.50} (sometimes also called the manifold assumption), which suggests that although most datasets seem to be high-dimensional in the original data space, data points can actually be described by a low-dimensional manifold which resides within the observed high-dimensional space. This is often used to explain why datasets that appear to require a great number of parameters to be represented, can in practice be encoded using latent variables with few dimensions.\marginnote{Often the term manifold is abused in ML and AI.} As a disclaimer, note that the term `manifold' is used loosely in this context and not in a mathematically rigorous sense. There are no formal guarantees that the low-dimensional representation possesses the mathematical properties discussed earlier in Section~\ref{sec:Manifolds and Differential Geometry}. For example, the space may not be perfectly smooth, locally Euclidean, or even have consistent local dimensionality.

\begin{figure}[htbp!]
  \centering
  \includegraphics[width=0.75\linewidth]{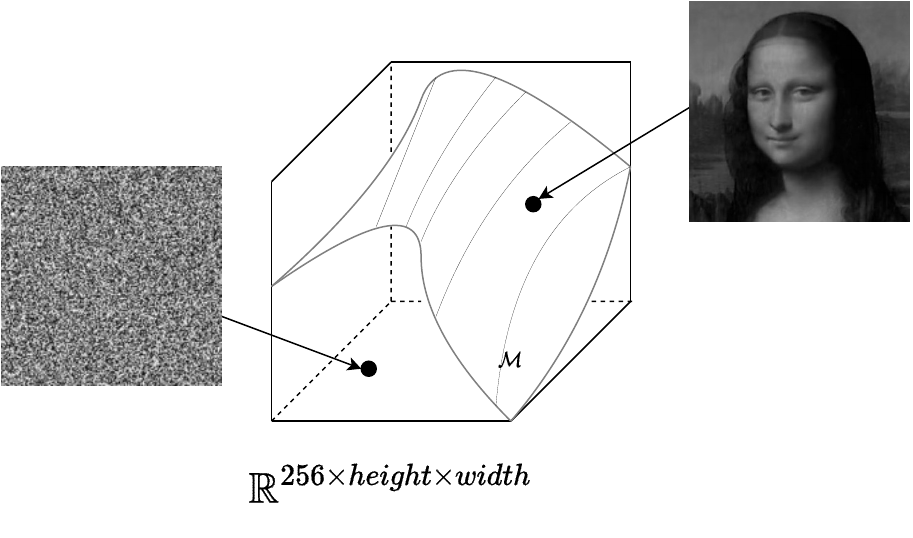}
  \caption{The manifold which encapsulates all images of faces, is expected to be substantially more low-dimensional than the space \(\mathbb{R}^{256 \times \textit{height} \times \textit{width}}\). Points on the manifold correspond to valid face images, whereas the remaining points in the hypercube are likely to produce meaningless, noisy images.}
  \label{manifold_learning_1}
\end{figure}

This idea can be more clearly illustrated with a simple example. Consider a dataset of grayscale images with fixed height and width. Although the dataset
$
\mathcal{D} \subset \mathbb{R}^{256 \times \textit{height} \times \textit{width}}
$
formally lies within a high-dimensional space, most points in this space correspond to meaningless noise. Only a small subset—those lying on the data manifold—represent valid images, such as faces. As shown in Figure~\ref{manifold_learning_1}, points on the manifold correspond to structured, coherent data, whereas random coordinates in the space typically yield unrecognizable outputs. A key goal in many machine learning approaches is to uncover this low-dimensional manifold that captures the true structure of the data.

\begin{figure}[htbp!]
  \centering
  \includegraphics[width=0.65\linewidth]{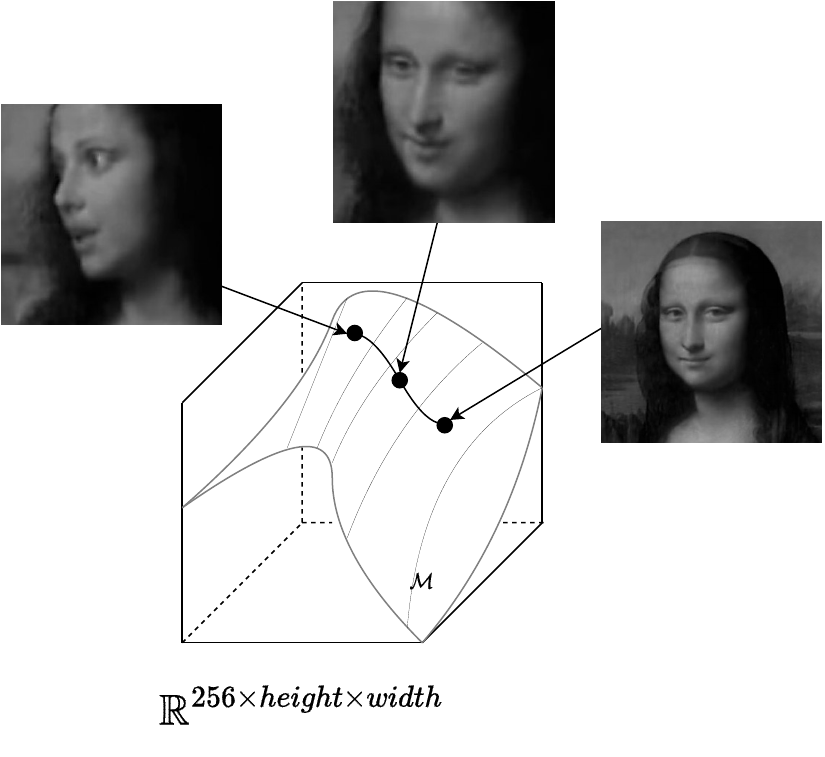}
  \caption{Depiction of interpolation between images along the surface of the manifold.}
  \label{manifold_learning_2}
\end{figure}

Figure~\ref{manifold_learning_2} illustrates that traversing the manifold allows for controlled variation (such as different facial expressions or poses) while remaining within the space of valid images. In contrast, simple linear interpolation between two images in pixel space generally produces noisy or implausible results. Empirically, smooth transitions can often be observed when interpolating in the latent space of models such as autoencoders. Still, there are no theoretical guarantees that smooth interpolations exist between any two arbitrary points.

\clearpage

\section{Functional Analysis}

Functional analysis is a branch of mathematical analysis that studies spaces of functions and the operators that act on them. Functional analysis provides a powerful framework for understanding infinite-dimensional spaces, where classical linear algebraic methods fail, and establishes the foundation for spectral theory. This section explores key concepts such as completeness, convergence, and the structural properties of vector spaces, with a focus on Banach and Hilbert spaces as fundamental mathematical structures.

\begin{tcolorbox}[colback=orange!20, colframe=orange!60]
\textbf{Banach and Hilbert Spaces in Geometric Deep Learning.} Banach and Hilbert spaces serve as a critical foundation for key concepts such as eigenfunctions, eigenvalues, and Fourier analysis, which we will study in Section~\ref{Spectral Theory} and which are widely used in many Geometric Deep Learning algorithms. We encourage readers to review the material on Banach and Hilbert spaces, operators, and functionals. While an in-depth study of these concepts may not be necessary, a basic understanding is useful to tackle spectral theory.

\end{tcolorbox}

\greylinelong

\subsection{Cauchy Sequences and Banach Spaces}
\begin{tcolorbox}[colback=gray!10, colframe=gray!40]
A sequence of vectors $ v_1, v_2, \ldots \in V $ in a normed vector space $ V $ is a \textit{Cauchy sequence} if for every $ \epsilon > 0 $, there exists an integer $ N $ such that  
$$
\|v_m - v_n\| < \epsilon \quad \text{for all } m, n > N.
$$  
\end{tcolorbox}

As indices $ m $ and $ n $ become arbitrarily large, the vectors $ v_m $ and $ v_n $ approach each other in norm, satisfying:
$$
\lim_{m, n \to \infty} \|v_m - v_n\| = 0.
$$

Critically, a Cauchy sequence does not inherently guarantee a limit within the space $ V $. \marginnote{Consider two spaces $V_1=(0,1]$ and $V_2=(0,1)$, and the sequence $d_n = 1 - \frac{1}{n},$ where $n$ is a positive integer. As $n\rightarrow\infty$, the sequence tends to $1$. In the case of $V_1$ the sequence converges within the space. On the other hand, in $V_2$ the boundary is not part of the space, and hence the sequence does not converge within $V_2$ even though it is Cauchy.} The existence of such a limit depends on the space's completeness.

\begin{tcolorbox}[colback=gray!10, colframe=gray!40]
A \textit{Banach space} is a normed vector space $ V $ that is complete, meaning every Cauchy sequence $ (v_n)_{n \geq 1} $ has a limit $ v \in V $ such that:
$$
\lim_{n \to \infty} \|v_n - v\| = 0,
$$
equivalently converging in the topology induced by the norm:
$$
\lim_{n \to \infty} v_n = v.
$$
\end{tcolorbox}

Banach spaces provide a framework for studying convergence in infinite-dimensional spaces, and they generalize the notion of completeness from real numbers to vector spaces.

A prototypical Banach space is $ \ell^p $ (for $ 1 \leq p < \infty $), defined by sequences $ (x_n)_{n \geq 1} $ satisfying:
$$
\|x\|_p = \left( \sum_{n=1}^\infty |x_n|^p \right)^{1/p} < \infty.
$$

\clearpage

The importance of completeness is illustrated by a counterexample in $ \mathbb{Q} $ with the absolute value norm. Consider the sequence approximating $ \sqrt{2} $:
$$
v_n = \text{the $ n $-th rational approximation of $ \sqrt{2} $}.
$$
This sequence is Cauchy in $ \mathbb{Q} $, but its limit $ \sqrt{2} $ lies outside $ \mathbb{Q} $. This demonstrates why completeness is crucial: it prevents Cauchy sequences from `escaping' the original space. 

\greylinelong

\subsection{Hilbert Spaces}

\begin{tcolorbox}[colback=gray!10, colframe=gray!40]
A {\em Hilbert space} is a complete inner product space. \marginnote{ Hilbert spaces combine the algebraic structure of inner products with the topological properties of completeness. Completeness ensures that the space is well-suited for analyzing convergence of Fourier series, solving partial differential equations, and modeling quantum systems. Hilbert spaces unify algebra, geometry, and analysis in an infinite-dimensional setting.}
\end{tcolorbox}

Hilbert spaces extend the notion of Banach spaces by introducing an inner product $ \langle \cdot, \cdot \rangle $ that induces the norm:  
$$
\|v\| = \sqrt{\langle v, v \rangle}.
$$  

The inner product allows Hilbert spaces to generalize the geometry of finite-dimensional Euclidean spaces to infinite dimensions. Key examples include $ L^2 $ (square-integrable) spaces, where functions are treated as infinite-dimensional vectors. 

\greyline

\paragraph*{Orthogonal Bases} Let $V$ be a Hilbert space and let $S\subseteq V$. 

$$\mathrm{span}(S) = \left\{ \sum_{i=1}^n \alpha_i v_i  : n\in \mathbb{N}, \, v_i \in S, \, \alpha_i \in \mathbb{C} \right\}$$ \marginnote{The equation states (contrapositive form) that if the linear combination equals the zero vector, then all the coefficients $\alpha_i$ must be zero. This is a defining property of linear independence.} 

is the set of all finite linear combinations from $S$. 

$S$ is {\em linearly independent} if for any finite subset $\{v_1,\ldots,v_n\} \subseteq S$ and any coefficients $\alpha_1,\ldots,\alpha_n \in \mathbb{C}$,
$$
\sum_{i=1}^n \alpha_i v_i = 0 \implies \alpha_i = 0 \, \forall i.
$$

$S$ is {\em orthogonal} if $\langle u, v\rangle = 0$ \, $\forall u, v  \in S$ s.t. $u\neq v$. \marginnote{As mentioned in Section~\ref{subsec:Inner Product}, orthogonality is typically denoted via $u \perp v$. To denote that vectors are orthonormal sometimes the following notation is used: $u \perp\!\!\perp v$.}

$S$ is {\em orthonormal} if it is orthogonal and all vectors have unit length, i.e. $\|u\| = 1$ \, $\forall u  \in S$. 

When $\{e_i\}_{i \in I}$ forms an orthonormal basis for $V$, every element $v \in V$ has a unique infinite representation:
$$
v = v_1 e_1 + v_2 e_2 + \dots = \sum_{i \in I} v_i e_i = v_ie_i = \sum_{i \in I} \langle v, e_i \rangle e_i,
$$
where $\langle v, e_i \rangle$ are the \textit{Fourier coefficients} (Section~\ref{subsec:Fourier_analysis}), and the series converges in the norm induced by the inner product.

\greyline

\paragraph*{Functions as Infinite-Dimensional Vectors in $ L^2 $} \marginnote{Sometimes Hilbert spaces are tacitly assumed separable, yielding the property of isometry to $\ell^2$.} A square-integrable function is a function $ f $ defined on a domain $ \Omega $ such that the square of its absolute value is integrable over $ \Omega $. Specifically, a function $ f(x) $ belongs to the space $ L^2(\Omega) $ if:

$$
\int_\Omega |f(x)|^2 \, dx < \infty.
$$

Functions in $ L^2 $ spaces can be understood as infinite-dimensional vectors by representing them in terms of a set of basis functions. Just as finite-dimensional vectors in $ \mathbb{R}^n $ can be expressed using a basis (e.g., $ v = v_1 e_1 + v_2 e_2 + \dots + v_n e_n = v_ie_i $), a function $ f(x) $ in $ L^2 $ can be written as a linear combination of basis functions:
$$
f(x) = f_1 \phi_1(x) + f_2 \phi_2(x) + f_3 \phi_3(x) + \dots
$$
Here, $ \{\phi_k(x)\}_{k=1}^\infty $\marginnote{An orthonormal set of basis functions are orthogonal $ \langle \phi_i, \phi_j \rangle = 0 $ for $ i \neq j $, and normalized $ \langle \phi_i, \phi_i \rangle = 1 $.} is a set of orthonormal basis functions, and the coefficients $ f_k $ represent how much of each basis function $ \phi_k(x) $ contributes to $ f(x) $. The coefficients $ f_k $ are computed using the inner product of $ f(x) $ with the basis function $ \phi_k(x) $:
$$
f_k = \langle f, \phi_k \rangle = \int f(x) \phi_k(x) \, dx.
$$
This step is analogous to finding the components of a vector in $ \mathbb{R}^n $ by projecting it onto the coordinate axes. \marginnote{Analogous to the expression above: $\sum_{i \in I} v_i e_i = \sum_{i \in I} \langle v, e_i \rangle e_i$.} Once the coefficients $ f_1, f_2, f_3, \dots $ are determined, the function $ f(x) $ can be viewed as an infinite-dimensional vector:
$$
f \equiv [f_1, f_2, f_3, \dots].
$$
In this sense, the `vector' $ [f_1, f_2, f_3, \dots] $ describes $ f(x) $ completely, just as the coordinates $ [v_1, v_2, \dots, v_n] $ describe a vector in finite-dimensional space. 

For example, consider the interval $ X = [0, 1] $ with basis functions $ \phi_1(x) = 1 $, $ \phi_2(x) = \sin(10\pi x) $, and $ \phi_3(x) = \cos(\pi x) $. A function $ f(x) = 2 + 17\sin(10\pi x) - \cos(\pi x) $ can be written as:
$$
f(x) = 2 \cdot \phi_1(x) + 17 \cdot \phi_2(x) - 1 \cdot \phi_3(x).
$$
In this case, the coefficients are $ f_1 = 2 $, $ f_2 = 17 $, and $ f_3 = -1 $, and the function $ f(x) $ is represented as the vector $ [2, 17, -1] $. Extending this idea to infinitely many basis functions gives the full $ L^2 $ perspective, where $ f(x) $ is reconstructed as a weighted sum of basis functions.

This approach provides an intuitive understanding of functions as vectors in infinite-dimensional spaces, where concepts like orthogonality, projection, and decomposition of functions naturally extend from finite-dimensional vector spaces.

\greylinelong

\subsection{Operators and Functionals}

In the context of Banach and Hilbert spaces, operators and functionals serve as essential tools for understanding the relationships between elements within and across spaces. They form the backbone of functional analysis. For the sake of brevity, here we only provide a very concise and high-level description of the aforementioned concepts.

\greyline

\paragraph*{Operators on Banach and Hilbert Spaces} Operators are mappings that transform elements from one space into another while preserving structure. Through operators, we can describe how vectors interact, how they transform, and how these transformations affect the overall structure of the space.

\begin{tcolorbox}[colback=gray!10, colframe=gray!40]
An {\em operator} in this context is a map $A: U \rightarrow V$ between two spaces $U$ and $V$ (Banach or Hilbert), usually preserving some structure.
\end{tcolorbox}

Let $(U, \| \,\|_U)$ and $(V, \| \,\|_V)$ be Banach spaces with their respective norms, and consider an operator $A: U \rightarrow V$.

$A$ is {\em continuous} \marginnote{$u_n$ here refers to a sequence in the space $u_n = (u_n)_{n \geq 1}$.} if it preserves convergence, i.e., $u_n\overset{\| \, \|_{U}}{\longrightarrow} u \Rightarrow Au_n \overset{\| \, \|_{V}}{\longrightarrow} Au$.

$A$ is {\em bounded} if $\exists c >0$ s.t. $\| Au \|_V \leq c \| u\|_U$ \, $\forall u \in U$.

$A$ is {\em linear} if $A(\alpha u + \beta w) = \alpha A u + \beta A w$ \, $\forall u,w \in U$ and $\alpha, \beta \in \mathbb{C}$.

$A$ is an {\em isometry} if it is length-preserving, i.e. $\| Au \|_V = \| u\|_U$.

Let $(V, \langle \, , \, \rangle))$ be a Hilbert space and consider an operator $A: V \rightarrow V$. \marginnote{In the context of Hilbert spaces we use the asterisk symbol $(\cdot)^*$ to denote adjoint operators.}

$A^*$ is {\em adjoint to $A$} if $\langle Au, v \rangle = \langle u, A^* v\rangle$ \quad $\forall u, v \in V$.

$A$ is {\em self-adjoint} if $A^* = A$, i.e. $\langle Au, v \rangle = \langle u,  Av\rangle$ \quad $\forall u, v \in V$. 

$A$ \marginnote{Weak limit ($ v_n \rightharpoonup v $): $ v_n $ converges weakly to $ v $ if $ \langle v_n, w \rangle \to \langle v, w \rangle $ for all $ w \in V $. This means that $ v_n $ converges to $ v $ in the sense of how they interact with other vectors, but not necessarily in norm.

Strong limit ($ v_n \to v $): $ v_n $ converges strongly to $ v $ if $ \| v_n - v \| \to 0 $, i.e., the distance between $ v_n $ and $ v $ in the norm goes to zero.} is {\em compact} if it maps weak limits to strong limits, i.e. $v_n \rightharpoonup v$ \quad $\Rightarrow$ \quad $Av_n \rightarrow Av$.

The \textit{rank} of an operator $ A $, denoted as $\text{rank}(A)$, is the dimension of the image of $A$, i.e., the number of linearly independent vectors in the set of vectors that $A$ maps to.

Note that in the space of finite-dimensional real vectors, operators can be expressed as matrices: $\langle Au, v\rangle = (A u )^\top v  = u^\top (A^\top v) = \langle u, A^\top v\rangle$. More on this next.

\greyline

\paragraph*{Functionals on Hilbert Spaces} Functionals are maps that assign scalar values to vectors. They provide a way to probe and measure elements of a space.

\begin{tcolorbox}[colback=gray!10, colframe=gray!40]
A {\em functional} is a map of the form $\phi : V \rightarrow \mathbb{C}$ on a Hilbert space $V$. 
\end{tcolorbox}

$\phi$ is {\em continuous}\marginnote{Note that continuity implies boundedness, that is, there exists a constant $C$ such that $ |\phi(v)| \leq C \| v \|_V.$} if it preserves convergence, i.e., if $ v_n \overset{\| \, \|_V}{\longrightarrow} v $ in $ V $, then $ \phi(v_n) \overset{}{\longrightarrow} \phi(v) $, where $ \| \cdot \|_V $ is the norm on $ V $.

$\phi$ is a {\em linear} functional if $\phi(\alpha v + \beta w) = \alpha \phi(v) + \beta \phi(w)$ $\forall v,w \in V$ and $\alpha, \beta \in \mathbb{C}.$

\noindent {\em Dual} (or {\em conjugate}) {\em space} to $V$ is the space of \textit{linear continuous functionals} on $V$, denoted
$$
V^* = \{ \phi: V \rightarrow \mathbb{C} \,\,\, \text{linear+continuous}
\}
$$
The elements of $V^*$ are called {\em dual vectors}.

\clearpage

\section{Spectral Theory}
\label{Spectral Theory}

Spectral theory studies the properties of operators and matrices by analyzing their spectra, that is, their eigenfunctions and associated eigenvalues.

\greylinelong

\subsection{Eigenfunctions and Eigenvalues}

Eigenfunctions and eigenvalues arise when we study linear transformations, whether on finite-dimensional vector spaces or infinite-dimensional spaces like function spaces. They allow us to decompose and \textit{diagonalize} operators. This can enable us to work with a simplified version of the original problem, one that might exhibit complex, non-linear dynamics in the original space. These concepts are particularly central to \textit{spectral theory}.

\begin{tcolorbox}[colback=gray!10, colframe=gray!40]
Let $A: V \rightarrow V$ be an operator on Hilbert space $V$. A vector $v\neq 0$ satisfying for some $\lambda$ 
$$
Av = \lambda v
$$
is  called an {\em eigenfunction} of $A$, and $\lambda$ is the corresponding {\em eigenvalue}. 
\end{tcolorbox}

Note that eigenfunctions are defined up to scale: if $v$ is an eigenfunction of $A$, so is $\alpha v$ for any $\alpha \neq 0$, since we can multiply both sides of the equation by $A (\alpha v) = \lambda (\alpha v)$ by $\alpha$. 
It is common to assume eigenfunctions of unit length, i.e. $\| v\| = 1$. 

\greyline

\paragraph*{Eigenvectors and Eigenvalues in Finite-Dimensional Vector Spaces} When we are first introduced to eigenvectors and eigenvalues, $A$ typically denotes a matrix, which is a finite, rectangular array of numbers that defines a linear transformation in a finite-dimensional vector space. Eigenvectors are the vectors in the vector space that are scaled by the linear transformation represented by $A$. In finite-dimensional spaces, eigenfunctions are essentially eigenvectors, but the term \textit{eigenfunction} is more commonly used in the context of infinite-dimensional spaces. For example, if $ A $ is an $ n \times n $ matrix, eigenvalues and eigenvectors are solutions to the equation:  

$$
A v = \lambda v, \quad v \neq 0,
$$
where $ v $ is a vector in $ \mathbb{R}^n $ or $ \mathbb{C}^n $. This is usually solved finding values of $\lambda$ that satisfy the \textit{characteristic equation}:\marginnote{There can be multiple eigenvectors corresponding to the same eigenvalue. If $ \lambda>0$ the direction of $v$ remains unchanged, but it is stretched if $|\lambda|>1$ or compressed if $|\lambda|<1$. Eigenvalues can be negative. If $ \lambda<0$ the direction of $v$ is reversed, since multiplication by a negative scalar reflects the vector across the origin.}  

$$
\det(A - \lambda I) = 0,
$$  

where $ I $ is the $ n \times n $ identity matrix. The solutions $ \lambda_1, \lambda_2, \dots, \lambda_n $ are the eigenvalues of $ A $, and for each eigenvalue $ \lambda $, we find the corresponding eigenvector(s) $ v $ by solving the system of linear equations:  

$$
(A - \lambda I)v = 0.
$$

The eigenvalue $ \lambda $ determines how $ A $ stretches or compresses the direction $ v $, which remains unchanged under the transformation, except for sign flips. 

\greyline

\paragraph*{Generalization to Hilbert Spaces: Eigenfunctions and Eigenvalues} Here, we are interested in the generalization from finite-dimensional vector spaces to infinite-dimensional spaces. In this generalized setting, $ A $ is a \textit{linear operator} $ A : V \to V $ which acts on vectors in the Hilbert space $ V $, instead of a matrix. In a finite-dimensional space, a matrix $ A $ maps vectors in $ \mathbb{R}^n $ to $ \mathbb{R}^n $, whereas in an infinite-dimensional space, an operator $ A $ maps functions in a space such as $ L^2$ to itself. The characteristic equation for eigenvectors $ A v = \lambda v $ still applies in this case, but here $ v $ might be a function (hence called an \textit{eigenfunction}), and $ \lambda $ is a scalar \textit{eigenvalue} associated with $ v $.

\greyline

\paragraph*{The Spectral Theorem} The spectral theorem states that self-adjoint operators, both in finite and infinite-dimensional spaces, can be fully diagonalized in terms of their eigenvalues and eigenfunctions. This theorem plays a crucial role in understanding the structure of such operators in Hilbert spaces. Remember that $A$ is {\em self-adjoint} if $A^* = A$, i.e. $\langle Au, v \rangle = \langle u,  Av\rangle$ \quad $\forall u, v \in V$. 

We begin by discussing important properties of self-adjoint operators.

\begin{tcolorbox}[colback=blue!10, colframe=blue!60]
\begin{theorem}[Spectral Theorem for Self-Adjoint Operators]
\label{thrm:eval_real}
Self-adjoint operators have real eigenvalues. 
\end{theorem}
\end{tcolorbox}

\begin{proof}[Proof] 
Let $A v = \lambda v$, with $v \neq 0$. Since $A = A^*$, we have:
$$
\langle A v, v \rangle = \langle v, A v \rangle.
$$
Substituting $A v = \lambda v$, we get:
$$
\langle \lambda v, v \rangle = \langle v, \lambda v \rangle.
$$
Because $\lambda$ is a scalar, we can factor it out of both inner products:
$$
\lambda \langle v, v \rangle = \overline{\lambda} \langle v, v \rangle.
$$
Note that on the right, we have applied conjugate linearity from Section~\ref{subsec:Inner Product}. Since $v \neq 0$, $\langle v, v \rangle > 0$. Thus, we can divide both sides by $\langle v, v \rangle$ to obtain:
$$
\lambda = \overline{\lambda},
$$
which implies that $\lambda \in \mathbb{R}$.
\end{proof}

\begin{tcolorbox}[colback=blue!10, colframe=blue!60]
\begin{theorem}[Orthogonality of Eigenfunctions]\label{thrm:orthogonality_eigenfunctions}
Eigenfunctions of self-adjoint operators corresponding to different eigenvalues are orthogonal.
\end{theorem}
\end{tcolorbox}

\begin{proof}[Proof]
Let $A v = \lambda v$ and $A w = \mu w$ with $\lambda \neq \mu$ and $v, w \neq 0$. Since $A = A^*$, we have:
$$
\langle A v, w \rangle = \langle v, A w \rangle.
$$
Substituting the eigenvalue equations, we get:
$$
\langle \lambda v, w \rangle = \langle v, \mu w \rangle.
$$
Since $\lambda$ and $\mu$ are real (from Theorem~\ref{thrm:eval_real}), we can factor out the scalars without conjugation:
$$
\lambda \langle v, w \rangle = \mu \langle v, w \rangle.
$$
Thus, 
$$
(\lambda - \mu) \langle v, w \rangle = 0.
$$
Since $\lambda \neq \mu$, it follows that:
$$
\langle v, w \rangle = 0,
$$
i.e., $v \perp w$.\marginnote{The set of eigenvalues can be either finite or countably infinite. A set is countable if there is a way to list its elements in a sequence, that is, there is a one-to-one correspondence between the set and the set of natural numbers, $\mathbb{N}$. When we say that the spectrum is discrete we mean that each eigenvalue is separated by some positive distance from others, that is, the eigenvalues are isolated. The only exception is $\lambda=0$. There is no continuous spectrum where eigenvalues can form a continuous range or interval.}
\end{proof} 

\begin{tcolorbox}[colback=blue!10, colframe=blue!60]
\begin{theorem}[Spectral Theorem]
A compact self-adjoint operator $A: V \to V$ has eigenvalues $\{ \lambda \}$ with corresponding eigenfunctions $\{ v_\lambda \}$ such that:
$$
A v_\lambda = \lambda v_\lambda.
$$
These eigenfunctions form an orthonormal basis of $V$, and the set of eigenvalues is countable.  Furthermore, the eigenvalue spectrum is discrete, with the only possible accumulation point being $\lambda = 0$.
\end{theorem}
\end{tcolorbox} 

This statement implies that the eigenvalues of a compact self-adjoint operator form a countable set, all of which are real. The corresponding eigenfunctions form an orthonormal basis of the Hilbert space $V$. If $\lambda \neq 0$, then $\lambda$ is an isolated eigenvalue (discrete spectrum). The only possible accumulation point of the spectrum is $\lambda = 0$.

Thus, the Spectral Theorem builds on the properties established in Theorems~\ref{thrm:eval_real} and~\ref{thrm:orthogonality_eigenfunctions} and provides a complete characterization of the structure of a Hilbert space under a compact self-adjoint operator. \marginnote{Principal Component Analysis (PCA) uses a finite-dimensional version of the Spectral Theorem to identify key directions in data.}

\greyline

\paragraph*{Spectral Theorem Example} In the following, we provide an illustration of the spectral theorem in the context of a differential operator and verify key properties like self-adjointness and orthogonality of eigenfunctions.

Let us work with

$$L^2([-\pi,+\pi]) = \left\{ f: [-\pi,+\pi]:\rightarrow \mathbb{C} \quad \text{s.t.} \quad \int_{-\pi}^{+\pi} |f(x)|^2 dx < \infty \right\},$$ 

the space of square-integrable periodic functions, meaning their squared magnitude integrates to a finite value, with standard inner product $$\langle f, g \rangle = \frac{1}{2\pi} \int_{-\pi}^{+\pi} f(x)\overline{g(x)} dx,$$

where $\overline{g(x)}$ denotes the complex conjugate of $g(x)$.

Consider the Laplacian operator (second-order derivative, see Section~\ref{sec: Vector calculus}): $\Delta = \frac{d^2}{dx^2}.$ First, we verify that $\Delta$ is self-adjoint. To do so, we must show:

$$
\langle \Delta f, g \rangle = \langle f, \Delta g \rangle \quad \forall f, g \in L^2([-\pi, \pi]).
$$

From the product differentiation rule,

$$
\frac{d}{dx} (f(x) g(x)) = f'(x) g(x) + f(x) g'(x).
$$

Also, the fundamental theorem of calculus tells us, \marginnote{Assuming continuity and differentiability, or piecewise smoothness.}

$$
\int_{-\pi}^{+\pi} \frac{d}{dx} (f(x) g(x)) \, dx = \left. f(x) g(x) \right|_{-\pi}^{+\pi},
$$

and given that we are considering periodic functions, we have the boundary conditions $f(\pi)=f(-\pi)$ and $g(\pi)=g(-\pi)$. Hence,

$$
\left. f(x) g(x) \right|_{-\pi}^{+\pi} = f(\pi)g(\pi) - f(-\pi)g(-\pi) = f(\pi)g(\pi) - f(\pi)g(\pi) = 0.
$$

Therefore,

$$
\int_{-\pi}^{+\pi} \frac{d}{dx} (f(x) g(x))  dx = 0 \implies 
\int_{-\pi}^{+\pi} f(x) g'(x) dx = - \int_{-\pi}^{+\pi} f'(x) g(x) dx, 
$$
where for simplicity, we ignore complex conjugates. Applying this result to $f' g'$ we have \marginnote{Let $f$ and $g$ swap roles to obtain both sides of the equation and perform a change of variables.}

\begin{eqnarray*}
- \int_{-\pi}^{+\pi} f'(x) g(x) dx
 = \int_{-\pi}^{+\pi} f'(x) g'(x) dx = - \int_{-\pi}^{+\pi} f(x) g'(x) dx
\end{eqnarray*}
from which self-adjointness $\langle \Delta f, g \rangle = \langle f, \Delta g \rangle$ follows 

$$
\langle \Delta f, g \rangle = \int_{-\pi}^{+\pi} f'(x) g(x) dx = \int_{-\pi}^{+\pi} f(x) g'(x) dx = \langle f, \Delta g \rangle. 
$$

After having verified the self-adjointness of the Laplacian, let us now consider the Laplacian acting on the function $e^{inx}$ where $n\in\mathbb{Z}$. From $\Delta e^{inx} = \frac{d^2}{dx^2} e^{inx}= -n^2 e^{inx},$ \marginnote{$\frac{d}{dx}e^{ax} = ae^{ax}$}it immediately follows that eigenfunctions have the form $e^{inx}$ with corresponding real eigenvalues $-n^2$. Remember that in infinite-dimensional space, eigenfunctions are linear operators: indeed the Laplacian scales linearly the function $e^{inx}$ by a factor of $-n^2$.

In Theorem~\ref{thrm:eval_real} we stated that self-adjoint operators have real eigenvalues: $-n^2$ is real. Next, to verify orthogonality and Theorem~\ref{thrm:orthogonality_eigenfunctions}, write\marginnote{Remember we need to consider the complex conjugate of the eigenfunction: $\overline{e^{imx}} = e^{-imx}$}
\begin{eqnarray*}
\langle e^{inx}, e^{imx}\rangle  =  \frac{1}{2\pi} \int_{-\pi}^{+\pi} e^{inx}e^{-imx} dx = \frac{1}{2\pi} \int_{-\pi}^{+\pi} e^{i(n-m)x} dx.
\end{eqnarray*}

For $n \neq m$ \marginnote{This can be shown using the integral of a complex exponential and rewriting the result in terms of the sine function.},

$$
\int_{-\pi}^{+\pi} e^{i(n-m)x} dx = 0 \implies \langle e^{inx}, e^{imx}\rangle = 0.
$$

This is because the function is periodic with zero average over the full period and shows that distinct eigenfunctions are orthogonal. For $n=m$,

$$
\int_{-\pi}^{+\pi} e^{i(n-m)x} dx = \int_{-\pi}^{+\pi} 1 dx = 2 \pi \implies \langle e^{inx}, e^{inx}\rangle = 1,
$$

which reflects normalization, that is, the eigenfunctions are orthonormal. 

Hence we have that,\marginnote{The Kronecker delta is defined as $$\delta_{nm} =
\begin{cases}
1, & \text{if } n = m, \\
0, & \text{if } n \neq m.
\end{cases}$$}

$$
\langle e^{inx}, e^{imx}\rangle = \delta_{nm},
$$

where $\delta_{mn}$ is the Kronecker delta.

\greyline

\paragraph*{Singular Values} The spectral theorem focuses on self-adjoint operators. For more general operators, we turn to the concept of \textit{singular values} and their corresponding \textit{singular vectors}. Singular values provide a more general way to characterize how an operator transforms vectors in a space, and they are particularly useful when dealing with non-self-adjoint operators, such as general matrices.

Before providing formal definitions, let us clarify the intuitive difference between eigenvalues and singular values. These quantities capture different aspects of how a linear operator transforms elements of the space. Eigenvalues measure how much a transformation stretches or compresses an eigenfunction along its direction, without changing that direction (except for sign flips). On the other hand, singular values measure the overall magnitude of an operator's action, independent of any specific direction, that is, they describe how much the operator stretches or compresses functions in general. These concepts provide fundamental tools for analyzing operators, whether finite-dimensional (as matrices) or infinite-dimensional.

\begin{tcolorbox}[colback=gray!10, colframe=gray!40]
An operator $A : V \rightarrow V$ is  \textit{compact}  iff it can be written in the form 
$$
Aw = \sum_{n\geq 1} \sigma_n \langle v_n, w \rangle u_n, \quad \forall w \in V.
$$

$\{\sigma_n \}_{n\geq 1}$ are the {\em singular values} and $\{v_n\}_{n\geq 1}$, $\{u_n\}_{n\geq 1}$ are the corresponding (left- and right-) {\em singular vectors} of $A$. \marginnote{Singular vectors, both left and right, represent directions in the domain and codomain of $A$.} 
\end{tcolorbox}

Note that this is an alternative definition of compactness. Compact operators are often studied because they have certain nice properties, such as having a countable set of singular values. Importantly, these singular values can accumulate only at zero. This means that after some index $N$, the singular values become zero, indicating that the operator has finite rank. In this case, the rank of $A$ is equal to $N$, and we have:
$$
\text{rank}(A) = N.
$$

In the finite-dimensional case the rank corresponds to the number of linearly independent rows or columns in the matrix representing the operator, whereas in the infinite-dimensional case the rank is the number of non-zero singular values. Even though the operator may act on an infinite-dimensional space, its rank remains finite.

After discussing the most general case, let us now examine particular cases. If $ A $ is self-adjoint, we can write it in the form:
$$
A w = \sum_{n \geq 1} \lambda_n \langle v_n, w \rangle v_n, \quad \forall w \in V.
$$
Here, $ \{ \lambda_n \} $ are the eigenvalues of $ A $, and $ \{ v_n \} $ are the corresponding eigenvectors of $ A $. This is a special case of the more general singular value decomposition, where the singular values coincide with the eigenvalues, and the left and right singular vectors are the same.

Next, let us discuss singular value decomposition (SVD) of matrices.

\begin{tcolorbox}[colback=gray!10, colframe=gray!40]
An $ m \times n $ matrix $ A $ can be written in the \textit{singular value decomposition (SVD)} form:
$$
A = U \Sigma V^* = \begin{pmatrix} | & & | \\ u_1 & \dots & u_n \\ | & & | \end{pmatrix}
\begin{pmatrix} \sigma_1 & & \\ & \ddots & \\ & & \sigma_n \end{pmatrix}
\begin{pmatrix} - & \overline{v}_1 & - \\ & \vdots & \\ - & \overline{v}_n & - \end{pmatrix},
$$
where $U$ is an $ m \times m $ unitary matrix whose columns are the \textit{left singular vectors} $ u_i $, $ \Sigma $ is an $ m \times n $ diagonal matrix containing the singular values $ \sigma_i $, and $ V^* $ is the conjugate transpose of the $ n \times n $ unitary matrix $ V $, whose rows are the \textit{right singular vectors} $ \overline{v}_i $.
\end{tcolorbox}

\greyline

\paragraph*{Example Eigenvalues vs Singular Values} To further build on our intuition regarding the difference between eigenvalues and singular values, let us consider a rotation matrix. A rotation matrix has no real eigenvalues because it does not stretch or compress space along fixed directions. However, it has singular values all equal to $1$, reflecting that it preserves lengths. More concretly, a rotation matrix $ R $ in 2D is defined as:

$$
R = \begin{pmatrix}
\cos \theta & -\sin \theta \\
\sin \theta & \cos \theta
\end{pmatrix},
$$

where $ \theta $ is the rotation angle. To find the eigenvalues, we solve the characteristic equation:

$$
\lambda^2 - 2\lambda\cos \theta + 1 = 0.
$$

Thus, the eigenvalues are:

$$
\lambda = e^{i\theta}, \quad \lambda = e^{-i\theta}.
$$

These eigenvalues are complex and lie on the unit circle in the complex plane. Hence, there are no real eigenvalues unless $\theta = 0$ or $\pi$ (identity and reflection).

The singular values of $ R $ are obtained from the eigenvalues of $ R^T R $:

Multiplying $ R^T R $:

$$
R^T R = \begin{pmatrix}
\cos \theta & \sin \theta \\
-\sin \theta & \cos \theta
\end{pmatrix}
\begin{pmatrix}
\cos \theta & -\sin \theta \\
\sin \theta & \cos \theta
\end{pmatrix}= \begin{pmatrix}
1 & 0 \\
0 & 1
\end{pmatrix} = I.
$$

The eigenvalues of $ R^T R $ are therefore both $ 1 $, and the singular values of $ R $ (the square roots of these eigenvalues) are:
$\sigma_1 = 1$ and $\sigma_2 = 1.$

\greylinelong

\subsection{Fourier analysis}
\label{subsec:Fourier_analysis}

In eigenfunctions and eigenvalues, singular value decomposition, and Fourier series, the fundamental concept is the decomposition of an object—whether a self-adjoint operator, an operator, or a function—into a sum of components along specific directions or bases. More commonly, Fourier series are associated with a trigonometric basis (sine, cosine, or complex exponential). However, the concept is general and applies to any orthonormal basis.

\begin{tcolorbox}[colback=gray!10, colframe=gray!40]
Let $\{v_\alpha\}$ be an orthonormal basis in a Hilbert space $V$. Then, $u \in V$ can be expressed as a {\em Fourier series}
$$
u = \sum_{\alpha} \langle u, v_\alpha \rangle v_\alpha
$$
The coefficients $\langle u, v_\alpha \rangle = \hat{u}_\alpha$ in the above series are called {\em Fourier coefficients} (or {\em transforms}) of $u$.
\end{tcolorbox}

For clarity, remember that the expression above can be expanded as follows:

$$
u = \sum_{\alpha} \langle u, v_\alpha \rangle v_\alpha = \hat{u}_{\alpha}v_{\alpha} = \hat{u}_1 v_1 + \hat{u}_2 v_2 + \hat{u}_3 v_3 + ...
$$

\greyline

\paragraph*{Fourier Decomposition for Vectors} For vectors, the Fourier decomposition can be written in matrix form:

$$
u = 
\underbrace{\begin{pmatrix}
| & & | \\
v_1 & \cdots & v_n \\
| & & | 
\end{pmatrix}}_{V}
\underbrace{\begin{pmatrix}
- & \overline{v}_1^\top & - \\
& \vdots & \\
- & \overline{v}_n^\top & - 
\end{pmatrix}}_{V^\dagger}
u,
$$

where $V \in \mathbb{C}^{n \times n}$ is the matrix whose columns are the basis vectors $v_i$, $V^\dagger$ is the Hermitian conjugate (conjugate transpose) of $V$, and $V^\dagger u = (\langle u, v_1 \rangle, \langle u, v_2 \rangle, \ldots, \langle u, v_n \rangle)^\top$ contains the Fourier coefficients. Thus
$$
u = V (V^\dagger u),
$$

where $V^\dagger u$ gives the coefficients, and $V (V^\dagger u)$ reconstructs the vector. From this, it is evident that it is a unitary operation (see below). \marginnote{A unitary operation is a linear operation that preserves the inner product in a complex vector space.}

\greyline

\paragraph{Continuous Fourier Transform} Note that in general, $\alpha$ here can be a continuous index, in which case the sum should be replaced with an integral:

$$\langle u, v_\alpha \rangle = \int u(x) \overline{v_\alpha(x)} \, dx.$$

This is the case with the continuous \textit{Fourier transform} using a basis of the form $e^{i\omega x}$, with $\omega \in \mathbb{R}$:

$$
   f(x) = \int_{-\infty}^{\infty} \hat{f}(\omega) e^{i\omega x} \, d\omega,
$$

where $\hat{f}(\omega)$ are the Fourier coefficients of $f(x)$, representing the contribution of each frequency component. The Fourier coefficients are obtained based on the inner product:

$$\hat{f}(\omega) = \int_{-\infty}^\infty f(x) e^{-i\omega x} \, dx.$$

Note that the computations in the continuous case are analogous to obtaining the coefficients and reconstructing the vector using matrix multiplication, as discussed earlier in the context of vectors.

\greyline

\paragraph*{Fourier Series Example} Consider $L^2([-\pi,+\pi])$, the space of square-integrable periodic functions, with the standard inner product $\langle f, g \rangle = \frac{1}{2\pi} \int_{-\pi}^{+\pi} f(x)\overline{g(x)} dx$ and the basis $\{e^{inx}\}_{n\geq 1}$. 
The Fourier series assume the classical form 
$$
f(x) = \sum_{n\geq 1} \frac{1}{2\pi} \int_{-\pi}^{+\pi} f(y) e^{-iny} dy \, e^{inx}.
$$

The Fourier series provides a discrete decomposition because the function being considered is periodic, leading to discrete frequencies.

\greyline

\paragraph*{Parseval's Identity} Parseval's identity establishes that the inner product, and hence the geometry, of a Hilbert space $V$ is perfectly captured by the Fourier coefficients. The identity guarantees that this mapping is an isometry, and it allows us to work with Fourier coefficients as a proxy for the original function or vector.

\begin{tcolorbox}[colback=blue!10, colframe=blue!60]
\begin{theorem}[Parseval's identity]
Let 
$
u = \sum_\alpha \hat{u}_\alpha v_\alpha \quad \text{and} \quad
w = \sum_\alpha \hat{w}_\alpha v_\alpha 
$
be Fourier series of $u, w \in V$ with respect to the orthonormal basis $\{ v_\alpha \}$. 
Then $\langle u, w \rangle = \sum_\alpha \hat{u}_\alpha \overline{\hat{w}}_\alpha$. 
\end{theorem}
\end{tcolorbox}

In other words, we can define a map $V \ni u \mapsto \hat{u} = \{ \langle u, v_\alpha\rangle \} \in \ell^2$ from vectors to (square summable) sequences. This map is an isometry \marginnote{Recall that an isometry is length-preserving.}: 
$$
\| u\|^2_V = \sum_\alpha | \langle u, v_\alpha\rangle |^2 = \sum_\alpha |\hat{u}_\alpha |^2 = \| \hat{u} \|^2_{\ell^2}. 
$$
This, in turn, is nothing else but the application of the Pythagorean theorem, Theorem~\ref{theorem:pythagoras} (possibly in infinite dimensions),
$$
\| u\|^2 = \Big\| \sum_\alpha \langle u, v_\alpha \rangle v_\alpha \Big\|^2 = \sum_\alpha \|  \langle u, v_\alpha \rangle v_\alpha \|^2 = \sum_\alpha  | \langle u, v_\alpha\rangle |^2,
$$
where we used the orthonormality of the basis $\{ v_\alpha \}$.

\greyline

\paragraph*{The Heat Equation}

Consider the following partial differential equation, called the {\em heat equation}, under Dirichlet boundary conditions:
$$
\left\{
\begin{array}{lc}
\Delta f(x,t) = f_t(x,t) & \\
f(x,0) = g(x) & \text{(initial conditions)}
\end{array}
\right.
$$
on a circle, where $ f : S^1 \times [0, \infty) \rightarrow \mathbb{R} $ (periodic in the first coordinate) represents the temperature \marginnote{$f(x,t)$ is the temperature at point $x$ at time $t$.}, $\Delta = \frac{\partial^2}{\partial x^2}$ is the one-dimensional Laplacian operator, and $g(x)$ is the initial temperature distribution at time $t=0$. Since $S^1$ is a circle, there are no boundary conditions on the spatial domain.

Fourier analysis was originally developed for solving this kind of partial differential equation (PDE), and we will show how it applies here. First, assume the solution has a separable form: 
$$
f(x,t) = X(x) T(t),
$$
where $X(x)$ is the spatial part and $T(t)$ is the temporal part. Assuming $X, T$ never vanish, we substitute this into the heat equation:
$$
\Delta f - \frac{\partial}{\partial t} f = X' T - X T' = 0.
$$
Since the above holds for any $(x,t)$, it follows that:
\begin{eqnarray*}
\frac{X'}{X} = \frac{T'}{T} = -\lambda \quad \text{(some constant)}.
\end{eqnarray*}
In other words, the spatial and temporal parts of the solution are eigenfunctions of the Laplacian and first-order derivative operators, respectively:
$$
X' = \Delta X = -\lambda X, \quad T' = \frac{\partial}{\partial t} T = -\lambda T,
$$
which we can express in closed form as:
$$
\Delta e^{inx} = -n^2 e^{inx}, \quad \frac{\partial}{\partial t} e^{-n^2 t} = -n^2 e^{-n^2 t},
$$
where $\lambda = -n^2$ is the corresponding eigenvalue.

Hence, solutions to the equation take the form $f_n(x,t) = e^{inx} e^{-n^2 t}$. Due to the linearity of the equation, any linear combination of such solutions is also a solution, so the general solution can be written as:
$$
f(x,t) = \sum_{n=-\infty}^{\infty} a_n e^{inx} e^{-n^2 t}.
$$
Note that we sum over all integer values of $n$ (including both positive and negative values) to account for the full Fourier expansion.

To find a unique solution, we must use the initial condition. The set $\{ e^{inx} \}_{n \in \mathbb{Z}}$ forms an orthonormal basis for $L^2(S^1)$. Therefore, we can express the initial condition $g(x)$ as a Fourier series:
\marginnote{Note that $g(x)$ does not depend on time, so we use the eigenfunctions of the Laplacian.}
$$
g(x) = \sum_{n=-\infty}^{\infty} \langle g, e^{inx} \rangle e^{inx},
$$
where $\langle g, e^{inx} \rangle$ is the Fourier coefficient for $g(x)$.

Since $f(x,0) = g(x)$, we can identify $a_n = \hat{g}_n = \langle g, e^{inx} \rangle$. Using the standard inner product for periodic functions, we obtain the general solution:
\begin{eqnarray*}
f(x,t) &=& \sum_{n=-\infty}^{\infty} \frac{1}{2\pi} \int_{-\pi}^{\pi} g(y) e^{-iny} \, dy \, e^{inx} e^{-n^2 t} \\
&=& \frac{1}{2\pi} \int_{-\pi}^{\pi} g(y) \underbrace{\sum_{n=-\infty}^{\infty} e^{-n^2 t} e^{-in(x-y)}}_{h_t(x-y)} \, dy = g \star h_t,
\end{eqnarray*}
where $\star$ denotes convolution.

The function $h_t(x)$ is called the {\em fundamental solution} of the heat equation, or the {\em heat kernel}. In particular, for the case where the initial condition is the Dirac delta function, $g(x) = \delta(x)$ (an impulse initial condition), we have: \marginnote{
The Dirac delta function is analogous to the Kronecker delta but in the continuous case. It is defined as:

$$
\delta(x - y) =
\begin{cases}
\infty, & \text{if } x = y, \\
0, & \text{if } x \neq y,
\end{cases}
$$
with the important property that its integral over the entire real line is equal to 1:

$$
\int_{-\infty}^{\infty} \delta(x - y) \, dy = 1.
$$

In the context of Fourier analysis, the Dirac delta function can be represented as:

$$
\delta(x - y) = \sum_{n=-\infty}^{\infty} e^{in(x - y)}.
$$

The Dirac delta function acts as an identity element in the Fourier transform, meaning that for any function $f(x)$:

$$
\int_{-\infty}^{\infty} f(y) \delta(x - y) \, dy = f(x).
$$

}
$$
\langle \delta, e^{inx} \rangle = e^{in0} = 1, \quad \text{so} \quad a_n = 1 \quad \forall n,
$$
which implies that the solution is:
$$
f(x,t) = \sum_{n=-\infty}^{\infty} e^{-n^2 t} e^{inx} = h_t(x).
$$
In signal processing terms, $h_t$ is referred to as the {\em impulse response} of the system.

\greyline

\paragraph*{A Short Note on Wavelets} Wavelets are a generalization of Fourier transforms. While Fourier transforms decompose functions into globally defined sinusoidal components, wavelets decompose functions using basis functions (often orthogonal, but not exclusively so) that are localized in both time and frequency. This localization enables wavelets to represent transient and hierarchical features in data. Before the advent of AlexNet in 2012 and the rise of deep learning, wavelet transforms were widely used in computer vision and signal processing due to their ability to simultaneously capture spatial and frequency information, making them particularly effective for tasks such as image compression, denoising, and texture analysis.

\begin{tcolorbox}[colback=orange!20, colframe=orange!60]
\textbf{Spectral Theory in Geometric Deep Learning.} Spectral theory provides a mathematically rigorous framework for extending traditional Deep Learning approaches for Euclidean data to irregular domains such as graphs and manifolds while maintaining important properties like translation invariance and locality.
\end{tcolorbox}

\clearpage

\section{Graph Theory}
\label{Graph Theory}

While\marginnote{Note however that Geometric Deep Learning is a broader framework that extends Deep Learning techniques to non-Euclidean domains, with one such instantiation being learning over graphs.} continuous geometry might examine smooth curves or surfaces, discrete geometry focuses on structures that can be enumerated or broken down into distinct, countable elements. Graph theory~\cite{Wilson1972IntroductionTG} is a subset of discrete geometry that is central to GNNs, which are perhaps the quintessential artificial neural network architecture in Geometric Deep Learning.

\greylinelong

\subsection{Preliminaries on Graphs and Notation}
\label{subsec: Preliminaries on graphs and notation}

We start by discussing basic definitions and notation to describe graphs.\marginnote{The famous Königsberg Bridge Problem was solved by Euler in 1736 and is one of the earliest examples of graph theory. It is also deeply connected to topology.}

\begin{tcolorbox}[colback=gray!10, colframe=gray!40] A \textit{graph} is an ordered tuple:

$$
G = (V, E),
$$

where $V$ is a set of nodes (or vertices), and $E \subseteq (V \times V)$ is a 2-tuple set representing the edges (or links) in the graph. 
\end{tcolorbox}

\marginnote{J. Sylvester mentions the term `graph' as early as 1878 in a chemical context.

\includegraphics[width=\linewidth]{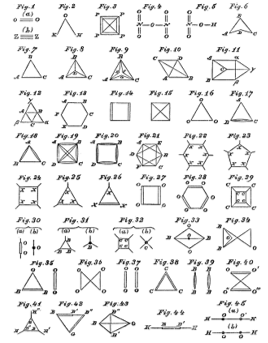}}Edges may be directed or undirected. Directed edges are uni-directional relations from a source node $v_i$ to a target node $v_j$; thus, $(v_i, v_j) \in E$, and importantly, $(v_i, v_j) \neq (v_j, v_i)$. 

\begin{tcolorbox}[colback=gray!10, colframe=gray!40]
A \textit{directed graph} (or \textit{digraph}) is a graph $G = (V, E)$ where each edge in $E$ is an ordered pair of nodes.
\end{tcolorbox}

In contrast, undirected edges are bidirectional, so $(v_i, v_j) = (v_j, v_i)$. When an edge connects a node to itself, we call it a self-loop $(v_i, v_i)$. 

\begin{tcolorbox}[colback=gray!10, colframe=gray!40] 
The (one-hop) \textit{neighborhood} of a node $v_i$ is the set of nodes that share an edge with $v_i$, denoted as $$\mathcal{N}(v_i) = \mathcal{N}_i = \{v_j | (v_i, v_j) \in E\}.$$
\end{tcolorbox}

\begin{tcolorbox}[colback=gray!10, colframe=gray!40]
A \textit{subgraph} $H = (V_H, E_H)$ of a graph $G = (V_G, E_G)$ is a graph where $V_H \subseteq V_G$ and $E_H \subseteq E_G$.
\end{tcolorbox}

If we consider the set $\{v_i\} \cup \mathcal{N}(v_i)$ as nodes and include all edges in $E$ that connect these nodes, this defines a \textit{neighborhood subgraph} of $v_i$, which is a subgraph of $G$.

\begin{figure}[htbp!]
  \centering
  \includegraphics[width=0.3\linewidth]{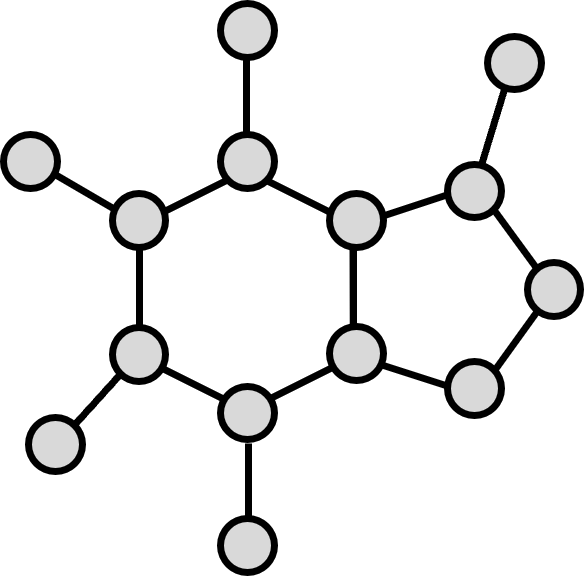}
  \caption{Diagram of a graph with nodes in gray and edges in black.}
  \label{graph}
\end{figure}

\greyline
\paragraph*{The Adjacency Matrix} Graphs can be represented using matrices. For a graph with $N=|V|$ number of nodes, its adjacency matrix $A \in \mathbb{R}^{N \times N}$ represents the connectivity structure between nodes. $A$ can be weighted or unweighted. If it is weighted, its entries $A_{ij}\in\mathbb{R}$ represent the weight or strength of the connection, and if $(v_i, v_j) \notin E$, then $A_{ij} = 0$. $ w: E \rightarrow \mathbb{R}^+ $ is the weight function assigning positive real numbers to edges: if $e=(v_i,v_j)$, then $w(e)=A_{ij}$. In the case of an unweighted adjacency matrix, $A_{ij} = 1$ when there is an edge and $A_{ij} = 0$ when there is no edge. So that,

$$
    A_{ij}=\begin{cases}
    1 & \text{if $(v_i,v_j) \in E$} \\
    0 & \text{if $(v_i,v_j) \notin E$}.
    \end{cases}
$$

Hence, if the graphs' edges are unweighted and undirectional, the corresponding adjacency matrix is binary and symmetric. On the other hand, the adjacency matrix of a digraph is generally asymmetric, since $A_{ij} \neq A_{ji}$ in the case of directed edges. Lastly, the diagonal degree matrix $D\in \mathbb{R}^{N \times N}$ is defined as the matrix where each entry on the diagonal is the row-sum of the adjacency matrix: $D_{ii}=\sum_jA_{ij}$, which is also symmetric for undirected graphs.

For example, we can number the nodes of an undirected infinite binary tree in \emph{level order}. Let $V=\{v_1,v_2,v_3,\dots\}$, where $v_1$ is the root and for each $v_i$, its left child is $v_{2i}$ and its right child is $v_{2i+1}$. For $v_1$ we have $i=1$, the left child is $v_{2\cdot 1} = v_2$, and the right child is $v_{2\cdot 1 + 1} = v_3$. Likewise for $v_3$, $i=3$, and hence its left child is $v_{2\cdot 3} = v_6$ and its right child is $v_{2\cdot 3 + 1} = v_7$. In summary, the weight function is defined as
\[
w(v_i,v_j) = w(v_j,v_i) =
\begin{cases}
1, & \text{if } \{i,j\} \text{ is a parent-child pair (i.e., } j = 2i \text{ or } j = 2i+1\text{)},\\[1mm]
0, & \text{otherwise}.
\end{cases}
\]
This weight function defines the entries of the adjacency matrix $A$ of the infinite binary tree $A_{ij} = A_{ji} = w(v_i,v_j).$ Since the tree is infinite, the full adjacency matrix is an infinite matrix too:
\[
A = \begin{pmatrix}
0 & 1 & 1 & 0 & 0 & 0 & 0 & \cdots \\
1 & 0 & 0 & 1 & 1 & 0 & 0 & \cdots \\
1 & 0 & 0 & 0 & 0 & 1 & 1 & \cdots \\
0 & 1 & 0 & 0 & 0 & 0 & 0 & \cdots \\
0 & 1 & 0 & 0 & 0 & 0 & 0 & \cdots \\
0 & 0 & 1 & 0 & 0 & 0 & 0 & \cdots \\
0 & 0 & 1 & 0 & 0 & 0 & 0 & \cdots \\
\vdots & \vdots & \vdots & \vdots & \vdots & \vdots & \vdots & \ddots
\end{pmatrix}.
\]

\greyline

\paragraph*{Graph Connectivity} Whether a graph is connected or not determines if information propagation across all vertices of the graph is possible.\marginnote{Convolutional Neural Networks operate on images and preserve the connectivity equivalent to that of a grid.

\includegraphics[width=0.75\linewidth]{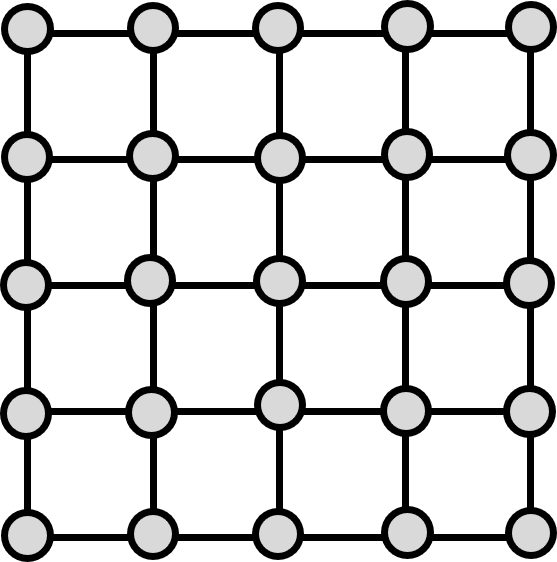}
}

\begin{tcolorbox}[colback=gray!10, colframe=gray!40]
A graph $G=(V,E)$ is said to be \textit{connected} if there is a path between every pair of nodes in the graph. In other words, for any two nodes $ v_i $ and $ v_j $, there exists a sequence of edges $ e_1, e_2, \dots, e_k \in E $ such that $ v_i $ and $ v_j $ are endpoints of this sequence.
\end{tcolorbox}

Conversely, in a \textit{disconnected} graph, there exist pairs for which no such path exists.

At the node level, degree centrality is a measure of the importance or influence of a node in a graph based on its connectivity.

\begin{tcolorbox}[colback=gray!10, colframe=gray!40] 
The \textit{degree centrality} of a node measures the number of direct connections a node has. In an undirected graph, the degree $ d(v_i) $ of a node $ v_i $ is simply the number of edges connected to it:
\vspace{-5pt}
\[
deg(v_i) = \sum_j A_{ij} = \sum_j A_{ji}.
\]

In directed graphs, the \textit{in-degree} and \textit{out-degree} are defined as the number of incoming and outgoing edges, respectively:
\[
deg_{in}(v_i) = \sum_j A_{ji}, \quad deg_{out}(v_i) = \sum_j A_{ij}.
\]
\end{tcolorbox}

Intuitively, a node with high degree centrality is likely to have smaller shortest path distances to other nodes. 

\begin{tcolorbox}[colback=gray!10, colframe=gray!40]
The \textit{shortest path (graph geodesic) distance} between two nodes $ v_i, v_j \in V $ in a weighted graph $ G = (V,E) $, denoted $ d_{G}(v_i, v_j) $, is the minimum total weight of any path connecting these nodes. Formally, for a path $ P = (e_1, \ldots, e_k) $ where $ e_i \in E $, we define:
\[
d_{G}(v_i, v_j) = \min_{P \in \mathcal{P}_{ij}} \sum_{e_k \in P} w(e_k)
\]
where $ \mathcal{P}_{ij} $ is the set of all paths from $ v_i $ to $ v_j $ in $ G $, and $ w: E \rightarrow \mathbb{R}^+ $ is the weight function assigning positive real numbers to edges.
\end{tcolorbox}

Consider a weighted graph $G$ with the node set $V=\{v_1, v_2, v_3, v_4\},$ and weighted edges defined by $w(v_1,v_2)=2, w(v_1,v_3)=4, w(v_2,v_3)=1, w(v_3,v_4)=3$.
For all other pairs of nodes the weights are $0$. This weight function is reflected in the adjacency matrix $A\in \mathbb{R}^{4\times 4}$, where each entry is given by
\[
A_{ij}=\begin{cases}
w(v_i,v_j) & \text{if } (v_i, v_j)\in E,\\[1ex]
0 & \text{if } (v_i, v_j)\notin E.
\end{cases}
\]
In our example, the explicit adjacency matrix is:
\[
A =
\begin{pmatrix}
0 & 2 & 4 & 0 \\
0 & 0 & 1 & 0 \\
0 & 0 & 0 & 3 \\
0 & 0 & 0 & 0
\end{pmatrix}.
\]

The possible paths from $v_1$ to $v_4$ are: $P_1: v_1 \rightarrow v_2 \rightarrow v_3 \rightarrow v_4$, with total weight $w(P_1)=w(v_1,v_2)+w(v_2,v_3)+w(v_3,v_4)=2+1+3=6.$ $P_2: v_1 \rightarrow v_3 \rightarrow v_4$, with total weight $w(P_2)=w(v_1,v_3)+w(v_3,v_4)=4+3=7.$ Thus, the shortest path (graph geodesic) distance between $v_1$ and $v_4$ is $d_G(v_1,v_4)=\min\{6,7\}=6.$

If no path exists between $ v_i $ and $ v_j $, we define the shortest path to be $ d_{G}(v_i, v_j) = \infty.$ \marginnote{For optimization purposes alternative definitions of the distance between disconnected nodes may be more appropiate than using $\infty$.} For unweighted graphs, the distance equals the minimum number of edges in any path between the nodes. Note that the shortest path distance induces a metric space $(V,d_{G})$ over the vertex set of the graph $G$.

The diameter of a graph is the longest shortest path between any two nodes in a graph.

\begin{tcolorbox}[colback=gray!10, colframe=gray!40]
The \textit{diameter} $ \text{diam}(G) $ is defined as the maximum value of the shortest path distances between all pairs of nodes:
\[
diam(G) = \max_{v_i, v_j \in V} d_G(v_i, v_j),
\]
where $ d(v_i, v_j) $ is the shortest path distance between nodes $ v_i $ and $ v_j $.
\end{tcolorbox}

Let us compute the diameter for an undirected graph $G$ with nodes $V=\{v_1,v_2,v_3,v_4\}$ and weighted edges $w(v_1,v_2)=2, w(v_1,v_3)=4, w(v_2,v_3)=1, w(v_3,v_4)=3.$ For any pair of nodes that are not directly connected, we set the weight to $0$. First we must compute the shortest paths between nodes. Between $v_1$ and $v_2$ we have $d_G(v_1,v_2)=2$ since there is a direct edge. Between $v_1$ and $v_3$, there is a direct edge with weight $4$, but we also have the path $v_1 \rightarrow v_2 \rightarrow v_3$ with total weight $2+1=3.$ Hence, $d_G(v_1,v_3)=\min\{4,3\}=3.$ Between $v_1$ and $v_4$, there are two possible paths: $v_1 \rightarrow v_3 \rightarrow v_4$, with weight $3+3=6$ (using the shorter $v_1 \to v_3$ path computed above), and $v_1 \rightarrow v_2 \rightarrow v_3 \rightarrow v_4$, with weight $2+1+3=6$. Thus, $d_G(v_1,v_4)=\min\{6,6\}=6.$ Similarly, we have $d_G(v_2,v_3)=1,\quad d_G(v_2,v_4)=d_G(v_2,v_3)+d_G(v_3,v_4)=1+3=4,\quad \text{and} \quad d_G(v_3,v_4)=3.$ Therefore, the pairwise distances (ignoring the trivial zero distances from a node to itself) are $\{2,\,3,\,6,\,1,\,4,\,3\},$ and given that the diameter of a graph is defined as the maximum shortest path distance between any two nodes we obtain: $\text{diam}(G) = \max\{2,\,3,\,6,\,1,\,4,\,3,\,0\}=6.$

\greyline

\marginnote{The term point cloud is often associated with points (or nodes) having coordinates in $\mathbb{R}^2$ or $\mathbb{R}^3$, while the term null graph is more commonly used in graph theory textbooks to refer to graphs without feature vectors. 

\includegraphics[width=0.75\linewidth]{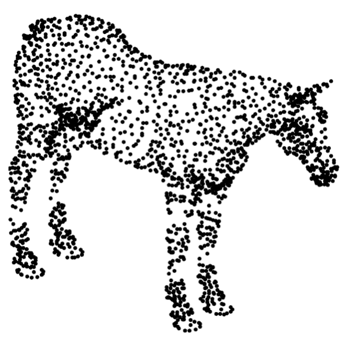}
However, in the GNN literature, point clouds do not necessarily have spatial coordinates.} \paragraph*{Types of Graphs} Next, we discuss important types of graphs based on their connectivity structures, or \textit{graph topology}. At one extreme, we can consider graphs that are completely disconnected, known as point clouds. These are actually common in many applications, such as remote sensing technology and surface reconstruction.

\begin{tcolorbox}[colback=gray!10, colframe=gray!40]
A \textit{point cloud} (or \textit{null graph} $N_N$, where the subscript stands for $N=|V|$) is a graph $G=(V,E)$ whose edge set is the empty set $E=\emptyset$.
\end{tcolorbox}

At the other end of the spectrum, we have complete graphs, which represent the maximum possible number of edges in a graph with $N$ vertices, where every vertex is directly connected to every other vertex.

\begin{tcolorbox}[colback=gray!10, colframe=gray!40]
A \textit{complete graph} is a graph in which every pair of distinct vertices is connected by a unique edge. A complete graph with $ N $ vertices is denoted $ K_N $.
\end{tcolorbox}

Thus, in a complete graph there are no disconnected components and all vertices are reachable from each other, with a graph geodesic distance equal to 1 for unweighted graphs.\marginnote{The ubiquitous attention mechanism in Transformers performs computations over a complete graph, where $N$ is the number of tokens in the context window.}

\begin{tcolorbox}[colback=gray!10, colframe=gray!40]
A \textit{bipartite graph} \( G = (V, E) \) consists of a set of vertices \( V \), which can be partitioned into two disjoint subsets \( V_1 \) and \( V_2 \), such that \( V = V_1 \cup V_2 \) and \( V_1 \cap V_2 = \emptyset \), and a set of edges \( E \subseteq \{ \{u, v\} \mid u \in V_1, v \in V_2 \} \), meaning that edges only connect vertices in \( V_1 \) to vertices in \( V_2 \).
\end{tcolorbox}

In simpler terms, a bipartite graph is a graph in which the vertices can be divided into two disjoint sets, such that no two vertices within the same set are adjacent, and edges connect only vertices from different sets. Bipartite graphs are commonly used for modeling in recommendation systems and for matching products to users.

\greyline

\paragraph*{Paths and Cycles} Next, we discuss paths and cycles as graph substructures.

\begin{tcolorbox}[colback=gray!10, colframe=gray!40]
A \textit{path graph} \marginnote{Path graphs can represent linear sequences or chains in networks.

\includegraphics[width=1\linewidth]{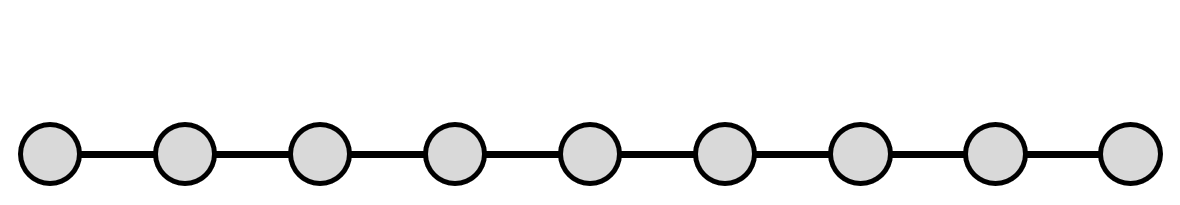}} is a graph where the vertices are arranged in a linear sequence, such that each vertex is connected to at most two others. A path graph with $N$ vertices is denoted $P_N$.
\end{tcolorbox}

$P_N$ consists of $N$ vertices and $N-1$ edges, where the endpoints (also called leaves) have degree 1, and all other vertices have degree 2. For instance, consider the vertex set \(V = \{1, 2, \dots, N\}\), where each vertex corresponds to an element of \(\mathbb{N}\) and the edge set is \(E = \{(v_i, v_{i+1}) \mid i \in \{1, 2, \dots, N-1\}\}\), representing the connections between consecutive numbers. This construction discretizes the natural numbers by treating them as evenly spaced points on a line.

A cycle in a graph is a path that starts and ends at the same node, with all intermediate vertices being distinct. An acyclic graph is one that does not contain any cycles (or closed loops).

\begin{tcolorbox}[colback=gray!10, colframe=gray!40]
A \textit{cycle graph} \marginnote{The circular structure of a cycle graph can be used to represent periodic phenomena.} is a graph that consists of a single cycle, where each vertex is connected to exactly two others, forming a closed loop. A cycle graph with $ N $ vertices is denoted $ C_N $.
\end{tcolorbox}

\begin{tcolorbox}[colback=gray!10, colframe=gray!40]
A \textit{directed acyclic graph (DAG)} \marginnote{DAGs are often used to describe causality.}is a directed graph that contains no cycles. In a DAG, the edges have a direction, and there is no directed path that leads back to the starting node.
\end{tcolorbox}

\begin{tcolorbox}[colback=gray!10, colframe=gray!40]
A \textit{tree} is a connected, acyclic graph where there is exactly one path between any two nodes. It has $ |V| - 1 $ edges for $ |V| $ vertices.
\end{tcolorbox}

A directed tree is a type of DAG, but trees can also be undirected. \marginnote{Trees have negative curvature and exhibit exponential volume growth.

\includegraphics[width=0.75\linewidth]{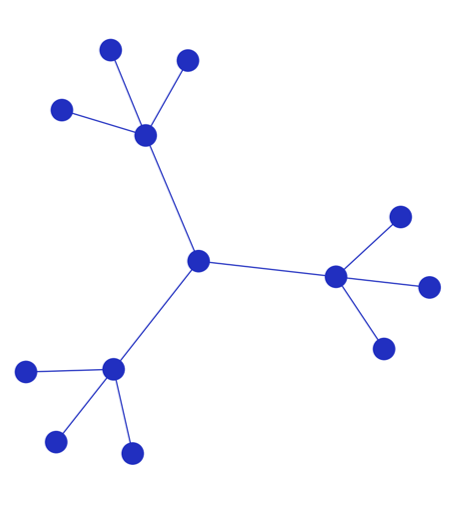}
}

\greyline

\paragraph*{Regular Graphs} In many applications where the underlying graph connectivity is unknown, such as in latent graph inference and bioinformatics, one assumes the underlying graph to be regular.

\begin{tcolorbox}[colback=gray!10, colframe=gray!40]
A \textit{regular graph} is a graph where every vertex has the same degree. If each vertex has degree $k$, the graph is called \textit{$k$-regular}.
\end{tcolorbox}

\begin{itemize}
    \item The \textit{null graph} $N_N$, which is $0$-regular (no edges).
    \item The \textit{cycle graph} $C_N$, which is $2$-regular.
    \item The \textit{complete graph} $K_N$, which is $(N-1)$-regular.
    \item \textit{Cubic graphs}, a special class of $3$-regular graphs, such as the Petersen graph\marginnote{The Petersen graph is a 10-vertex, 15-edge undirected graph that plays a prominent role in graph theory, often used as a key example or counterexample in various problems.}.
\end{itemize}

\greyline

\paragraph*{Geometric Graphs} In geometric graphs nodes are represented as points in Euclidean space and their relationships are often defined based on distance or some other notion of geometric proximity according to the space's metric.\marginnote{Proximity is used to infer the graph connectivity of molecules based on electron cloud images obtained through X-ray crystallography.}

\begin{figure}[htbp!]
  \centering
  \includegraphics[width=0.35\linewidth]{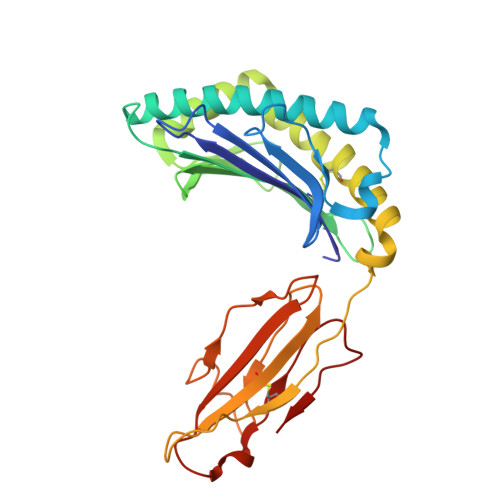}
  \caption{Geometric graphs can be used as mathematical abstractions of biomolecules.}
  \label{bio}
\end{figure}

\begin{tcolorbox}[colback=gray!10, colframe=gray!40]
A \textit{geometric graph} $ G = (V, E) $ is a graph where each node $ v_i \in V $ is associated with a point in a geometric space, typically $ \mathbb{R}^2$ or $\mathbb{R}^3$, and edges $ (v_i, v_j) \in E $ are determined by the positions of the nodes.
\end{tcolorbox}

As discussed in the preliminaries in Section~\ref{subsec: Preliminaries on graphs and notation}, connections between nodes are represented by an adjacency matrix, but they also have geometric positions (e.g., atoms in 3D) and geometric features (e.g., velocities).

\marginnote{k-NN type properties might be desirable if the graph's density is intended to remain consistent, as it can also prevent the occurrence of disconnected components. However, it imposes constraints on the graph's connectivity structure and may result in connections between nodes that are unreasonably distant.}Often, in geometric graphs we use the \textit{unit disk graph} approach where edges \( (v_i, v_j) \in E \) are included if the distance \( d(v_i, v_j) \) between nodes \( v_i \) and \( v_j \) is less than or equal to a fixed threshold \( \epsilon \), i.e., \( d(v_i, v_j) \leq \epsilon \).

An alternative approach is to use \textit{k-nearest neighbor (k-NN) graphs}. In a k-NN graph, each node is connected to its $k$-closest neighbors in the geometric space, based on the distance metric $ d(v_i, v_j) $. This method does not rely on a fixed threshold, but instead ensures that each node is connected to exactly $ k $ other nodes, that is, it is a k-regular graph. 

\greyline

\paragraph*{Homophily and Heterophily} We can assign class labels $y_i$ to each node $v_i$\marginnote{It is also possible to assign labels at the graph or edge level.}. Most real-world graph datasets adhere to the principle of homophily, where connected nodes tend to belong to the same class. For example, in citation networks, similar research works cite each other. Homophily can be calculated as the fraction of intra-class graph edges:\marginnote{Example highly homophilic graph.

\includegraphics[width=1\linewidth]{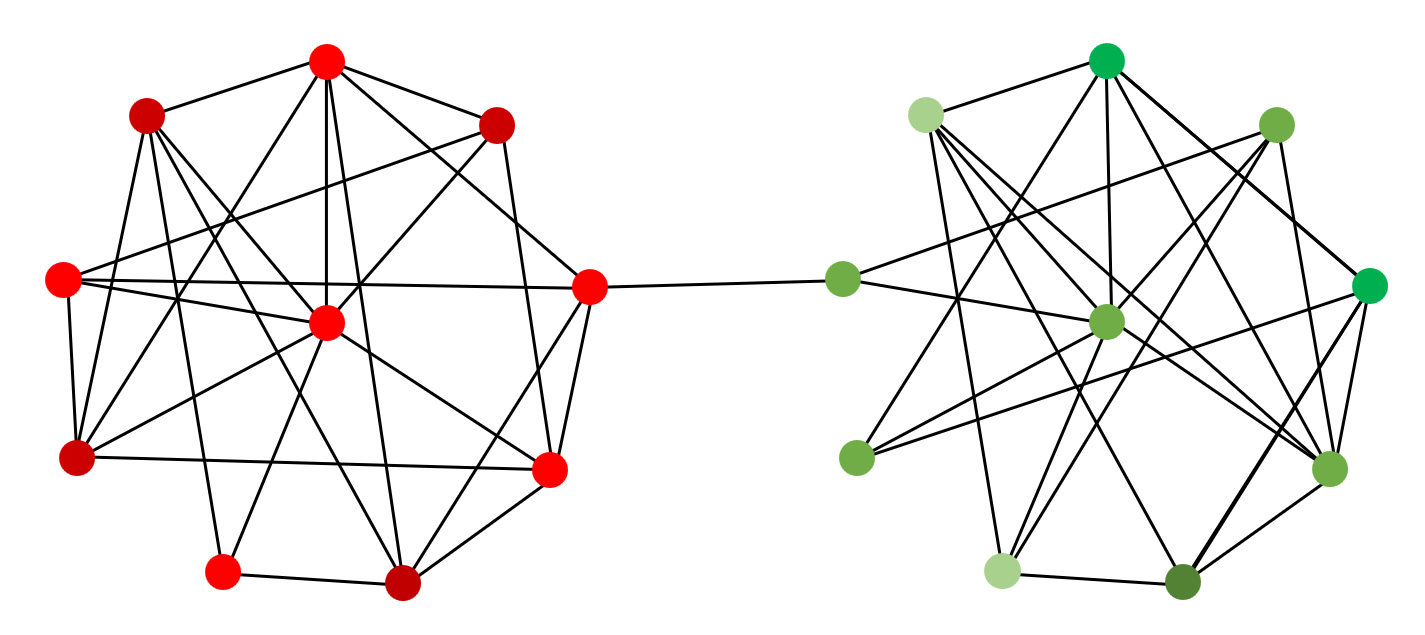}
}

$$
    h = \frac{1}{|E|}\sum_{(v_i,v_j)\in E}\mathds{1}(y_i=y_j),
$$

where $\mathds{1}$ is the indicator function evaluating to one when the labels of adjacent nodes are equal. The homophily level $h$ can take values between 0 and 1. We refer to graphs with low $h$ values as being heterophilic or non-homophilic. Most classical GNN architectures rely on the implicit assumption that graph labels are homophilic.

\greyline

\paragraph{Meshes and other Discrete Structures} Although the main focus in this section is on graphs, other structures such as meshes and simplicial complexes are also important in many computational applications. Rather than delving into the details, our goal here is to make the reader aware of the existence of such mathematical objects.

\begin{tcolorbox}[colback=gray!10, colframe=gray!40] A \textit{mesh} is a discrete representation of a geometric domain, typically composed of vertices, edges, and faces (often triangles or polygons) that approximate a continuous surface or manifold. \end{tcolorbox}

Meshes are widely used in computer graphics, geometry processing, and physical simulation such as in computational fluid dynamics and other engineering applications.

\begin{figure}[htbp!]
  \centering
  \includegraphics[width=0.4\linewidth]{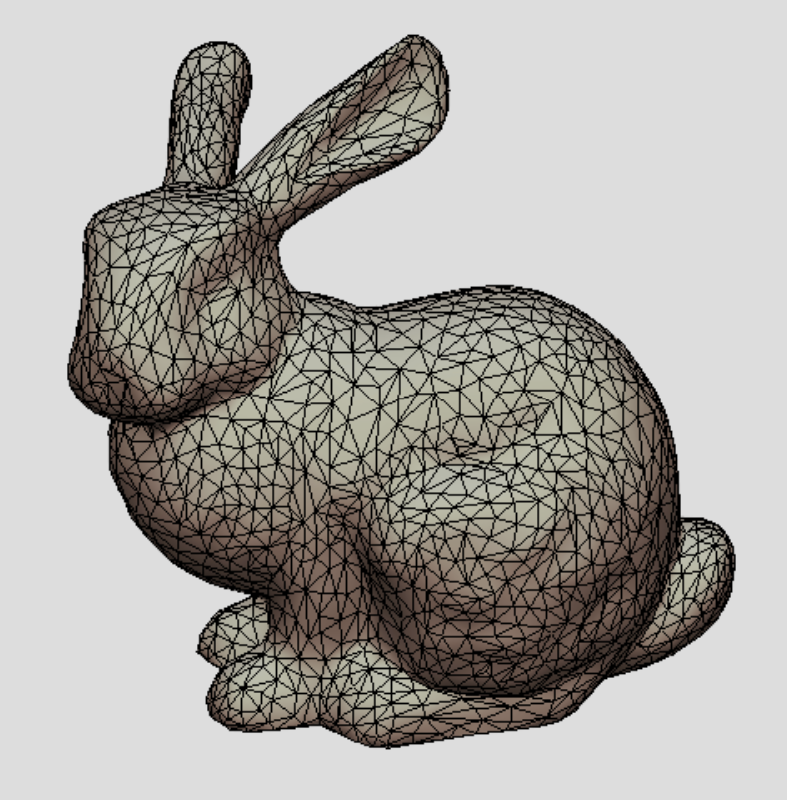}
  \caption{The Stanford Bunny is now one of the most recognizable 3D test models in computer graphics. It was originally developed by Greg Turk and Marc Levoy in 1994 at Stanford University.}
  \label{stanford bunny}
\end{figure}

\begin{tcolorbox}[colback=gray!10, colframe=gray!40] A \textit{simplicial complex} is a combinatorial object built from simplices (points, line segments, triangles, tetrahedra, etc.) that are glued together in a way that satisfies certain intersection and inclusion rules.\end{tcolorbox}

Simplicial complexes generalize meshes by allowing the construction of higher-dimensional elements. These structures allow for richer notions of locality and multi-scale representation, and are also key to extending graph-based methods into the realm of topological deep learning.

\greylinelong

\subsection{Group Theory and Graphs}

\paragraph*{Permutation-invariance} In many graph machine learning applications, it is important to preserve the structure of the data under reordering, since the numbering of the nodes is arbitrary to begin with. This is where symmetric groups and permutation-invariant aggregators come into play.

\begin{tcolorbox}[colback=gray!10, colframe=gray!40]
Let \( S \) be a set with \( |S| = N \). The \textit{symmetric group} of \( S \), denoted by \( S_N \), is the set of all bijections from \( S \) to itself:
\[
S_N = \{ \sigma : S \to S \mid \sigma \text{ is a bijection} \}.
\]
\end{tcolorbox}

\begin{tcolorbox}[colback=gray!10, colframe=gray!40]
A \textit{permutation-invariant aggregator} is a function \( \bigoplus: \mathcal{X}^N \to \mathcal{Y} \) that satisfies the condition

\[
\bigoplus(x_1, x_2, \dots, x_N) = \bigoplus(x_{\sigma(1)}, x_{\sigma(2)}, \dots, x_{\sigma(N)}),
\]

for any permutation \( \sigma \in S_N \), and \( \mathcal{X}^N \) denotes the set of all ordered tuples of \( N \) elements from the set \( \mathcal{X} \). 
\end{tcolorbox}

Common examples of permutation-invariant aggregators include summation $\sum_{i=1}^{N} x_i$, mean $\frac{1}{N} \sum_{i=1}^{N} x_i$, and maximum $\max_{i=1}^{N} x_i$, where $x_i$ are features vectors associated to each node $v_i$ as later discussed in Section~\ref{subsec:Vector Fields on Graphs}. These operations are commonly used at the end of GNN architectures to pool the features from all the nodes in the graph into a single feature vector which can be used for graph level classification or regression. 

Permutation matrices formalize the reordering or relabeling of nodes in a graph. Such reordering preserves the intrinsic graph structure, as the node labeling is arbitrary.

\begin{tcolorbox}[colback=gray!10, colframe=gray!40]
A \textit{permutation matrix} $P$ is a square binary matrix where exactly one entry in each row and each column is equal to 1, and all other entries are 0. Formally, for an $N \times N$ permutation matrix $P$, it holds that:
\[
P_{ij} = \begin{cases}
1 & \text{if node } i \text{ is mapped to node } j,\\[1ex]
0 & \text{otherwise}.
\end{cases}
\]
Such a matrix corresponds uniquely to an element of the symmetric group $S_N$.
\end{tcolorbox}

Permutation matrices are orthogonal, which implies that $P^{-1}=P^T$ and thus $PP^T=P^TP=I$, where $I$ is the identity matrix. When applying a permutation matrix $P$ to a graph with adjacency matrix $A$, the adjacency matrix transforms as follows:

\[
A' = P A P^T,
\]

where $A'$ is the permuted adjacency matrix, corresponding to the same graph with vertices relabeled according to $P$. Importantly, graph invariants such as the eigenvalues of the adjacency matrix, node degrees, and connectivity structure remain unchanged by permutations.

\greyline

\paragraph*{Graph Homomorphisms} Similar to group homomorphisms which allow us to relate equivalent groups that can be realized differently (Section~\ref{subsec:Groups}), graph homomorphisms provide a mathematical framework for studying mappings between graphs that preserve their structural properties. This can be particularly relevant in the context of network compression, graph colorings, and GNN expressivity analysis.

\begin{tcolorbox}[colback=gray!10, colframe=gray!40]
A \textit{graph homomorphism} is a mapping $F: V_G \to V_H$ between the vertex sets of two graphs $G = (V_G, E_G)$ and $H = (V_H, E_H)$ such that if $(v_i, v_j) \in E_G$, then $(F(v_i), F(v_j)) \in E_H$.
\end{tcolorbox}

Intuitively, a graph homomorphism maps edges of $G$ to edges of $H$, preserving the adjacency structure: if $v_i$ and $v_j$ are adjacent in $G$, their images $F(v_i)$ and $F(v_j)$ are adjacent in $H$. Note that in general, a homomorphism can map multiple vertices or edges of $G$ onto a single vertex or edge in $H$. This enables the simplification (or \textit{coarsening}) of graph structures while retaining connectivity properties.

\begin{tcolorbox}[colback=gray!10, colframe=gray!40]
A \textit{graph isomorphism} is a bijective mapping $F: V_G \to V_H$ between the vertex sets of two graphs $G = (V_G, E_G)$ and $H = (V_H, E_H)$ such that $(v_i, v_j) \in E_G$ if and only if $(F(v_i), F(v_j)) \in E_H$.
\end{tcolorbox}

Graph isomorphisms are a specific class of graph homomorphisms in which the mapping must be bijective, and the edge-preservation condition is bidirectional.

\begin{figure}[htbp!]
  \centering
  \includegraphics[width=\linewidth]{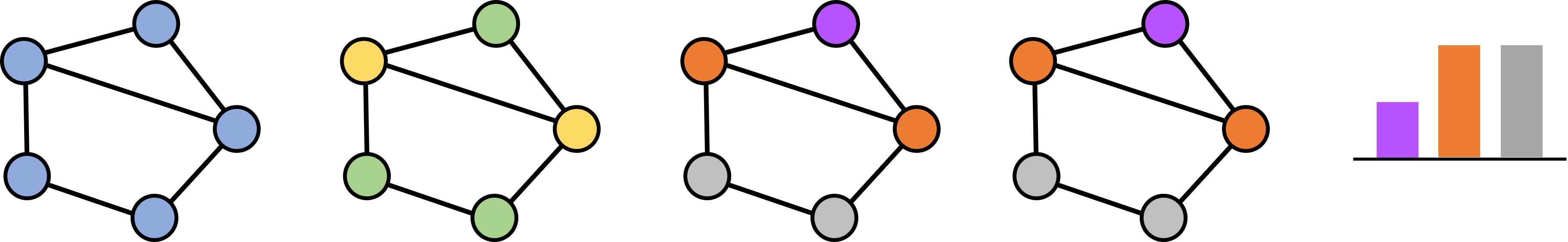}
  \caption{The Weisfeiler-Lehman (WL) test is a method used to determine whether two graphs are isomorphic by iteratively refining node labels based on their neighborhoods.}
  \label{wl}
\end{figure}

\greyline
\paragraph*{Examples of Graph Homomorphisms}

\begin{itemize}
\item Consider a cycle graph $C_6$ with six vertices and a complete graph $K_3$. A homomorphism $F: V_{C_6} \to V_{K_3}$ exists, where vertices of $C_6$ are mapped to vertices of $K_3$ in a repeating pattern.
\item For bipartite graphs, any homomorphism maps vertices in one partition to one set of vertices in the target graph and the other partition to the other set.
\item Let $P_{11}$ be a path graph with eleven vertices, and $C_{10}$ be a cycle graph with ten vertices. A homomorphism $F: V_{P_{11}} \to V_{C_{10}}$ exists where each vertex of $P_{11}$ is mapped to a vertex of $C_{10}$, and edges of $P_{11}$ are mapped to edges of $C_{10}$. Note that in this case the vertices at the start and end of the path graph would be mapped (or collapsed) to a single vertex.
\end{itemize}

\greylinelong

\subsection{Vector Fields on Graphs} 
\label{subsec:Vector Fields on Graphs}

Although so far our discussion has centered on graphs in terms of their connectivity structure, in practical scenarios and particularly in the context of Geometric Deep Learning, we primarily deal with graphs that have node attributes. Next, we consider graphs where each node has associated feature vectors and introduce relevant notation.

\begin{tcolorbox}[colback=gray!10, colframe=gray!40]
A \textit{feature vector} $ x_i $ at node $ v_i $ is a $ D $-dimensional vector that represents the characteristics or attributes of the node in the graph. 
\end{tcolorbox}

These vectors are organized into a matrix $ X \in \mathbb{R}^{N \times D} $ for all nodes $N=|V|$ in the graph. In the following expression, each entry $ x_{ij} $ represents the $ j $-th feature of node $ i $:

\[
X = \begin{bmatrix}
-x_1^\top- \\
-x_2^\top- \\
\vdots \\
-x_N^\top-
\end{bmatrix}
= \begin{bmatrix}
x_{11} & x_{12} & \cdots & x_{1D} \\
x_{21} & x_{22} & \cdots & x_{2D} \\
\vdots & \vdots & \ddots & \vdots \\
x_{N1} & x_{N2} & \cdots & x_{ND}
\end{bmatrix}.
\]

Equivalently, linking this discussion back to Section~\ref{subsec:Scalar Fields, Vector Fields and Signals}, we can define the feature vector field $ F $ as a mapping from the graph domain (nodes in the graph) to $\mathbb{R}^D$, where $D$ is the number of features for each node:

\[
F: V \to \mathbb{R}^D, \quad F(v_i) = x_i \in \mathbb{R}^D, \quad \forall v_i \in V.
\]

In geometric graphs, the matrix \( S \in \mathbb{R}^{N \times D} \) is sometimes used to denote scalar node features, while \( X \in \mathbb{R}^{N \times 3} \) is reserved to represent 3D coordinates, and \( V \in \mathbb{R}^{N \times 3} \) is used to represent additional geometric features.

\greyline

\paragraph*{Permuting Feature Vectors} Next, we give concrete examples, showing how the output produced by permutation-invariant aggregators remains unchanged when applying the permutation matrix to a matrix containing feature vectors. Let \(N=3\) and \(D=2\).  Suppose our node feature matrix is
\[
X = \begin{bmatrix}
1 & 2 \\  
3 & 4 \\  
5 & 6     
\end{bmatrix},
\]
so \(x_1 = [1,2]^\top,\;x_2=[3,4]^\top,\;x_3=[5,6]^\top\). Consider the permutation \(\sigma\) that swaps nodes 1 and 2 (and leaves 3 fixed).  The corresponding permutation matrix is
\[
P = 
\begin{bmatrix}
0 & 1 & 0 \\ 
1 & 0 & 0 \\ 
0 & 0 & 1
\end{bmatrix}.
\]
Applying \(P\) to \(X\) yields
\[
P\,X
= \begin{bmatrix}
0 & 1 & 0 \\ 
1 & 0 & 0 \\ 
0 & 0 & 1
\end{bmatrix}
\begin{bmatrix}
1 & 2 \\ 
3 & 4 \\ 
5 & 6
\end{bmatrix}
=
\begin{bmatrix}
3 & 4 \\  
1 & 2 \\  
5 & 6     
\end{bmatrix}.
\]
Check the sum:
\[
\sum_{i=1}^3 x_i
= 
\begin{bmatrix}1+3+5 \\ 2+4+6\end{bmatrix}
= \begin{bmatrix}9 \\ 12\end{bmatrix},
\qquad
\sum_{i=1}^3 (P X)_i
= 
\begin{bmatrix}3+1+5 \\ 4+2+6\end{bmatrix}
= \begin{bmatrix}9 \\ 12\end{bmatrix}.
\]
Thus \(\sum_i x_i = \sum_i (P X)_i = \sum_i P x_i = P \sum_i x_i\), illustrating permutation‑invariance. Also, it is trivial to verify that for the mean the same logic holds:

\[
\mathrm{mean}(X)
= \frac{1}{3}\sum_{i=1}^3 x_i
= \frac{1}{3}
\begin{bmatrix}1+3+5 \\ 2+4+6\end{bmatrix}
= \begin{bmatrix}3 \\ 4\end{bmatrix},
\]
\[
\mathrm{mean}(P\,X)
= \frac{1}{3}\sum_{i=1}^3 (P\,X)_i
= \frac{1}{3}
\begin{bmatrix}3+1+5 \\ 4+2+6\end{bmatrix}
= \begin{bmatrix}3 \\ 4\end{bmatrix}.
\]
Thus \(\mathrm{mean}(X)=\mathrm{mean}(P\,X)\). Finally, for the max:

\[
\max_{i=1}^3 x_i
= \begin{bmatrix}
\max\{1,3,5\} \\[0.5ex]
\max\{2,4,6\}
\end{bmatrix}
= \begin{bmatrix}5 \\ 6\end{bmatrix},
\]
\[
\max_{i=1}^3 (P\,X)_i
= \begin{bmatrix}
\max\{3,1,5\} \\[0.5ex]
\max\{4,2,6\}
\end{bmatrix}
= \begin{bmatrix}5 \\ 6\end{bmatrix}.
\]
Hence \(\max_i x_i = \max_i (P\,X)_i\).

\greyline

\paragraph*{The Graph Laplacian} The Laplacian plays a key role in analyzing graph structures, particularly in spectral graph theory.

\begin{tcolorbox}[colback=gray!10, colframe=gray!40]
The \textit{graph Laplacian} matrix $ L $ for a graph $ G = (V, E) $ is defined as:

\[
L = D - A,
\]
where $A$ and $ D $ are the adjacency and degree matrices of the graph, respectively.
\end{tcolorbox}

For undirected graphs, the graph Laplacian is symmetric and positive-semidefinite.

The quadratic form associated with the graph Laplacian can be written as:

\[
x^\top L x = x^\top (D - A) x = \sum_{i=1}^n d_i x_i^2 - \sum_{(v_i, v_j) \in E} w_{ij} x_i x_j = \frac{1}{2} \sum_{(v_i,v_j) \in E} w_{ij} (x_i - x_j)^2,
\]
where $ w_{ij} = w(e_{ij}) = w((v_i, v_j)) $ is the weight of the edge $ (v_i, v_j) $, and $ x_i $ and $ x_j $ are the feature values at nodes $ v_i $ and $ v_j $, respectively. Note that this is effectively computing a gradient-like quantity over the graph, which measures the smoothness of the vector field over the graph and is analogous to the Dirichlet energy in continuous settings. \marginnote{The Dirichlet energy is the continuous setting is the quadratic functional $\langle f, \Delta f\rangle = \langle \nabla f, \nabla f \rangle $.} It is often referred to as the \textit{graph Dirichlet energy} or simply the \textit{Dirichlet energy on a graph}.

\begin{tcolorbox}[colback=gray!10, colframe=gray!40]
The \textit{normalized graph Laplacian} matrix $ L_{\text{norm}} $ is defined as:

\[
L_{\text{norm}} = I - D^{-1/2} A D^{-1/2},
\]
where $ I $ is the identity matrix, $ A $ is the adjacency matrix, and $ D $ is the degree matrix.
\end{tcolorbox}

The form above\marginnote{The multiplicity of the eigenvalue refers to the number of times a specific eigenvalue appears in the spectrum of a matrix.} has several useful properties: the eigenvalues of $ L_{\text{norm}} $ lie in the range $[0, 2]$ and the multiplicity of the eigenvalue $ 0 $ corresponds to the number of connected components in the graph. 

\greyline

\paragraph*{Spectral Properties and Graph Frequencies} The eigenvectors of the graph Laplacian provide a natural generalization of the classical Fourier basis to graphs. This spectral perspective enables us to decompose signals over a graph into components of varying smoothness.

\begin{tcolorbox}[colback=gray!10, colframe=gray!40]
Let \( L \in \mathbb{R}^{N \times N} \) be the graph Laplacian of a graph \( G = (V,E) \) and $N=|V|$. Since \( L \) is symmetric and positive-semidefinite for undirected graphs, it admits an eigen-decomposition:
\[
L = U \Lambda U^\top,
\]
where \( U = [u_1, u_2, \dots, u_N] \) is an orthonormal basis of eigenvectors (the \textit{graph Fourier basis}) and \( \Lambda = \text{diag}(\lambda_1, \lambda_2, \dots, \lambda_N) \) is the diagonal matrix of eigenvalues.
\end{tcolorbox}

Each eigenvector \( u_k \) defines a basis function over the graph nodes, and its associated eigenvalue \( \lambda_k \) determines the `frequency' of that basis: lower eigenvalues correspond to smooth, slowly-varying functions over the graph, while higher eigenvalues capture more oscillatory variations. This frequency structure allows us to design filtering operations analogous to classical low-pass or high-pass filters.

\marginnote{The graph Laplacian eigenvectors form a global coordinate system over the graph. They can be leveraged as \textit{positional encodings} by assigning each node \( v_i \) a coordinate vector \( p_i = (u_1(i), u_2(i), \dots, u_k(i)) \) obtained from the first \( k \) nontrivial eigenvectors. These encodings are isomorphism-invariant and capture the intrinsic geometry of the graph.}

Given a signal \( f : V \to \mathbb{R} \) defined on the graph nodes, it can be expressed as a linear combination of these eigenvectors:
\[
f = \sum_{k=1}^{N} \langle f, u_k \rangle u_k.
\]
The coefficients \( \langle f, u_k \rangle \) constitute the graph Fourier transform of \( f \), allowing for the design of frequency-aware processing steps.

\begin{tcolorbox}[colback=gray!10, colframe=gray!40]
The \textit{graph Fourier transform} of a signal \( f \) is defined as \( \hat{f} = U^\top f \), and the inverse transform is given by \( f = U \hat{f} \).
\end{tcolorbox}

There will be as many eigenvectors as nodes in the graph. However, note that if there are degenerate eigenvalues (i.e., if an eigenvalue has multiplicity greater than one), the corresponding eigenvectors are not unique, but one can always choose an orthonormal basis consisting of \( N = |V| \) eigenvectors.

This spectral framework forms the foundation of many techniques in graph signal processing and also serves as a key tool in developing expressive architectures that incorporate both local and global graph structure.

\greyline

\paragraph*{Message-Passing on Graphs} For GNNs, we say we are \textit{learning a signal over a graph}, where the graph structure guides the flow of information between nodes. Typically, the graph on which the signal is defined is coupled with the computational graph of the artificial neural network.

More concretely, a message passing GNN layer $l$ over a graph $G$ is computed as

$$
x_{i}^{(l+1)}=\phi\Big(x_{i}^{(l)},\bigoplus_{j\in\mathcal{N}(v_{i})}\psi(x_{i}^{(l)},x_{j}^{(l)})\Big),
    \label{mp_gnn}
$$

\noindent
\marginnote{Transformers perform attentional message passing over a fully connected graph. Alternatively, one can interpret the attention scores as `discovering' the underlying graph.

\includegraphics[width=1\linewidth]{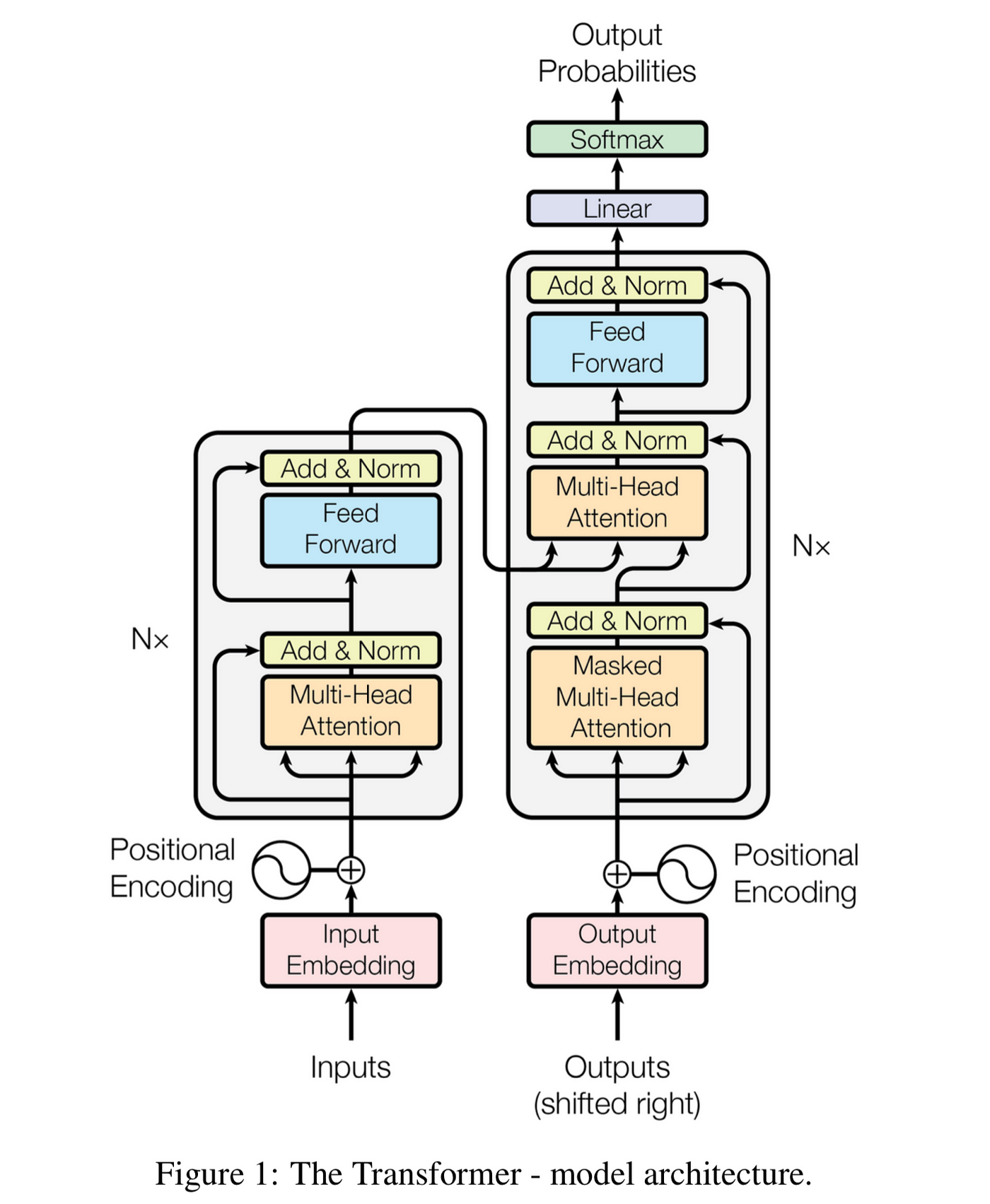}}where $\psi$ and $\phi$ are non-linear functions, and $\bigoplus$ is an aggregation function, which must be permutation-invariant. The above equation constrains the information flow for each layer to local neighbourhoods and can be further decomposed into three update rules:

\begin{equation}
    m_{ij}^{(l)} \leftarrow \psi(x_{i}^{(l)},x_{j}^{(l)}), \tag{Message}
\end{equation}

\begin{equation}
    a_{i}^{(l)} \leftarrow \bigoplus_{j\in\mathcal{N}(v_{i})} m_{ij}^{(l)}, \tag{Aggregate}
\end{equation}

\begin{equation}
    x_{i}^{(l+1)} \leftarrow \phi\Big(x_{i}^{(l)},a_{i}^{(l)}\Big). \tag{Update}
\end{equation}

\begin{tcolorbox}[colback=orange!20, colframe=orange!60]
\textbf{Graph Theory in Geometric Deep Learning.} Graph theory plays a central role in Geometric Deep Learning, particularly in the context of GNNs, which are designed to learn signals over graph structures. The underlying graph domain serves as a geometric prior, typically assuming that connected nodes share similar features. GNNs have been applied to diverse areas, including social networks, recommendation systems, and bioinformatics, for both supervised learning and generative modeling.
\end{tcolorbox}

\clearpage

\bibliographystyle{unsrtnat}
\bibliography{references}

\end{document}